\documentclass[pdflatex,sn-apa]{sn-jnl}

\usepackage{graphicx}
\usepackage{multirow}
\usepackage{amsmath,amssymb,amsfonts}
\usepackage{amsthm}
\usepackage{mathrsfs}
\usepackage[title]{appendix}
\usepackage{xcolor}
\usepackage{textcomp}
\usepackage{manyfoot}
\usepackage{booktabs}
\usepackage{algorithm}
\usepackage{algorithmicx}
\usepackage{algpseudocode}
\usepackage{listings}
\usepackage{subcaption}
\usepackage{float}
\theoremstyle{thmstyleone}

\usepackage{makecell}
\theoremstyle{thmstyletwo}

\theoremstyle{thmstylethree}

\begin{document}

\title[Article Title]{``Let's Agree to Disagree'': Investigating the Disagreement Problem in Explainable AI for Text Summarization}


\author[1]{\fnm{Seema} \sur{Aswani}}\email{p20210908@dubai.bits-pilani.ac.in}

\author*[2]{\fnm{Sujala D.} \sur{Shetty}}\email{sujala@dubai.bits-pilani.ac.in}

\affil*[1,2]{\orgdiv{Department of Computer Science}, \orgname{BITS Pilani, Dubai Campus}, \orgaddress{\street{Academic City}, \city{Dubai}, \country{U.A.E}}}

\abstract{Explainable Artificial Intelligence (XAI) methods in text summarization are essential for understanding the model behavior and fostering trust in model-generated summaries. Despite the effectiveness of XAI methods, recent studies have highlighted a key challenge in this area known as the "disagreement problem". This problem occurs when different XAI methods yield conflicting
explanations for the same model outcome. Such discrepancies raise concerns about the consistency of explanations and reduce confidence in model interpretations, which is crucial for secure and accountable AI applications. This work is among the first to empirically investigate the disagreement problem in text summarization, demonstrating that such discrepancies are widespread in state-of-the-art summarization models. To address this gap, we propose Regional Explainable AI (RXAI) a novel segmentation-based approach, where each article is divided into smaller, coherent segments using sentence transformers and clustering. We use XAI methods on text segments to create localized explanations that help reduce disagreement between different XAI methods, thereby enhancing the trustworthiness of AI-generated summaries. Our results illustrate that the localized explanations are more consistent than full-text explanations. The proposed approach is validated using two benchmark summarization datasets, Extreme summarization (Xsum) and CNN/Daily Mail, indicating a substantial decrease in disagreement. Additionally, the interactive JavaScript visualization tool is developed to facilitate easy, color-coded exploration of attribution scores at the sentence level, enhancing user comprehension of model explanations.
}


\maketitle

\section{Introduction}\label{sec1}
Text summarization condenses lengthy documents into meaningful summaries by emphasizing crucial information \citep{WIDYASSARI20221029}. Due to the exponential growth of information, its applications span diverse fields such as news, clinical text, and legal documents summarization \citep{SUPRIYONO2024100070}. Summarization models based on the transformers architecture \citep{vaswanietal} have significantly advanced text summarization, resulting in fluent and contextually meaningful summaries. However, due to their black-box nature, these models have raised concerns about reliability, transparency, and biases. \citep{holzinger2020explainable}

Explainable Artificial Intelligence (XAI) tries to solve these concerns with techniques that provide understandable explanations of a complex AI model \citep{Saranyaetal}. In text summarization, the XAI methods are intended to provide transparent and interpretable explanations for why certain words/sentences were included in or excluded from the summary. This work focuses on state-of-the-art abstractive summarization models - BART \citep{lewis-2020-bart} and PEGASUS \citep{zhang20ae}, fine-tuned on the XSum and CNN/Daily Mail datasets, respectively. These datasets present complementary challenges, with XSum providing single-sentence and CNN/Daily Mail providing multi-sentence summaries.

Despite the effectiveness of XAI methods, they face a key challenge known as the disagreement problem, where different methods produce conflicting explanations \citep{krishna2024etal}. This issue erodes trust and presents significant challenges in high-stakes domains such as healthcare, finance, and law.  For instance, conflicting explanations can affect crucial decisions in fraud detection, disease diagnosis, or legal compliance, leading to security breaches or ethical failures. It is essential to address the disagreement problem not only for enhanced interpretability but also to ensure secure and trustworthy deployment of AI. While efforts have been made to quantify the disagreement in other domains, no prior work has explored potential solutions tailored
specifically for text summarization models. To address and mitigate this issue we introduce a novel
Regional Explainable AI (RXAI) framework which segments the articles into semantically
coherent regions using Sentence Transformers \citep{Songea} and clustering. To quantify the disagreement based on the underlying meaning of explanations we propose a semantic similarity-based metric called Semantic Alignment Score (SAS). As the final contribution, we present an open-source JavaScript-based tool that generates interactive sentence-level attribution visualization, providing insights into the contributions of input sentences to model predictions\footnote{Access the tool here: \url{https://github.com/SVC04/Explainable-Summarization}}.\\
Following are the main Research Questions(RQ) that we aim to address in this study.


\begin{itemize}
   
\item \textbf{RQ1}: To what extent do different XAI methods agree when applied to text summarization models? 
\item \textbf{RQ2}: How do dataset characteristics such as the input article length and summary type affect the disagreement among XAI methods in text summarization?
\item \textbf{RQ3}: Does the proposed RXAI framework, through text segmentation reduce disagreement between XAI methods in text summarization? 

\end{itemize}
In summary, the following are the key contributions of our research:
\begin{enumerate}
\item This paper offers a novel empirical analysis of the disagreement problem among different XAI methods applied to text summarization models, utilizing both predefined metrics and the proposed semantic similarity-based metric.
\item We propose RXAI, a novel framework that leverages segmentation-based explanations using sentence embeddings and k-means clustering to mitigate the disagreement problem.

\item We have developed an open-source interactive JavaScript visualization tool to analyze sentence-wise attribution scores for text summarization models.
\end{enumerate}

\section{Related Work}\label{sec2}
\subsection{XAI Methods and Attribution Generation Framework}
XAI methods have gained popularity for interpreting complex deep-learning models. In text summarization, XAI methods like LIME \citep{Ribeiroae}, Gradient SHAP \citep{Lundbergea}, attention mechanism \citep{bahdanau2014neural}, and DeepLIFT \citep{shrikumarea} provide diverse perspectives for understanding model decisions. LIME approximates a model locally by perturbing input sentences, Gradient SHAP approximates Shapely values using gradient-based expectations, attention mechanism helps transformer models focus on relevant parts of the input text, while DeepLIFT assigns attribution scores by comparing each neuron’s activation to a reference activation. Appendix \ref{secA1} contains a detailed description of these methods.

Transformer-based models applied for text generation tasks require specialized tools for attribution generation. The Inseq library \cite{sarti-etal} built on Captum \cite{kokhlikyan2020captum}, supports standard XAI methods for token and sentence-level attribution generation and an interactive heatmap visualization that enhances the interpretability of the seq2seq model prediction. 

\subsection{The Disagreement Problem and Its Evaluation}
The disagreement problem in XAI occurs when different methods give divergent explanations for the same input for feature importance, ranking, or directional influence \citep{krishna2024etal}. Prior research has made significant progress in validating the interpretability using ROAR (Remove and Retain) \citep{Saraea}. It has also focused on evaluating discrepancies in explanation quality with an emphasis on fidelity, stability, consistency, and sparsity \citep{Jessicaea}, as well as benchmarking feature attribution techniques with synthetic datasets \citep{liuea}. However, the crucial task of assessing the inconsistency in explanations is understudied. \cite{krishna2024etal} points out that most of the XAI methods differ in explanations and emphasize the need for systematic approaches to address them.
 
Previous studies have examined various disagreement metrics. \cite{neely2022song} discovered low correlations between XAI methods, questioning the validity of agreement-based evaluations. \cite{Royea} suggested an aggregated approach to reduce disagreement between LIME and SHAP through prioritizing shared features. These results highlight the need for consistent metrics to evaluate and control XAI disagreement.


\subsection{Proposed Solutions for Disagreement in XAI}
Recent studies have proposed promising solutions to tackle the disagreement problem such as the AGREE framework \cite{Piereae} and the application of FD-Trees \cite{laberge24a}. The FD trees execute localized feature interactions while the AGREE framework mitigates the problem through the use of alignment-based metrics. These frameworks highlight the potential of alignment-based strategies and the potential of segmentation in tackling disagreement problem. 

This work addresses the disagreement problem in text summarization by extending previous research. Building on local explanations \citep{laberge24a} and different agreement measurement strategies, we present the RXAI framework as a novel tool to improve the consistency of explanations between XAI methods.

\section{Methodology}
\subsection{Phase A: Disagreement Analysis}

This work follows a two-phase approach to measure and reduce disagreement between XAI methods. Phase A measures global disagreement on complete articles over two benchmark summarization datasets—XSum and CNN/Daily Mail. The system architecture of Phase A is illustrated in Fig. \ref{fig:phase_a}. We examined 11 separate random batches of 500 articles each (5500 total per dataset) to ensure robustness, reduce sampling bias, and assess statistical consistency between runs. This batched setup enhances reliability and facilitates Phase B, where analysis for each batch guarantees computational feasibility while maintaining generalizability. Appendix \ref{seca2} contains detailed statistics of both datasets. 

The input articles were tokenized to a maximum length of 1,024 tokens, utilizing the Hugging Face implementation \citep{wolf-etaltransformers}. As shown in Fig. \ref{fig:phase_a} the input articles were preprocessed and summarized using two state-of-the-art pre-trained models: BART \citep{lewis-2020-bart} fine-tuned on the XSum dataset, and PEGASUS \citep{zhang20ae}, fine-tuned on CNN/Daily Mail without additional fine-tuning. We employed four diverse XAI methods: attention (model-intrinsic), DeepLIFT (activation and backpropagation-based), LIME (perturbation-based), and Gradient SHAP (gradient-based) to generate attribution at token and sentence level. Each XAI method uses different input configurations and explanation strategies, allowing a comprehensive analysis. To quantify agreement between these attributions, both the predefined metrics proposed by \cite{krishna2024etal} and proposed semantic similarity-based metrics were utilized. To represent the resulting pairwise agreement scores we have used heatmaps and some other form of visualizations.



    
    

\subsection{Phase B: Tackling Disagreement through RXAI}
\begin{figure}[H]
    \centering
    \begin{subfigure}[t]{0.48\textwidth}
        \centering
        \includegraphics[width=\textwidth]{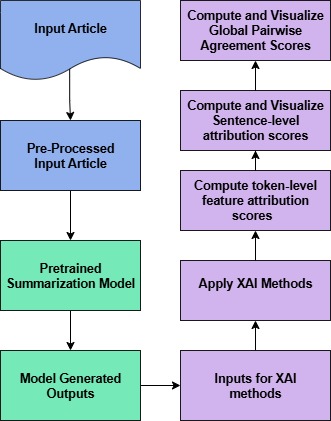}
        \caption{Phase: A Global agreement analysis}
        \label{fig:phase_a}
    \end{subfigure}
    \hspace{0.03\textwidth}
    \begin{subfigure}[t]{0.46\textwidth}
        \centering
        \includegraphics[width=\textwidth]{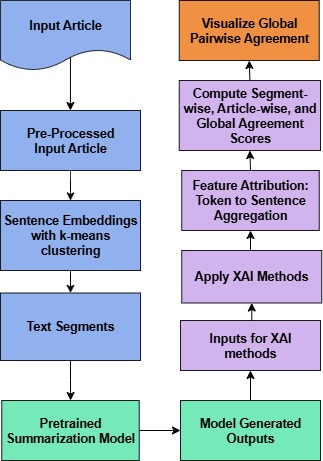}
        \caption{Phase: B RXAI framework for regional analysis}
        \label{fig:phase_b}
    \end{subfigure}
    \caption{System architectures for analyzing and addressing the disagreement problem in XAI methods. 
    (a) Phase A performs global agreement analysis. 
    (b) Phase B applies RXAI for regional evaluation. These phases
can be executed in parallel, enabling a transition from a global to a more fine-grained analysis for improved explanation consistency.}
    \label{fig:system_architectures}
\end{figure}

Phase B discusses the proposed RXAI framework, which segments input text into semantically coherent regions using Sentence Transformers \citep{Songea} and \texttt{k}-means clustering method \citep{hartigan1979algorithm}. The segmentation helps to analyze the disagreement problem at segment level, article level, and across all articles. By focusing on regional explanations, the RXAI framework complements the full-text disagreement analysis performed in Phase A and offers insights into the impact of explanation consistency at the segment level. Fig. \ref{fig:phase_b} outlines the RXAI system architecture.

\subsection{Customized Pre-processing steps}
Customized pre-processing steps were applied to the input articles to ensure precise detection of sentence boundaries, a critical factor for aggregating token-level attribution scores into sentence-level attribution. Periods are considered as indicators of sentence ending. The precise identification of sentence boundaries facilitates the calculation of sentence-wise attribution scores, in contrast to previous studies that focused on token-level attributions.
Following is the list of pre-processing steps applied:


\begin{itemize}
    \item \textbf{Adjusting Sentence Boundaries and Punctuation:} This step is added to ensure proper sentence endings by correcting misplaced periods near quotation marks, inserting missing periods, and standardizing punctuation such as question marks paired with periods.
    \item \textbf{Reducing Redundant Periods:} To avoid confusion in token aggregation multiple consecutive periods are replaced with a single period in this step.
    \item \textbf{Handling Non-Standard Period Usages:} 
    In this step, the periods in initials, URLs, email addresses, and decimal numbers are replaced with special tokens (\texttt{[NAME\_PERIOD\_TOKEN]}, \texttt{[WEB\_PERIOD\_TOKEN]}, \texttt{[EMAIL\_PERIOD\_TOKEN]}, \texttt{[NUMBER\_PERIOD\_TOKEN]}) to avoid errors in boundary detection.
\end{itemize}

\subsection{Attribution Computation and Interactive Visualization}
The four XAI methods: LIME, Gradient SHAP, Attention, and DeepLIFT were implemented using Inseq library to explain the results of text summarization models.\\
The Inseq library generates attribution scores token-wise using \texttt{model.attribute()} function. These token-wise scores are aggregated into sentence-wise scores using the \texttt{aggregate()} function. The attribution scores are normalized within the range of [0,1], ensuring consistency and comparability among attribution scores derived from several XAI methods.


To facilitate visual understanding of the link between the input text and the model-generated summary based on the generated attribution scores, we developed an interactive JavaScript-based tool. The tool accepts three inputs: Input Text Sentences, sentence-wise attribution score, and model-generated summary. Sentences are derived from the truncated input article, which is tokenized and shortened to the model’s maximum token limit. The \texttt{inseq} library provides normalized
token-level attribution scores, aggregated into sentence-level. However, the scale of the score differs across XAI methods. To
ensure consistency and enable comparability across methods, an additional normalization step is applied to standardize the scores by applying the formula below:

\[
\text{Normalized Score} = \frac{\text{Score} - \text{Min}}{\text{Max} - \text{Min}}
\]

\begin{figure}[h]
    \centering
    \includegraphics[width=0.75\textwidth]{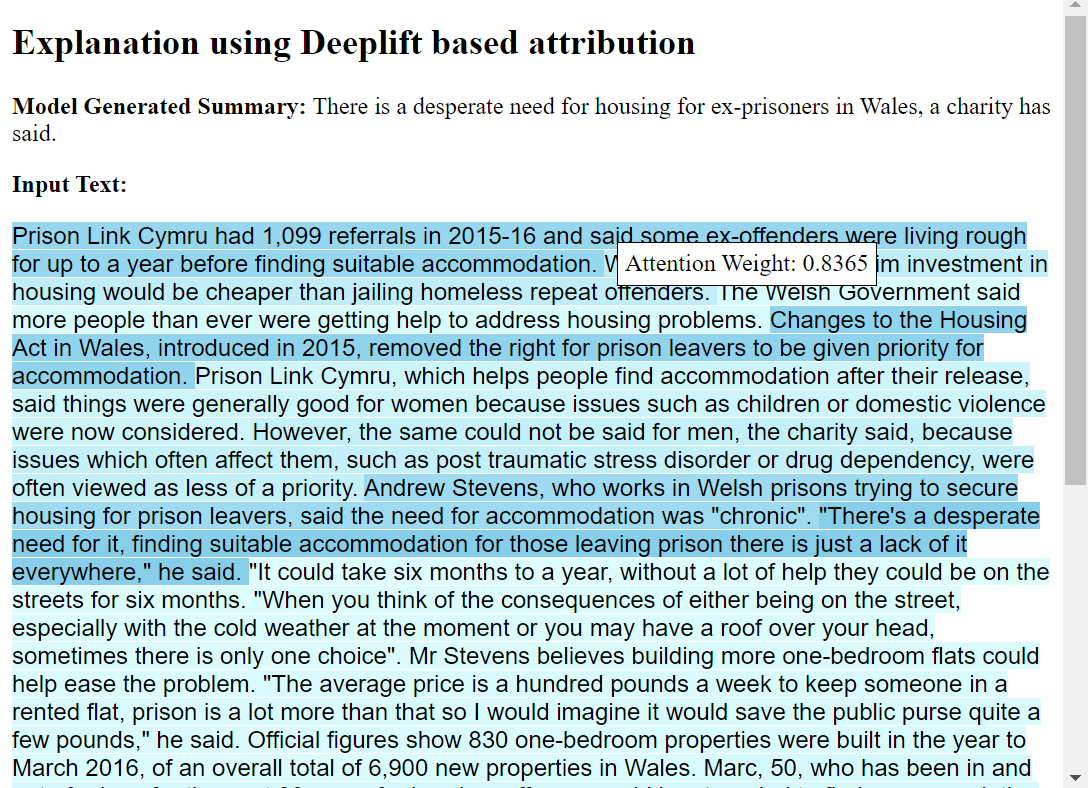}
    \caption{The text plot of the attribution weights assigned over the input sentences using DeepLIFT. Normalized attribution scores are represented using color coding, where darker-colored sentences represent higher attribution weights, signifying the greater contribution to the resultant summary.}
    \label{fig:Deeplift_Attribution}
\end{figure}

Upon receiving these inputs, the tool generates a color-coded plot where each sentence is assigned a color gradient (light blue to dark blue) based on its attribution score, with darker shades indicating higher importance. The tool offers an interactive hover functionality that allows users to view exact attribution scores for each sentence, enhancing transparency. For a detailed view of the tool's user interface, refer to Fig. \ref{fig:Js_textplot}  in the Appendix. 


Fig. \ref{fig:Deeplift_Attribution} presents the resulting visualization, which highlights the contributions at the sentence level to the generated summary with the help of DeepLIFT XAI method. The visualization tool provides an effective way to compare the attribution of input sentences to the generated summary, aiding in enhanced interpretability.

\subsection{Quantifying Disagreement between XAI methods}
We employed existing metrics proposed by \cite{krishna2024etal} to evaluate the explanation agreement. These metrics quantify agreement at several levels. For the comparison of explanations between XAI methods, we apply four existing metrics: Feature Agreement, Rank Agreements, Spearman Rank Correlation, and Pairwise Rank Agreement (PRA) \citep{krishna2024etal}. Additionally, we introduce a novel semantic similarity-based metric SAS that quantifies disagreement at the semantic level by comparing attribution-weighted sentence embeddings using cosine similarity. These metrics measure agreement based on the most significant (top-k) and relative feature ranking. Appendix \ref{seca3} provides specific details on these measures. The selection of top-k values is inspired by \cite{krishna2024etal} and adjusted to account for structural differences in the analysis. For global disagreement analysis we select $k \in [2, 10]$, which covers ~25-50\% of content in documents averaging 20-40 sentences. For RXAI we choose  $k \in [2, 4]$ which covers 44--66\% of shorter segments averaging 6-9 sentences.

    
    The Algorithm \ref{Globalalgo} outlines the steps of computing global agreement scores for text summarization datasets. For each XAI method the attribution scores are derived, followed by pairwise top-k and relative ranking agreement computations. The pairwise agreement scores of all the articles are aggregated to achieve global agreement scores. These scores were then analyzed using different visualizations.

\begin{algorithm}[H]  
\small
\caption{Compute top-k and Relative Ranking Agreement} \label{Globalalgo}
\begin{algorithmic}[1]
    \State \textbf{Input:} $sample\_news\_articles$, $methods = [\texttt{attention}, \texttt{deeplift}, \texttt{lime}, \texttt{gradient\_shap}]$
    \State \textbf{Output:}  Pairwise agreement scores based on top-k and relative rankings

    \State Preprocess $sample\_news\_articles$ and store it in $preprocessed\_articles$
     \State Initialize $global\_results \gets [\ ]$ 

    \For{each $article$ in $preprocessed\_articles$}
        \For{each $method$ in $methods$}
            \State Load $model\_instance$ with $method$
            \State $out \gets model\_instance.attribute(article)$ 
            \State Aggregate sentence spans from $out$ and assign to $global\_results$
        \EndFor
    \EndFor

    \For{each $(method1, method2)$ pair in $methods$}
        \For{$k = 2$ to $11$}
            \State Compute and store average top-k agreement
        \EndFor
        \State Compute and store:\newline
        (a) Average Spearman Correlation\newline
        (b) Average Pairwise Rank Agreement
    \EndFor

    \State Generate correlation matrices and visualizations of agreement
\end{algorithmic}
\end{algorithm}

\subsection{RXAI: Tackling the Disagreement Problem}
To tackle the disagreement problem among XAI methods we employed the RXAI approach, hypothesizing that segmenting the input article into semantically coherent segments would reduce the level of disagreement between explanations. Algorithm \ref{RXAIalgo} outlines the RXAI framework. First, the input articles are pre-processed and truncated into 1024 tokens to match the input length handled by pre-trained transformer models. Due to the high accuracy, all-mpnet-base-v2 sentence transformer \citep{Songea} is utilized to transform
input sentences into high-dimensional embeddings followed by applying k-means clustering \citep{hartigan1979algorithm} for segmentation. The value of optimal-\texttt {k} is determined by maximizing the Silhouette Score \citep{ROUSSEEUW198753} on a range $\texttt{k} \in [2, \min(10, \texttt{num\_sentences}-1)]$, to ensure clustering quality and preserving over segmentation. In cases where multiple k-values yield the same maximum silhouette score we applied a tie-breaking strategy that chooses smallest k-value. The rationale behind this choice is: (i) computational efficiency, since a smaller number of clusters decreases the overhead of segment-level attribution; and (ii) semantic coherence, since larger segments maintain contextual flow better, which enhances explanation reliability. The clusters with fewer than two sentences are merged with the most similar cluster using cosine similarity. The \texttt{k}-values across articles were averaged for uniformity. Segment-level attribution scores are computed for each XAI method. These scores at segment level are aggregated into article-level disagreement scores and visualized through pairwise heatmaps for both segment and article-level analysis.

\texttt{k}-means was chosen for its balance between segmentation quality and computational feasibility. While DBSCAN \citep{Esterea} can deal with arbitrarily shaped clusters, it does not work well with high-dimensional sentence embeddings and needs careful fine-tuning of the $\epsilon$ parameter. In contrast, \texttt{k}-means offers more control over the number of clusters (\texttt{k}). Agglomerative clustering \citep{mullner2011modern} effectively captures hierarchical relationships, but it was not employed due to its high computational complexity, making it challenging to cluster the high-dimensional sentence embeddings.

\begin{algorithm}[H] 
\small
\caption{Regional Explanations and Agreement Calculation} \label{RXAIalgo}
\begin{algorithmic}[1]
    \State \textbf{Input:} $sample\_news\_articles$, $methods = [\texttt{attention}, \texttt{deeplift}, \texttt{lime}, \texttt{gradient\_shap}]$
    \State \textbf{Output:} Average agreement scores and visualizations

    \State Preprocess and tokenize $sample\_news\_articles$ into $preprocessed\_articles$ (truncate to 1024 tokens)
    \State Encode sentences and determine optimal $k$ for k-means clustering
    \State Compute average $k$ for uniform segmentation

    \For{each $article$ in $preprocessed\_articles$}
        \State Segment $article$ using K-means with average $k$
        \State Merge small clusters (fewer than two sentences) into the most similar cluster
        \State Store segmented $article$ in $preprocessed\_segments$
    \EndFor

    \For{each $segmented\_article$ in $preprocessed\_segments$}
        \For{each $method$ in $methods$}
            \State Compute attribution scores for each segment using the XAI method
        \EndFor
        \State Aggregate segment-level scores into article-level disagreement
    \EndFor

    \State Compute average disagreement scores across all articles
    \State Generate pairwise agreement heatmaps and visualizations
\end{algorithmic}
\end{algorithm}

\section{Results}
\subsection{Disagreement Analysis Across XAI Methods} \label{Disgreement_results} The results section is divided based on three main research questions this study has focused on. The disagreement analysis is conducted on Xsum and CNN/DM datasets using metrics discussed in appendix \ref{seca3}. The results are evaluated on a total of 5,500 test samples from each dataset, examined across 11 random batches to ensure statistically robust results and computational feasibility for attribution generation for global and RXAI analysis. To assess the consistency and mean difference of agreement scores across all batches, we used one-way ANOVA for each method pair and k-value separately. The results of ANOVA revealed highly consistent results across the agreement metrics Feature, Rank, Semantic, and Rank Correlation agreement for Xsum (with p-value higher than 0.05) and moderate batch sensitivity for CNN/DM. Refer to Section: \ref{stats_test} for detailed results. While the entire study covers all 11 batches, this section provides detailed results and visualizations based on a single representative batch of 500 articles.
This section presents the analysis of the degree of disagreement between XAI methods (RQ1) and assesses the influence of dataset characteristics on disagreement (RQ2).

\subsubsection{\textbf{RQ1.}To what extent do different XAI methods agree when applied to text summarization models?} \label{global_analysis}

\begin{enumerate}

\item \textbf{Feature Agreement Analysis:} The feature agreement across both datasets is illustrated in Fig. \ref{fig:feature_agreement_side_by_side}. attention vs. DeepLIFT consistently shows the highest feature agreement for the Xsum dataset, which indicates that the two methods mostly highlight the same top features. For the CNN/DM dataset attention vs DeepLIFT exhibited the highest agreement up to \texttt{k = 5}. However, for \texttt{k > 5} the attention vs Gradient SHAP surpasses attention vs. DeepLIFT, indicating that the attribution assigned by Gradient SHAP aligns more closely with attention-based explanations when considering a larger set of features. The LIME-based method pairs exhibited lowest agreement scores. The feature agreement value increases as \texttt{k} increases. Variability analysis of feature agreement scores is depicted in Fig. \ref{fig:FA_boxplot_side_by_side} in Appendix. 

\begin{figure}[H]
    \centering
    \begin{subfigure}[b]{0.49\linewidth}
        \centering
        \includegraphics[width=\textwidth]{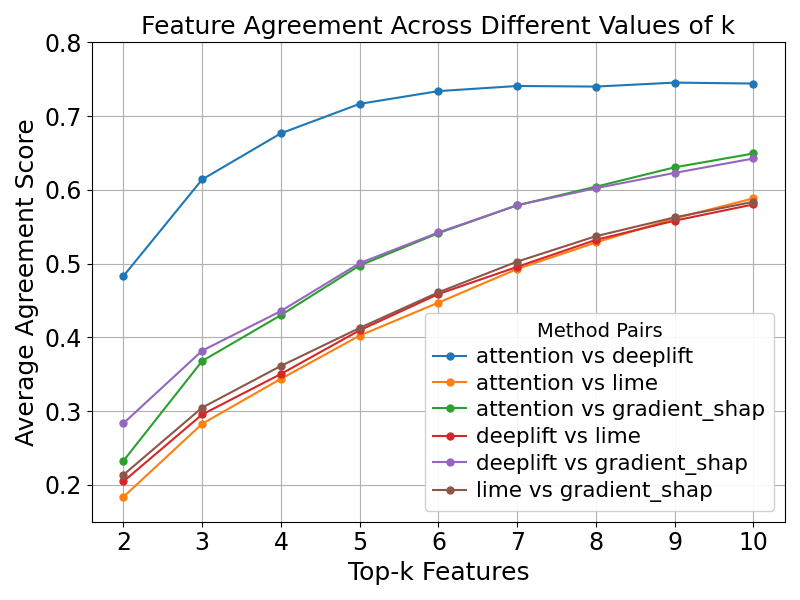}
        \caption{Feature agreement across different k-values for Xsum.}
        \label{fig:line_plot_xsum}
    \end{subfigure}
    \hfill
    \begin{subfigure}[b]{0.49\textwidth}
        \centering
        \includegraphics[width=\textwidth]{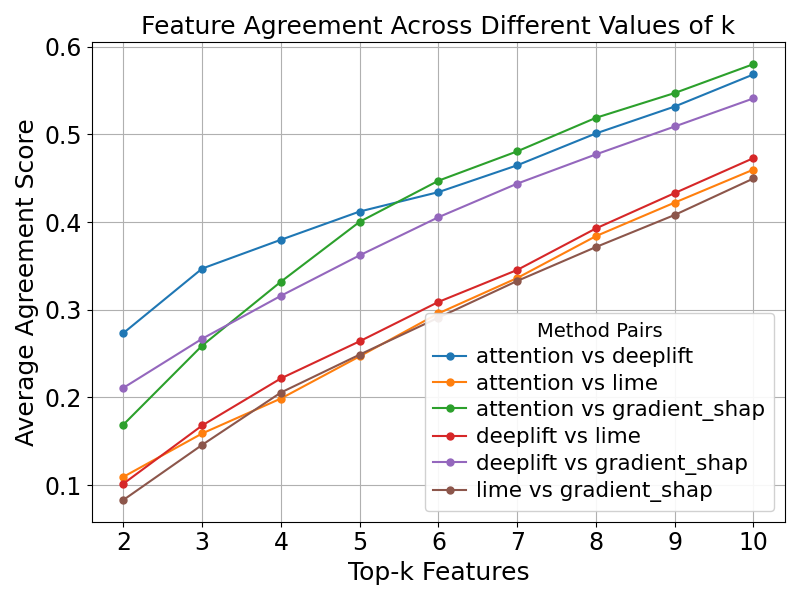}
        \caption{Feature agreement across different k-values for CNN/DM.}
        \label{fig:Line_plot_CNN}
    \end{subfigure}
    \caption{Average feature agreement scores across top-k (k=2 to 10) features for Xsum (Fig. \ref{fig:line_plot_xsum}) and CNN/DM (Fig. \ref{fig:Line_plot_CNN})datasets. Feature agreement scores increase as the value of k increases. For both the datasets mostly the attention vs. DeepLIFT exhibits the highest feature agreement. Feature agreement values for the CNN/DM  dataset are slightly low in magnitude compared to Xsum feature agreement scores.}
    \label{fig:feature_agreement_side_by_side}
\end{figure}

\item \textbf{Rank Agreement Analysis:} The rank agreement scores in Fig. \ref{fig:Rank_agreement_side_by_side} highlights that attention vs. DeepLIFT gives the highest rank agreement score, while the overall value of agreement is relatively low for both datasets. For all method pairs, rank agreement decreases consistently as the value of k increases. This indicates that, with more features taken into consideration, the methods disagree not only in which features they highlight but also in how they rank them. Fig. \ref{fig:RA_boxplot_side_by_side} in appendix \ref{Seca4} illustrates the variability of rank agreement scores across all method pairs.

\begin{figure}[H]
    \centering
    \begin{subfigure}[b]{0.49\linewidth}
        \centering
        \includegraphics[width=\textwidth]{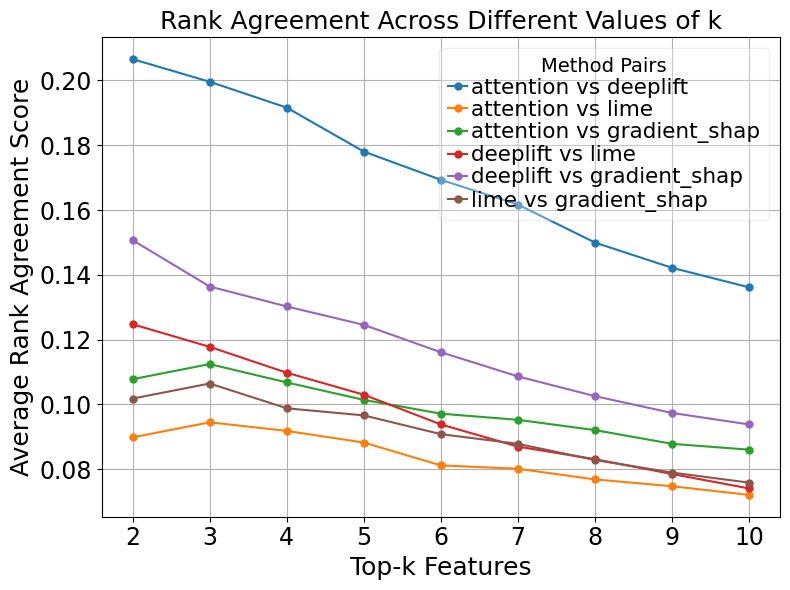}
        \caption{Rank agreement across different k-values for Xsum.}
        \label{fig:RAline_plot_xsum}
    \end{subfigure}
    \hfill
    \begin{subfigure}[b]{0.49\textwidth}
        \centering
        \includegraphics[width=\textwidth]{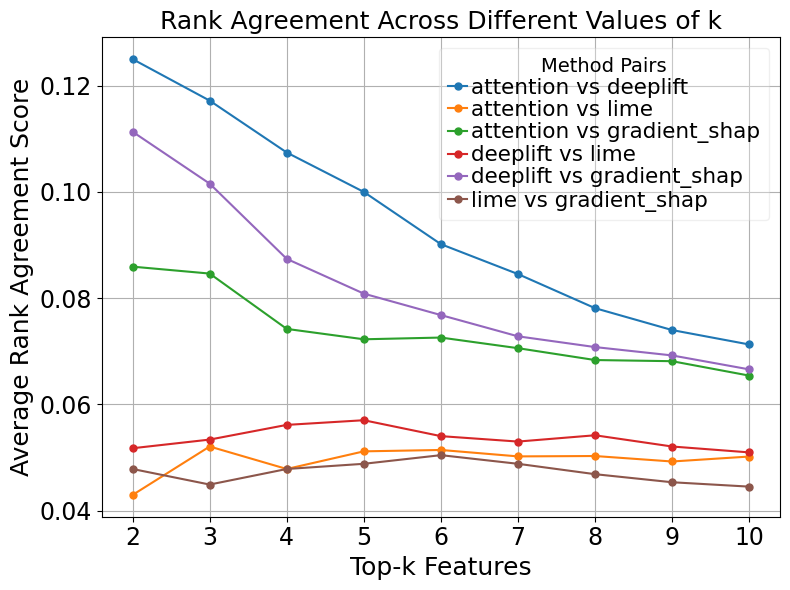}
        \caption{Rank agreement across different k-values for CNN/DM.}
        \label{fig:RALine_plot_CNN}
    \end{subfigure}
    \caption{Average rank agreement scores across top-k (k=2 to 10) features for Xsum (Fig. \ref{fig:RALine_plot_CNN}) and CNN/DM (Fig. \ref{fig:RAline_plot_xsum}) datasets. Rank agreement scores decrease as the value of k increases. For both the datasets the attention vs. DeepLIFT method pair exhibits the highest rank agreement with relatively low value. Rank agreement values for CNN/DM  dataset are low in magnitude compared to Xsum rank agreement scores.}
    \label{fig:Rank_agreement_side_by_side}
\end{figure} 

\item \textbf{Semantic Alignment Score Analysis:} The SAS scores illustrated in Fig. \ref{fig:cas_heatmap_xsum} and \ref{fig:cas_heatmap_cnn} were averaged across all k values due to their consistency, with relatively same scores across all method pairs. The semantic alignment is higher in the CNN/DM (0.64–0.66) compared to XSum (0.38–0.46), indicating stronger agreement between XAI methods in CNN/DM. This may be attributed to its discourse structure and less abstractive summary. SAS analysis per k is added in Tables: \ref{tab:cas_kwise_xsum} and \ref{tab:cas_kwise_cnn}. 

\begin{figure}[ht]
\centering
\begin{subfigure}[b]{0.49\textwidth}
    \centering
    \includegraphics[width=\textwidth]{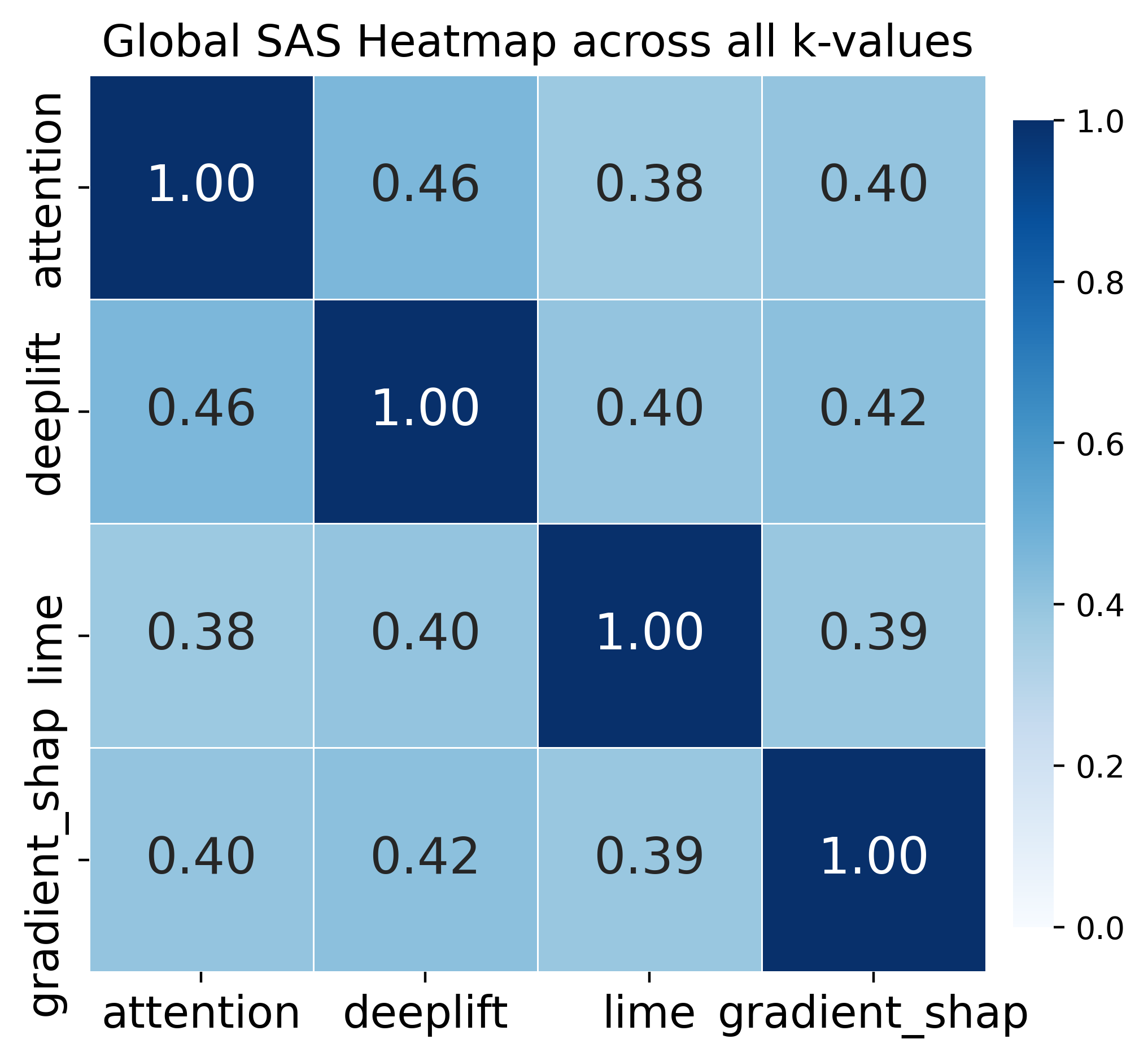}
    \caption{Average SAS across all k values for XSum Dataset}
    \label{fig:cas_heatmap_xsum}
\end{subfigure}
\hfill
\begin{subfigure}[b]{0.49\textwidth}
    \centering
    \includegraphics[width=\textwidth]{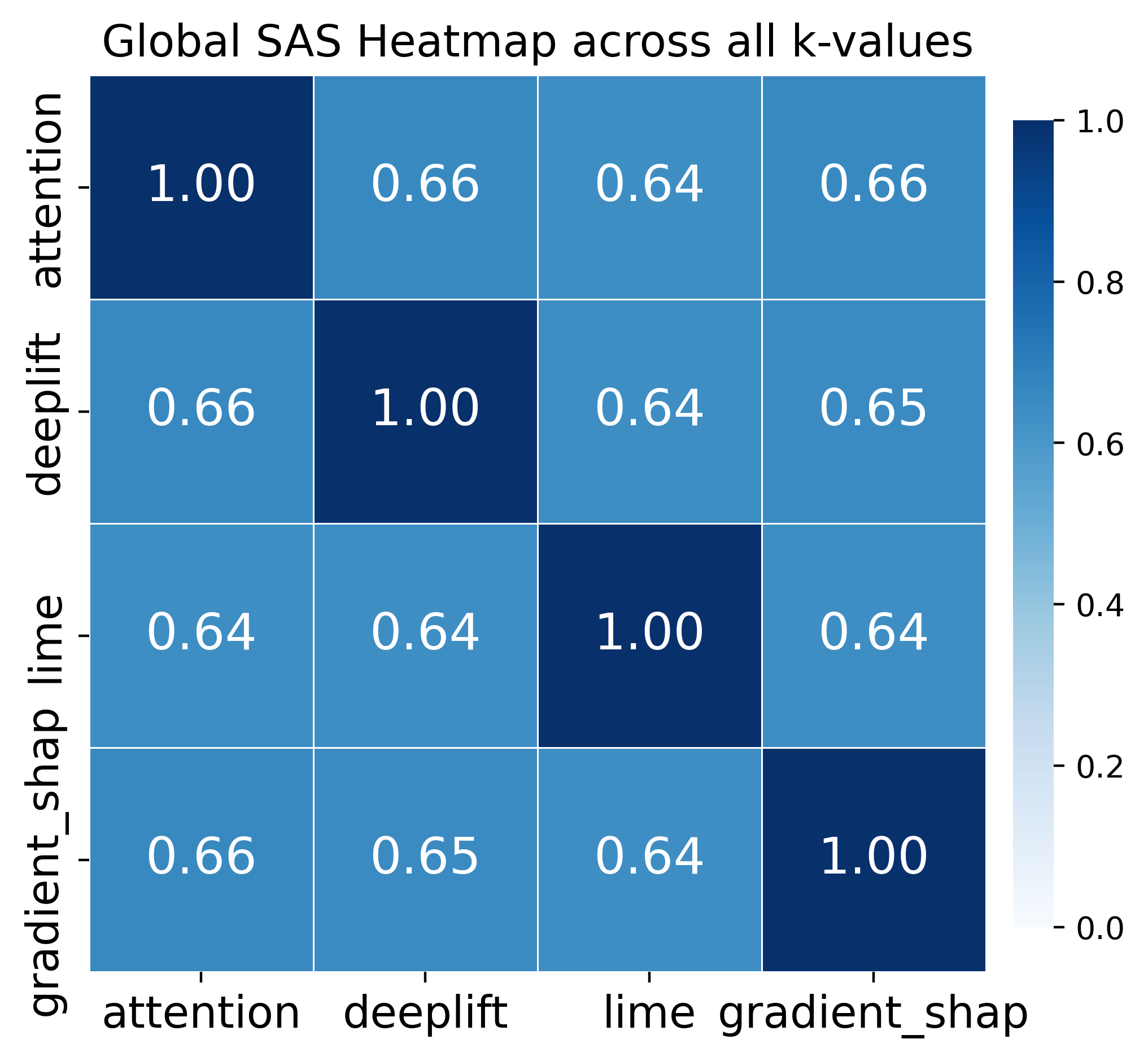}
    \caption{Average SAS across all k values for CNN/DM Dataset}
    \label{fig:cas_heatmap_cnn}
\end{subfigure}
\caption{Semantic Alignment heatmaps showing average SAS agreement between explanation methods on both datasets. Higher scores indicate stronger semantic alignment.}
\label{fig:cas_heatmap_sidebyside}
\end{figure}

\item \textbf{Rank Correlation Analysis:} 
We have also applied agreement metrics based on the relative ranking of features to quantify disagreement further. Fig. \ref{fig:SRA_heatmap_side_by_side}  presents the average rank correlation heatmaps. In the Xsum dataset, the attention vs. DeepLIFT method pair consistently exhibited high agreement, with a high average correlation value.  In contrast, for the CNN/DM dataset attention vs. Gradient SHAP has achieved the highest correlation value. However, the average correlation value is relatively low. Across both datasets, all other method pairs display significantly lower rank correlation scores, highlighting a strong disagreement in relative ranking of features among these methods. The results of pairwise rank agreement metric are discussed in appendix \ref{Seca4} in Fig. \ref{fig:PRA_heatmap_side_by_side}.

\begin{figure}[H]
    \centering
    \begin{subfigure}[b]{0.49\textwidth}
        \centering
        \includegraphics[width=\textwidth]{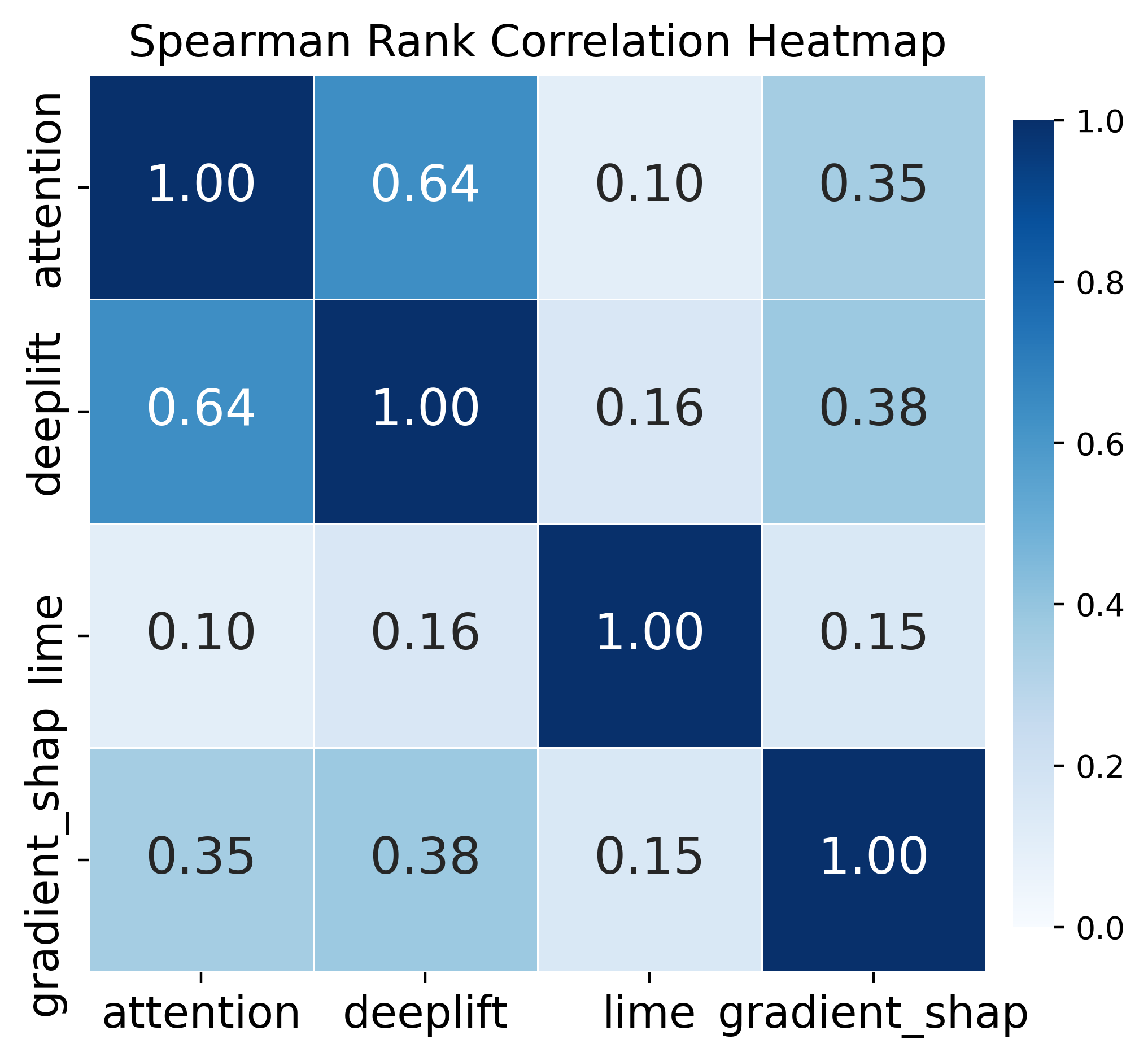}
        \caption{Rank correlation heatmap for Xsum.}
        \label{fig:SRA_heatmap_xsum}
    \end{subfigure}
    \hfill
    \begin{subfigure}[b]{0.49\textwidth}
        \centering
        \includegraphics[width=\textwidth]{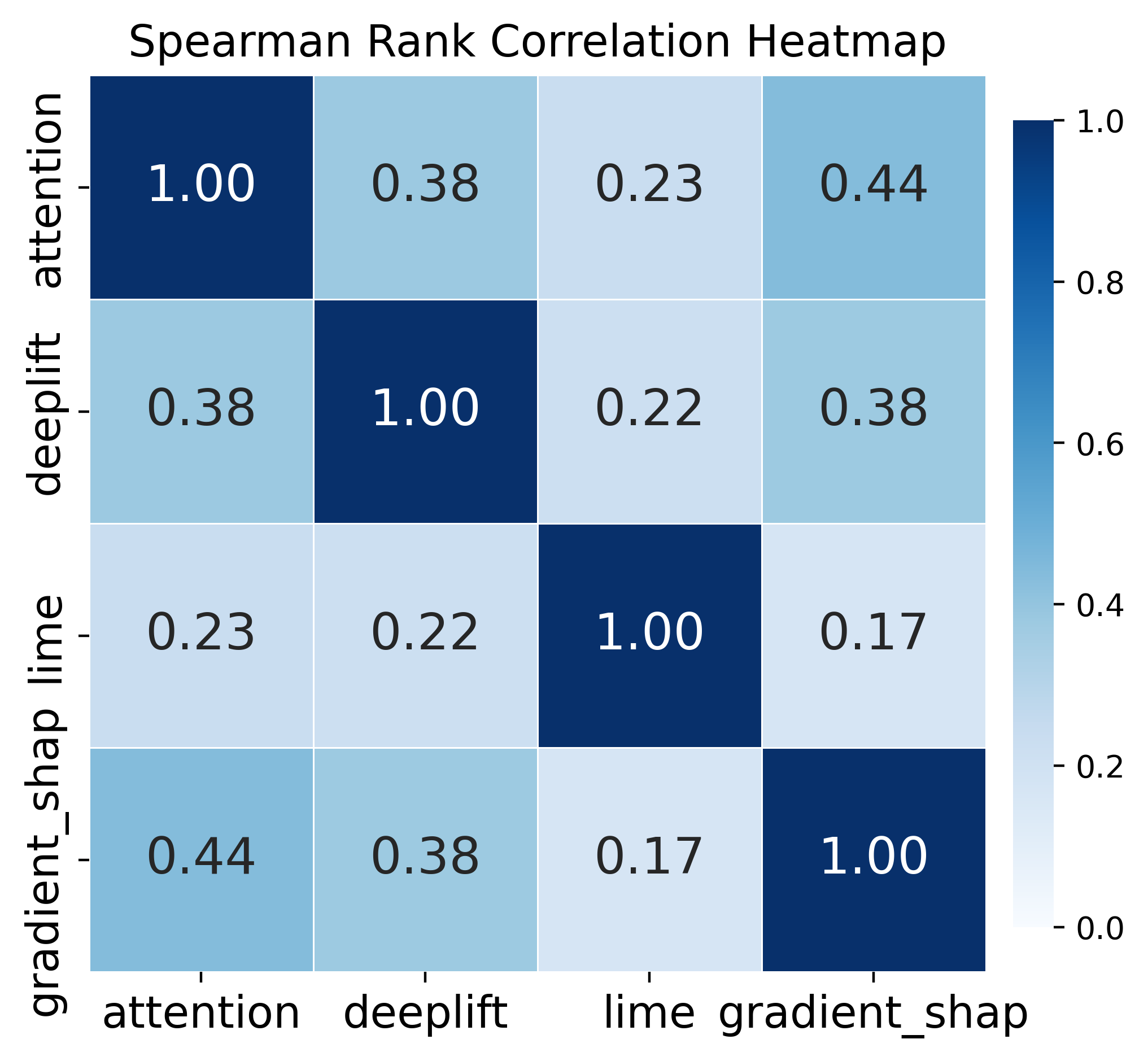}
        \caption{Rank correlation heatmap for CNN/DM.}
        \label{fig:SRA_heatmap_CNN}
    \end{subfigure}
    \caption{Average spearman rank correlation heatmap for Xsum (Fig. \ref{fig:SRA_heatmap_xsum}) and CNN/DM (Fig. \ref{fig:SRA_heatmap_CNN}) datasets. Overall agreement scores based on rank correlation are low, indicating significant disagreement between XAI methods for both Xsum and CNN/DM datasets.}
    \label{fig:SRA_heatmap_side_by_side}
\end{figure}

\subsubsection{\textbf{RQ2.}How do dataset characteristics such as the input article length and summary type affect the disagreement among XAI methods in text summarization?} \label{RXAI_results}
Based on the empirical analysis presented in section \ref{global_analysis}, the global agreement scores reveal significant disagreement among XAI methods across both datasets, Xsum and CNN/DM. The diversity of document length and summary type between datasets impacts the disagreement scores. The average length of the documents in XSum dataset is shorter than CNN/DM as discussed in Table: \ref{tab:dataset_statistics}. The Xsum highlights a single-sentence summary whereas CNN/DM consists multi-sentence summary. As a result, the disagreement scores of CNN/DM are lower than those in XSum. This highlights the challenges of handling longer lengths of input articles and summaries. However this trend is reversed for semantic alignment (SAS) where CNN/DM exhibits higher semantic alignment compared to Xsum, likely due to its structure and less abstractive summary. Despite the differences in the magnitude of disagreement scores, the trend of disagreement remains consistent across both datasets. This indicates that the disagreement observed has not occurred randomly and the dataset characteristics play a key role in the disagreement analysis.

\end{enumerate}
\fbox{
    \parbox{\textwidth}{\textbf{Key takeaways:} From our empirical analysis, it is evident that the XAI methods disagree significantly when applied to text summarization. The attention vs. DeepLIFT method pair frequently shows the highest agreement, indicating consistent explanations in contrast, LIME tends to diverge significantly from other methods and display consistently lower agreement scores. The diversity of documents in terms of length and summary types affects the overall magnitude of disagreement scores, resulting in higher magnitude of agreement scores for Xsum dataset with comparatively smaller documents than CNN/DM dataset, whereas the Semantic Alignment Scores exhibit opposite trend, with CNN/DM dataset having higher alignment scores.
}
}

\subsection{\textbf{RQ3. Does the proposed RXAI framework, through text segmentation, reduce disagreement between XAI methods in text summarization? }}
\begin{enumerate}
    \item \textbf{Feature Agreement Analysis:} The feature agreement analysis of top-k features for segmented articles is shown in Fig. \ref{fig:FA_regional_sidebyside}. The attention vs. DeepLIFT pair in XSum dataset (Fig. \ref{fig:FA_regional_xsum}) obtains the highest feature agreement score of roughly 0.85 for \texttt{k=4}, suggesting that it is highly aligned following text segmentation. In the CNN/DM dataset (Fig: \ref{fig:FA_regional_CNNDM}), method pairs that include Gradient SHAP demonstrate the most significant enhancement, with attention vs Gradient SHAP and DeepLIFT vs. Gradient SHAP attaining highest feature agreement scores of 0.67 and 0.66, respectively for \texttt{k=4}. These outcomes demonstrate the effectiveness of RXAI framework. In both datasets. LIME-based method pairs continue to exhibit the lowest agreement. However, agreement scores of LIME-based method pairs have significantly improved. 

To compare and verify the effectiveness of RXAI framework appendix Figures: \ref{fig:global_vs_regional_fa_xsum} of Xsum and \ref{fig:global_vs_regional_fa_cnndm} of CNN/DM dataset present global vs. regional feature agreement heatmaps at $k$=4. The heatmaps of both datasets highlight significant improvement in feature agreement scores compared to global feature agreement. To test the significant mean difference of agreement scores between global and segmented articles, we have used Wilcoxon signed rank test for all method pairs across different k-values. This non-parametric test was chosen because it does not assume normality and is well-suited for paired comparison between global and regional scores. For both datasets the $p$-values were below 0.05, highlighting significant mean difference. Appendix Tables \ref{tab:wilcoxon_xsum_fa} and \ref{tab:wilcoxon_batch2_cnn} report the $p$-values across different $k$-values and method pairs, based on the same representative batch of 500 articles used in the main analysis.

\begin{figure}[H]
    \centering
    \begin{subfigure}[b]{0.49\textwidth}
        \centering
        \includegraphics[width=\textwidth]{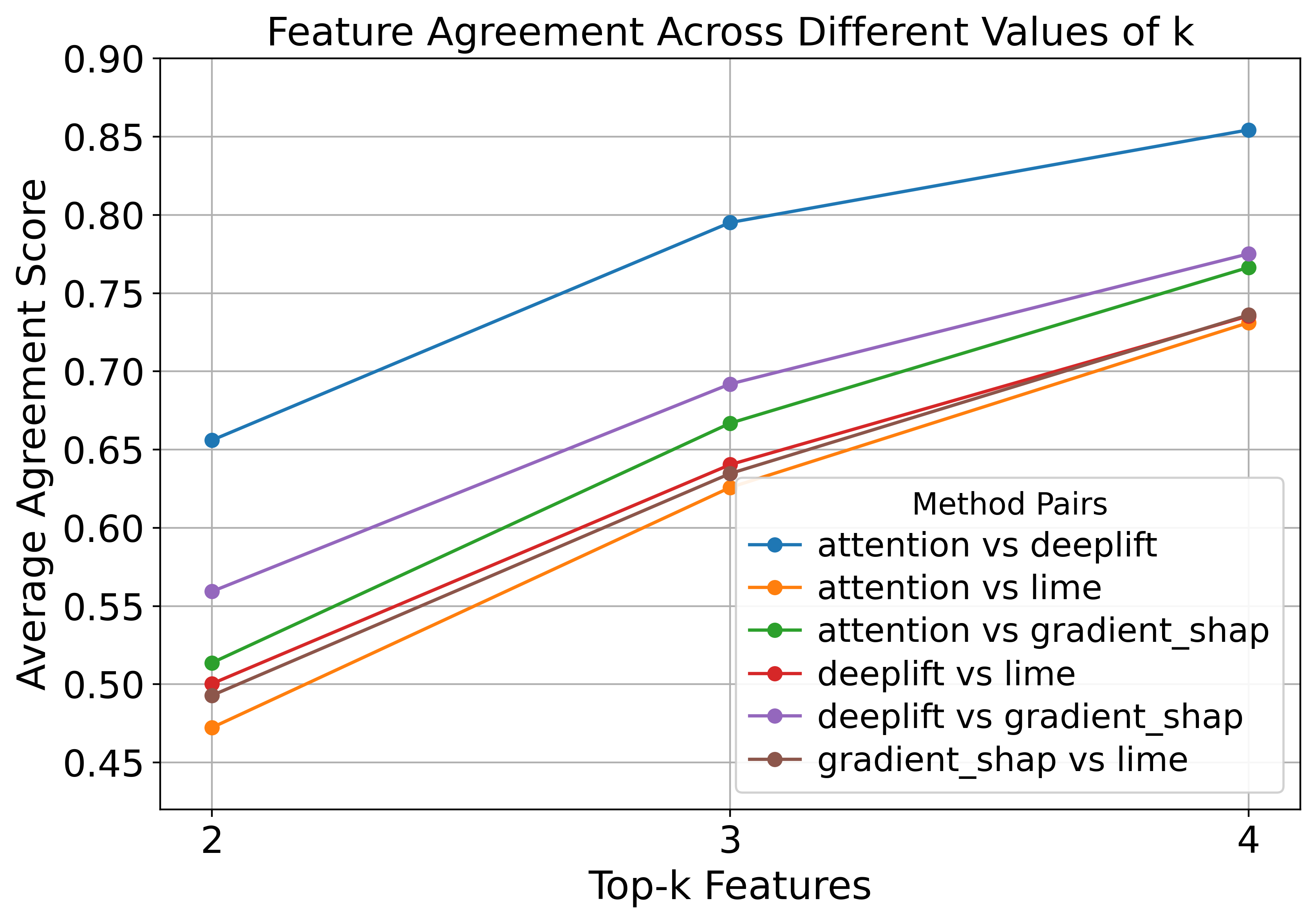}
        \caption{Feature Agreement across top-k values (Segmented Articles) for Xsum dataset.}
        \label{fig:FA_regional_xsum}
    \end{subfigure}
    \hfill
    \begin{subfigure}[b]{0.49\textwidth}
        \centering
        \includegraphics[width=\textwidth]{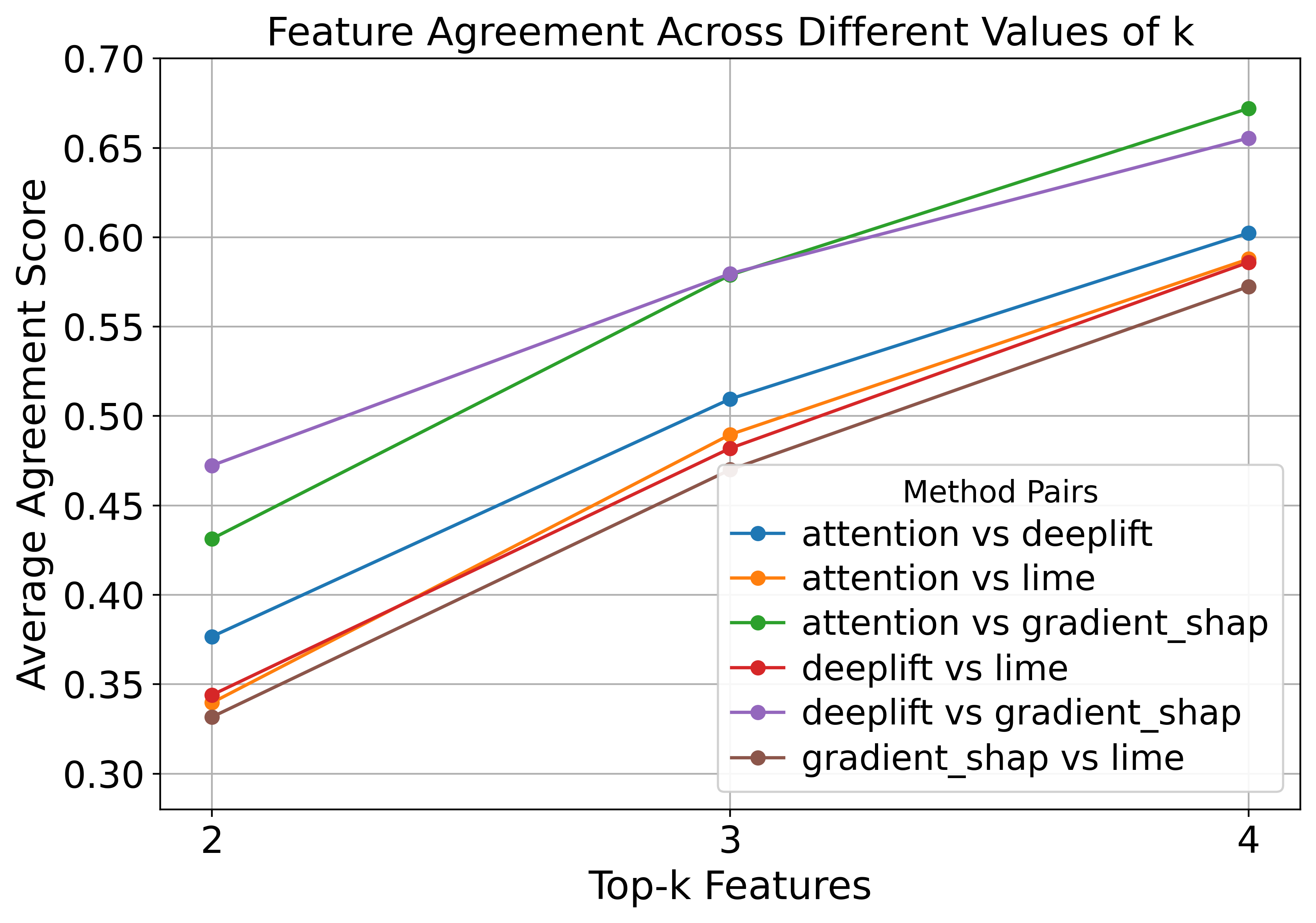}
        \caption{Feature Agreement across top-k values (Segmented Articles) for CNN/DM dataset.}
        \label{fig:FA_regional_CNNDM}
    \end{subfigure}
    \caption{Average feature agreement scores for segmented articles of Xsum (Fig.~\ref{fig:FA_regional_xsum}) and CNN/DM (Fig.~\ref{fig:FA_regional_CNNDM}). The line plot highlights significant improvements in agreement, with the attention vs. DeepLIFT pair achieving the highest score for Xsum and attention vs. Gradient SHAP for CNN/DM.}
    \label{fig:FA_regional_sidebyside}
\end{figure}
 
\item \textbf{Rank Agreement Analysis:} The regional rank agreement plot of the Xsum dataset in Fig. \ref{fig:RA_regional_xsum} shows increased and relatively stable rank agreement scores across the different values of k, with DeepLIFT vs. Gradient SHAP achieving the highest rank agreement of approximately 0.32 for k = 4 compared to the global score of 0.13, indicating improved agreement scores after segmentation.  
The LIME-based method pairs also achieve improved rank agreement scores after segmentation with DeepLIFT vs. LIME achieving the highest rank agreement of 0.28 at \texttt{k} = 4 compared to a global agreement score of 0.11.
The rank agreement plot of the CNN/DM shown in Fig. \ref{fig:RA_regional_CNNDM} highlights a similar trend as Xsum dataset with attention vs. Gradient SHAP and DeepLIFT vs. Gradient SHAP having the highest agreement. The segmentation approach does improve the rank agreement scores. 

The Analysis of Rank correlation and SAS revealed that the agreement scores based on rank correlation and SAS scores remain largely consistent as the global agreement scores. The graphs are added in the Appendix \ref{Regional_results}.

    \begin{figure}[H]
    \centering
    \begin{subfigure}[b]{0.49\textwidth}
        \centering
        \includegraphics[width=\textwidth]{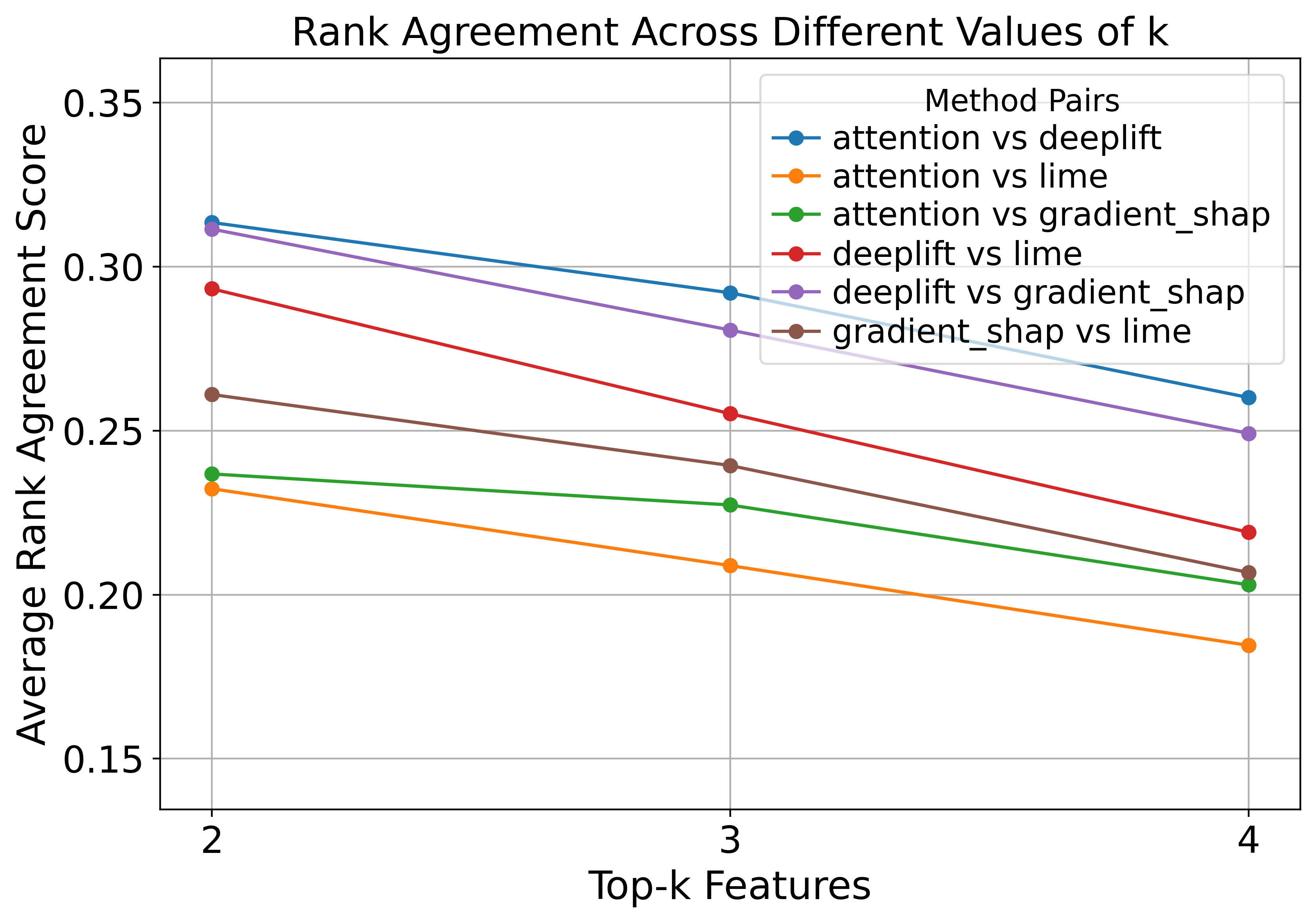}
        \caption{Rank Agreement across top-k values (Segmented Articles) for Xsum dataset.}
        \label{fig:RA_regional_xsum}
    \end{subfigure}
    \hfill
    \begin{subfigure}[b]{0.49\textwidth}
        \centering
        \includegraphics[width=\textwidth]{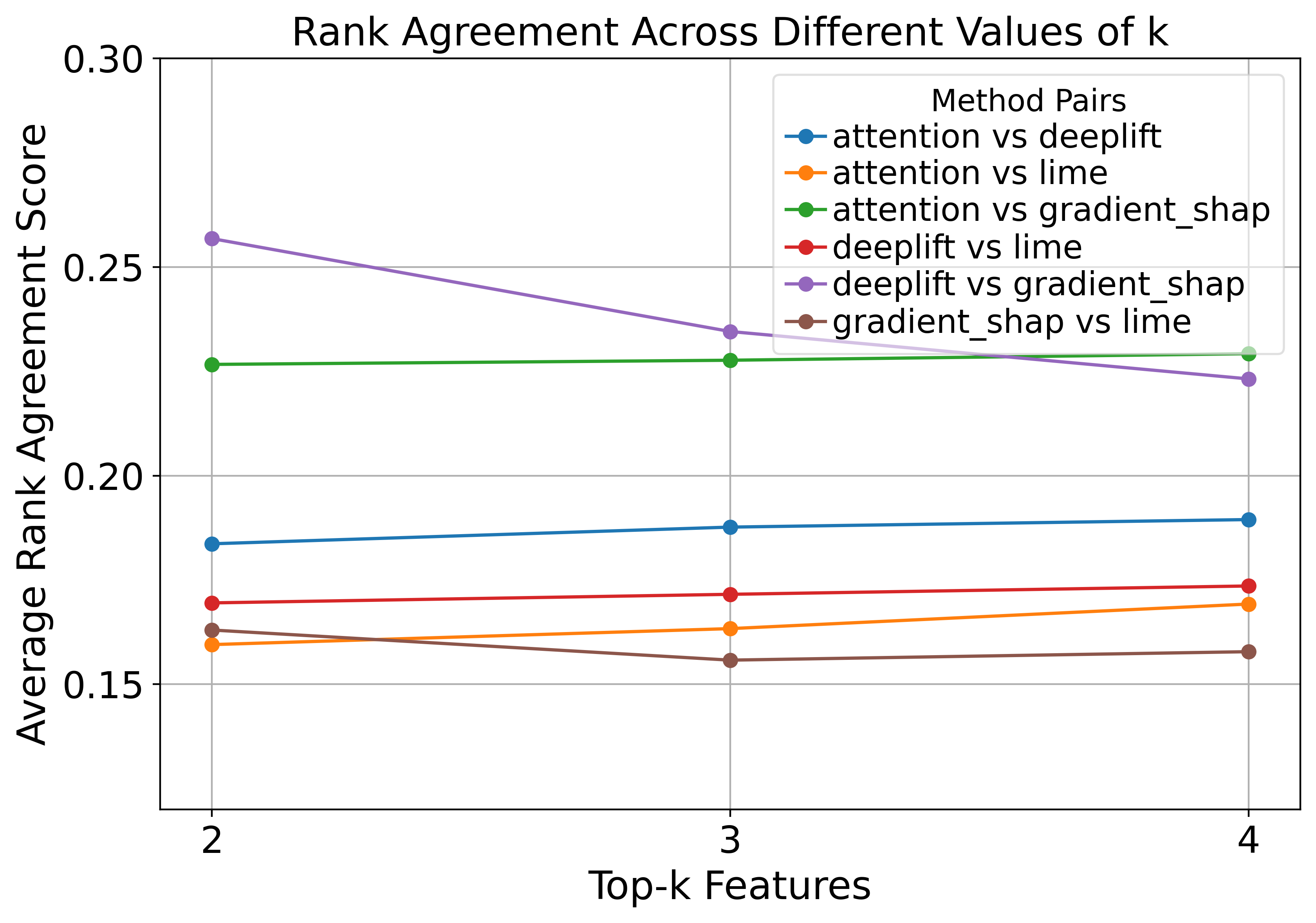}
        \caption{Rank Agreement across top-k values (Segmented Articles) for CNN/DM.}
        \label{fig:RA_regional_CNNDM}
    \end{subfigure}
    \caption{Average rank agreement scores across top-k values for segmented articles of Xsum (Fig. \ref{fig:RA_regional_xsum}) and CNN/DM dataset (Fig. \ref{fig:RA_regional_CNNDM}). The plots demonstrate overall improvement in rank agreement scores for both datasets text segmentation.}
    \label{fig:RA_regional_sidebyside}
\end{figure}


\noindent\fbox{%
  \begin{minipage}{0.95\textwidth}
    \textbf{Key takeaways:} The results of RXAI based on top-k features on both datasets highlight that while the overall agreement improves between XAI methods, the extent of this improvement varies between different agreement metrics. Feature agreement and rank agreement are significantly enhanced, but rank agreement metrics reveal a continued disagreement. SAS scores remain relatively unchanged after applying RXAI approach. On the contrary, relative ranking-based approaches do not exhibit improved agreement scores on both datasets. Overall, the results based on syntactic top-k feature metrics highlight that regional explanations increase the agreement level with statistically significant differences in mean scores, validating the hypothesis that text segmentation improves the consistency of explanations across methods. 
  \end{minipage}%
}

\end{enumerate}

\section{Discussion}
\subsection{Disagreement Problem among XAI methods}

 Our results confirm significant disagreement among XAI methods in text summarization. Model-specific methods, such as attention vs. DeepLIFT consistently demonstrate higher agreement, reflecting their focus on global attributions, while model-agnostic method like LIME shows lower agreement due to their perturbation-based approach.

The disagreement problem is critical in high-stake and secure AI applications. In healthcare, inconsistency of explanations may lead to conflicting diagnoses. In legal and financial domains, conflicting explanations can mislead contractual interpretations or risk assessments. One potential future direction would be to use Large Language Models (LLMs) as mediators to aggregate explanations of different XAI methods into more consistent explanations. Additionally, conversational AI agents could enhance interpretability by enabling interactive investigation of attribution discrepancies, helping users understand why disagreements occur.

\subsection{Impact of RXAI framework}
Silhouette-based \texttt{k}-means segmentation mitigates the disagreement problem. Our findings demonstrate that segmentation serves as a promising strategy to mitigate this problem, while also revealing that the impact of segmentation varies across different agreement metrics. The proposed approach of utilizing silhouette-based clustering in high-dimensional context has known limitations (as discussed in \ref{limitations}). Future work should explore alternate clustering validation indices or adapt dimensionality reduction techniques to address this limitation. Applying RXAI framework introduces computational overhead. The segmented articles of Xsum dataset requires 39 hours of computation compared to 29 hours of global analysis (34\% increase). For CNN/DM, it requires 78.6 hours of computation compared to 53.9 hours of global computation (45.7\% increase) for a 500 article batch. In case of LIME, which is particularly compute-intensive due to the lack of batched attribution support in the Inseq library, attribution generation for segmented data of 500 articles based on LIME required 29 hours for XSum and approximately 65 hours for CNN/DM—accounting for over 74\% and 80\% of the total execution time, respectively. The computational overhead stems from segmentation, XAI method execution (with LIME being primary contributor) on segments, and the number of articles to process. This results in O(n x s x m) complexity where s, n, and m represent segment, method, and number of articles. This computational cost represents a reasonable trade-off for applications requiring highly consistent explanations, though it limits
its deployment in time-sensitive scenarios. Future work should explore efficient segmentation approaches and optimization techniques.

While RXAI effectively enhances agreement between explanations it may introduce semantic or coherence drift between input article and segments, leading to diverging segments that may not fully preserve the original context. To evaluate this we measured semantic similarity and coherence between input articles and segments. The results illustrate high semantic similarity more than 90\% between input article and segments for both datasets (refer Table \ref{tab:semantic_similarity_all_batches}), ensuring semantic preservation. Additionally, the coherence analysis reveals minor drift between article and its segmented version, suggesting that overall coherence is largely maintained with moderate drift (refer Table \ref{tab:coherence_drift_split}). RXAI can be extended beyond news summarization with its careful adoption of domain-specific requirements and structured segmentation strategies in fields such as healthcare, legal, and financial text summarization. Future work could explore topic-aware segmentation and graph-based clustering to enhance scalability while preserving meaning and coherence when applying the RXAI approach.

\subsection{Limitations} \label{limitations}
\begin{enumerate}
\item This study demonstrates the existence of XAI disagreement in text summarization. However, it does not investigate the root cause of why these methods disagree. Future research could explore the root cause of disagreement among XAI methods.

\item Although silhouette score is widely used metric for identifying optimal number of clusters, its reliability in high-dimensional embedding spaces is affected by the curse of dimensionality. This may weaken the distinction between intra-cluster and inter-cluster distances, making it challenging to detect true cluster boundaries.

\item In this study, we have utilized English-language news summarization datasets as a foundational evaluation of RXAI in general news context. However, we acknowledge that adapting RXAI framework in summarization tasks such as legal, clinical or scientific text summarization may pose challenges due to longer input sequences, specialized terminologies and domain-specific structural complexity. 

\item We have tested explanation disagreement and RXAI framework on two encoder-decoder transformer model, BART and PEGASUS. While these models are strong baselines, they share similar architecture. The effectiveness of RXAI framework to other model types such as LED, T5, or decoder-only models like GPT and LLaMA, is untested and calls for future work.

\end{enumerate}

\subsection{Future Work}
For future work, we will investigate the performance of XAI methods in various text summarization domains such as health, finance, and legal. We will study the impact of domain-specific training on global and regional agreement of XAI methods. Future work could explore domain-adaptive segmentation strategies, where RXAI could be guided by domain-specific discourse markers, ontologies, or structural cues. Comparing agreement scores between models trained on domain-specific datasets and those trained on general-purpose datasets will allow researchers to determine whether specialized training affects the amount of agreement among XAI methods. Additionally, we will explore optimization strategies to improve the scalability of the RXAI approach on larger datasets. Future work should also assess a wide range of summarization models to evaluate whether the disagreement trend and RXAI effectiveness generalize across different model variants. Such evaluation will help evaluate architecture-dependent variations and ensure the effectiveness of RXAI in diverse paradigms. Finally, to assess the effectiveness of the proposed visualization tool
in enhancing interpretability and user trust, future iterations could involve human-centered evaluations.
\section{Conclusion}
Our work generates empirical evidence of the existence of the disagreement problem in XAI for text summarization. The results indicate that, on both datasets: Xsum and CNN/DM, attention and DeepLIFT reflect higher agreement consistently, while LIME shows lower agreement with other methods because of the inherently different explanation generation approach. To mitigate this, we propose RXAI, a novel segmentation-based approach. The proposed framework effectively enhances the agreement scores, especially feature agreement, by splitting articles into semantically coherent segments. The effectiveness of the RXAI approach in generating stable explanations can enhance trust in AI-driven decisions, specifically for security-sensitive applications. This paper contributes toward state-of-the-art improvement in the consistency and interpretability of Explainable AI for text summarization and also lays the base for further research in this direction.


\begin{appendices}
\section{Text Summarization Dataset} \label{seca2}
This study utilizes two benchmark datasets, XSum and CNN/Daily Mail, for evaluating text summarization models and their explanations. The XSum dataset comprises $\sim$204K training articles and $\sim$11K test articles, with one-sentence abstractive summaries from BBC News. CNN/Daily Mail includes $\sim$287K training articles and $\sim$11K test articles with multi-sentence extractive summaries. Random samples of 500 test articles were selected from each dataset for analysis. Table: \ref{tab:dataset_statistics} provides a comprehensive breakdown of the average statistics for both datasets.

\begin{table} [ht]
\centering
\caption{Comparison of Average Statistics between CNN/Daily Mail and XSum Datasets.}
\label{tab:dataset_statistics}
\begin{tabular}{p{4cm}p{3cm}p{3cm}} 
\hline
\textbf{Statistic}            & \textbf{CNN/Daily Mail} & \textbf{XSum} \\ 
\hline
Average Document Length       & $\sim$760 words         & $\sim$431 words \\ 
Average Summary Length        & $\sim$56 words          & $\sim$23 words \\ 
Type of Summaries             & Extractive (highlight-based summaries) & Abstractive (one-sentence summaries) \\ 
Domain                        & News articles (CNN and Daily Mail)     & BBC News (various topics) \\ 
\hline
\end{tabular}
\end{table}

\section{XAI methods}\label{secA1}
\begin{enumerate}
    \item \textbf{ LIME (Local Interpretable Model-agnostic Explanations)}
    
LIME locally approximates a complex model using an interpretable model to explain individual predictions \citep{Ribeiroae}. 
LIME perturbs the input samples, observes changes, and derives an interpretable model on the sampled perturbations. In the text summarization task, LIME tries to explain the importance of a particular word or sentence by removing some of them from the input text. It also observes the effects of such changes on the generated summary. The interpretable model describes the local behavior of the original model around a given prediction. To generate the attribution, LIME minimizes the objective function stated below :

\begin{equation}
  \xi(x) = \arg\min_{g \in G} \mathcal{L}(f, g, \pi_x) + \Omega(g)  
\end{equation}
where:
\begin{itemize}
    \item $\xi(x)$: Explanation of the input $x$.
    \item $g \in G$: A selected model $g$ from the class of interpretable models.
    \item $\mathcal{L}(f, g, \pi_x)$: The fidelity loss function
    \item $\pi_x$: The proximity measure
    \item $\Omega(g)$: A complexity measure of the interpretable model $g$. 
    \item $f$: The original complex model being explained, which maps the input space $x \in \mathbb{R}^d$ to a real number.
\end{itemize}

\item \textbf{Gradient SHAP (SHapley Additive exPlanations using gradients)}

Gradient SHAP provides an estimation for the accurate and efficient feature attribution, combining the SHAP values \citep{Lundbergea} and Integrated Gradients \citep{Sundararajanea}. Gradient SHAP can be applied in text summarization to assign the importance of every word or token by calculating the gradient with respect to the output for every input feature that resulted in a given summary. This method effectively points out which part of the input text contributes much to the resulting summary. It uses the following formula for the estimation of an attribution score:

\begin{equation}
  \text{Attribution} = \mathbb{E}_{\text{baselines}} \left[ \text{gradients} \times (\text{inputs} - \text{baselines}) \right]
\end{equation}
This formula captures the expected contribution of input features to the model's prediction by averaging the gradients over multiple baselines. 

\item \textbf{attention}

The attention mechanism introduced by \citep{bahdanau2014neural} is extensively applied in transformer models to concentrate on pertinent segments of the input text during output generation. The attention mechanism computes a weighted sum of the input features, where weights are calculated depending on the relevance of each feature to the output produced at a given step. Self-attention \citep{vaswanietal} improves this idea by allowing the model to evaluate the relevance of certain tokens over the whole input sequence. In text summarization, the attention mechanism helps the model to capture the most crucial information by allowing it to selectively focus on different parts of the input text throughout the generation of each summary word.

\begin{equation}
\text{Attention}(Q, K, V) = \text{softmax}\left(\frac{QK^T}{\sqrt{d_k}}\right)V
\end{equation}
where $Q$ (query), $K$ (key), and $V$ (value) are matrices derived from the input, and $d_k$ is the dimension of the key vectors.

\item \textbf{DeepLIFT (Deep Learning Important FeaTures)}

\citep{shrikumarea} proposed DeepLIFT, a method for assigning the prediction score of a deep learning model to its input features. The DeepLIFT method compares every neuron's activation to a reference activation. The method assigns the meaning of each word or phrase in the input text by analyzing how changes in the input affect the output summary, to provide insights into the decision-making process of the text summarization. DeepLIFT employs a ”summation-to-delta” principle that is pointed out in Equation 4 below.

\begin{equation}
\sum_{i=1}^{n} C_{\Delta x_i \Delta o} = \Delta o,
\label{eq:Deeplift}
\end{equation}
where ( $o$ = $f(x)$ ) is the model output, ( $\Delta$ $o$ = $f(x)$ - $f(r)$  ), ( $\Delta$ $x_i$ = $x_i$ - $r_i$ ), and $r$  is the reference input.

\end{enumerate}

\section{Disagreement Metrics}\label{seca3}
To quantify the degree of disagreement among various XAI methods, multiple metrics that assess various aspects of feature importance and ranking were utilized after the explanations were generated.

\subsubsection{ Agreement Measurement based on Top-k features} In this category of measurement following two metrics were employed to quantify the degree of agreement between two pairs of explanations based on their most important (top- \texttt{k}) features. The value of \texttt{k}  initiates at 2, denoting the top-2 features, and may be extended to any \texttt{k} value, dependent upon the length of the articles under analysis.

\begin{enumerate}

\item{ \textbf{Semantic Alignment Score (SAS)}}  

We propose SAS, a semantic similarity-based metric to evaluate semantic alignment between different XAI methods. Unlike previously suggested measures that heavily rely on syntactic similarity, SAS employs sentence embedding to capture semantic relationships across explanations. It provides a unique perspective for measuring disagreement between XAI methods for text data. The steps to calculate SAS are listed below:

\begin{itemize}
    \item Selection of Top-k Features: The top-k sentences for each XAI method are selected based on their attribution score.
    \item To convert the sentences into a dense vector space the sentence embeddings are generated for the selected top-k sentences.
    \item Weighting by Attribution Scores: The sentence embeddings are multiplied with their attribution score to emphasize their contribution
    \item Cosine Similarity Computation: Pairwise cosine similarity is computed between the weighted sentence embeddings.
    \item Averaging for Agreement: To derive the global SAS value across all method pairs, the average similarity across all top-k features is calculated. \\The SAS formula is defined as:
\end{itemize}

\[
\text{SAS}(E_a, E_b, k) = \frac{1}{k} \sum_{i=1}^{k} \text{cos\_sim}(S_{E_a,i} \cdot \text{Emb}_{E_a,i}, S_{E_b,i} \cdot \text{Emb}_{E_b,i})
\]

Where, \(E_a, E_b\) are Explanations from two methods, \(S_{E_a,i}, S_{E_b,i}\) are Attribution scores for the \(i\)-th top feature,
\(\text{Emb}_{E_a,i}, \text{Emb}_{E_b,i}\) are Sentence embeddings for the \(i\)-th top feature and \(k\) is Number of top features.

\item{\textbf{ Feature Agreement}}

Feature agreement was the first metric applied in this study, and it referred to the degree to which two explanations agree on their most important features. It calculates the ratio of the number of common features between the two top-k feature sets from both explanations.

Mathematically, it is defined as:

\[
\text{Feature Agreement}(E_a, E_b, k) = \frac{| \text{TF}(E_a, k) \cap \text{TF}(E_b, k) |}{k}
\]

where \(\text{TF}(E, k)\) returns the set of top-k features of explanation \(E\) based on the magnitude of feature importance values. Ea and Eb are explanations of two methods a and b.\\

 \item{ \textbf{ Rank Agreement}}

The rank agreement metric compares the sets of \texttt{top-k} features, including their relative ordering in addition to their intersection of common features. It offers a more rigorous measure than simple feature agreement because it includes the positional ordering of features, which can greatly influence the interpretation of model explanations. The metric computes the fraction of features that are both present in the top-k feature sets of both explanations and share the same rank inside those sets, considering two explanations $E_a$ and $E_b$.  
It can be computed using the formula below: $\text{Rank Agreement}(E_a, E_b, k)$ =
\end{enumerate}

\begin{equation}
\frac{ \left| \left\{ s \mid s \in T_a \cap T_b \land \text{rank}(E_a, s) = \text{rank}(E_b, s) \right\} \right| }{k}
\end{equation}
Here, \\
$T_a = \text{top\_features}(E_a, k)$ and $T_b =  \text{top\_features}(E_b, k)$.\\ \\
$S$ represents the entire set of features, $\text{top\_features}(E, k)$ is the set of top-$k$ features in explanation $E$, and $\text{rank}(E, s)$ returns the rank or position of the feature $s$ within the explanation $E$.

\subsubsection{Agreement Measurement based on Relative ordering of features} The following two measures rely on the relative feature ordering. Two explanations are considered to be different if the relative ordering of the features of interest varies between them.

\begin{enumerate}

\item{\textbf{Rank Correlation}}

To quantify the agreement between the relative ordering of features in two explanations, $E_a$ and $E_b$, Spearman's rank correlation coefficient is used. This metric quantifies the degree of correlation between the rankings of a selected set of features. The rank correlation between two explanations $\text{Rank Correlation}(E_a, E_b, F)$ can be computed as:

\begin{equation}
 r_s(\text{Ranking}(E_a, F), \text{Ranking}(E_b, F))
\end{equation}

Where $F$ is a selected set of features, $r_s$ represents Spearman rank correlation coefficient, and $\text{Ranking}(E, F)$ is the rank assigned to the features in $F$ based on explanation $E$. Lower correlation values indicate a higher level of disagreement in the ordering of features.

\item{\textbf{Pairwise Rank Agreement (PRA)}}

It quantifies the agreement of the relative ordering of feature pairs across two explanations. This metric shows whether the relative importance of every pair of features in one explanation matches with the same pair in another. Specifically, it evaluates the proportion of feature pairs for which the relative ranking stays constant across both models. The formula for Pairwise Rank Agreement is given below:

\begin{equation}
\text{PRA}(E_a, E_b, F) = \frac{ \sum_{i < j} \mathbf{1}[\text{Rank}_a(f_i, f_j) = \text{Rank}_b(f_i, f_j)] }{\binom{|F|}{2}}
\end{equation}

Where $F = \{f_1, f_2, \ldots\}$ is the set of selected features, and $\mathbf{1}$ is an indicator function that returns 1 if the relative ranking of $f_i$ and $f_j$ is the same in both explanations and 0 otherwise. The denominator $\binom{|F|}{2}$ represents the total number of feature pairs.  $\text{Rank}_a(f_i, f_j)$ represents the relative ranking of features $f_i$ and $f_j$ in explanation $E_a$. $\text{Rank}_b(f_i, f_j)$ represents the relative ranking of the same feature pair in explanation $E_b$.\\ \\

\end{enumerate}

\section{Results of Disagreement Analysis} \label{Seca4}
\subsection{Disagreement analysis on Full-text articles (Same Representative Batch used in Section:\ref{Disgreement_results})}

\begin{itemize}

\item{\textbf{Semantic Alignment Score}} To measure the agreement between explanation
methods based on semantic alignment, we have used the SAS metric. The 
Analysis of Xsum and CNN dataset across different top-k values (\texttt{k} = 2 to 8) is depicted in Table: \ref{tab:cas_kwise_xsum} and Table: \ref{tab:cas_kwise_cnn}, highlighting that the semantic agreement scores across different k-values remain largely consistent. Since the SAS values are consistent across different k-values, the top-k is selected from 2 to 8. As depicted in tables \ref{tab:cas_kwise_xsum} and \ref{tab:cas_kwise_cnn}, the semantic agreement scores for CNN/DM are higher than the scores of Xsum, diverging from the pattern observed in syntactic agreement metrics, where Xsum typically scored higher. This suggests that the higher agreement in CNN/DM may stem from its distinctive discourse structure when compared to Xsum.

    \begin{table}[ht]
    \caption{Average SAS scores across $k$ values (2 to 8) for each XAI method pair on the XSum dataset batch presented in section \ref{Disgreement_results}.}
    \label{tab:cas_kwise_xsum}
    \begin{tabular*}{\textwidth}{@{\extracolsep\fill}llrrrrrrr}
    \toprule
    \textbf{Method 1} & \textbf{Method 2} & \textbf{k=2} & \textbf{3} & \textbf{4} & \textbf{5} & \textbf{6} & \textbf{7} & \textbf{8} \\
    \midrule
    attention & deeplift       & 0.480 & 0.475 & 0.470 & 0.465 & 0.460 & 0.456 & 0.449 \\
    attention & gradient\_shap & 0.407 & 0.408 & 0.404 & 0.404 & 0.401 & 0.399 & 0.398 \\
    attention & lime           & 0.380 & 0.383 & 0.384 & 0.385 & 0.382 & 0.382 & 0.382 \\
    deeplift  & gradient\_shap & 0.432 & 0.423 & 0.418 & 0.417 & 0.413 & 0.409 & 0.407 \\
    deeplift  & lime           & 0.403 & 0.399 & 0.397 & 0.396 & 0.394 & 0.391 & 0.390 \\
    lime      & gradient\_shap & 0.388 & 0.390 & 0.389 & 0.392 & 0.391 & 0.390 & 0.388 \\
    \bottomrule
    \end{tabular*}
    \end{table}

    \begin{table}[ht]
    \centering
    \caption{Average SAS scores across $k$ values (2 to 8) for different XAI method pairs in the CNN/DM dataset batch presented in section \ref{Disgreement_results}.}
    \label{tab:cas_kwise_cnn}
    \begin{tabular*}{\textwidth}{@{\extracolsep\fill}llrrrrrrr}
    \toprule
    \textbf{Method 1} & \textbf{Method 2} & \textbf{k=2} & \textbf{3} & \textbf{4} & \textbf{5} & \textbf{6} & \textbf{7} & \textbf{8} \\
    \midrule
    attention & DeepLIFT & 0.660 & 0.661 & 0.662 & 0.662 & 0.662 & 0.660 & 0.659 \\
    attention & Gradient\_SHAP & 0.657 & 0.657 & 0.654 & 0.657 & 0.658 & 0.657 & 0.658 \\
    attention & lime & 0.634 & 0.640 & 0.644 & 0.648 & 0.649 & 0.649 & 0.650 \\
    DeepLIFT & Gradient\_SHAP & 0.653 & 0.656 & 0.653 & 0.653 & 0.654 & 0.653 & 0.653 \\
    DeepLIFT & lime & 0.629 & 0.636 & 0.641 & 0.643 & 0.644 & 0.647 & 0.649 \\
    lime & Gradient\_SHAP & 0.632 & 0.636 & 0.641 & 0.642 & 0.644 & 0.643 & 0.645 \\
    \bottomrule
    \end{tabular*}
    \end{table}

    \item \textbf{Box Plot of Feature Agreement} The Fig. \ref{fig:FA_boxplot_side_by_side} illustrates the variability of feature agreement scores across highlighted top-\texttt{k} values (\texttt{k} = all even numbers between 2 to 10) for both datasets. The box plot of Xsum as shown in Fig. \ref{fig:box_plot_xsum} presents consistent feature agreement scores with low variability compared to CNN/DM dataset. The box plot of the CNN/DM dataset, as shown in Fig. \ref{fig:box_plot_CNN}, presents a high variability of feature agreement scores compared to Xsum, with LIME-based pairs having the highest variability across all other method pairs.

\begin{figure}[H]
    \centering
    \begin{subfigure}[b]{0.49\linewidth}
        \centering
        \includegraphics[width=\textwidth]{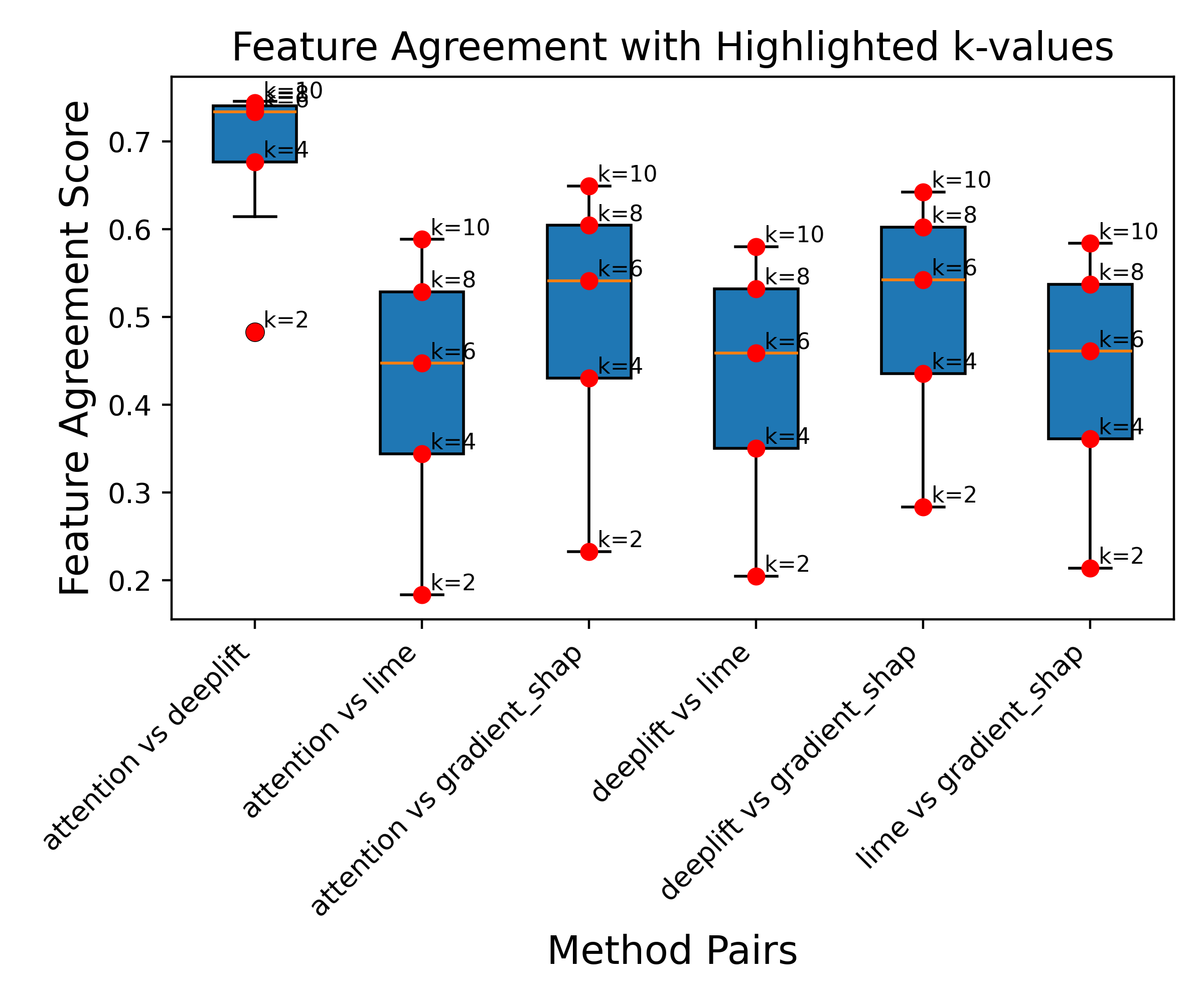}
        \caption{Box plot of feature agreement with highlighted k-values for Xsum.}
        \label{fig:box_plot_xsum}
    \end{subfigure}
    \hfill
    \begin{subfigure}[b]{0.49\linewidth}
        \centering
        \includegraphics[width=\textwidth]{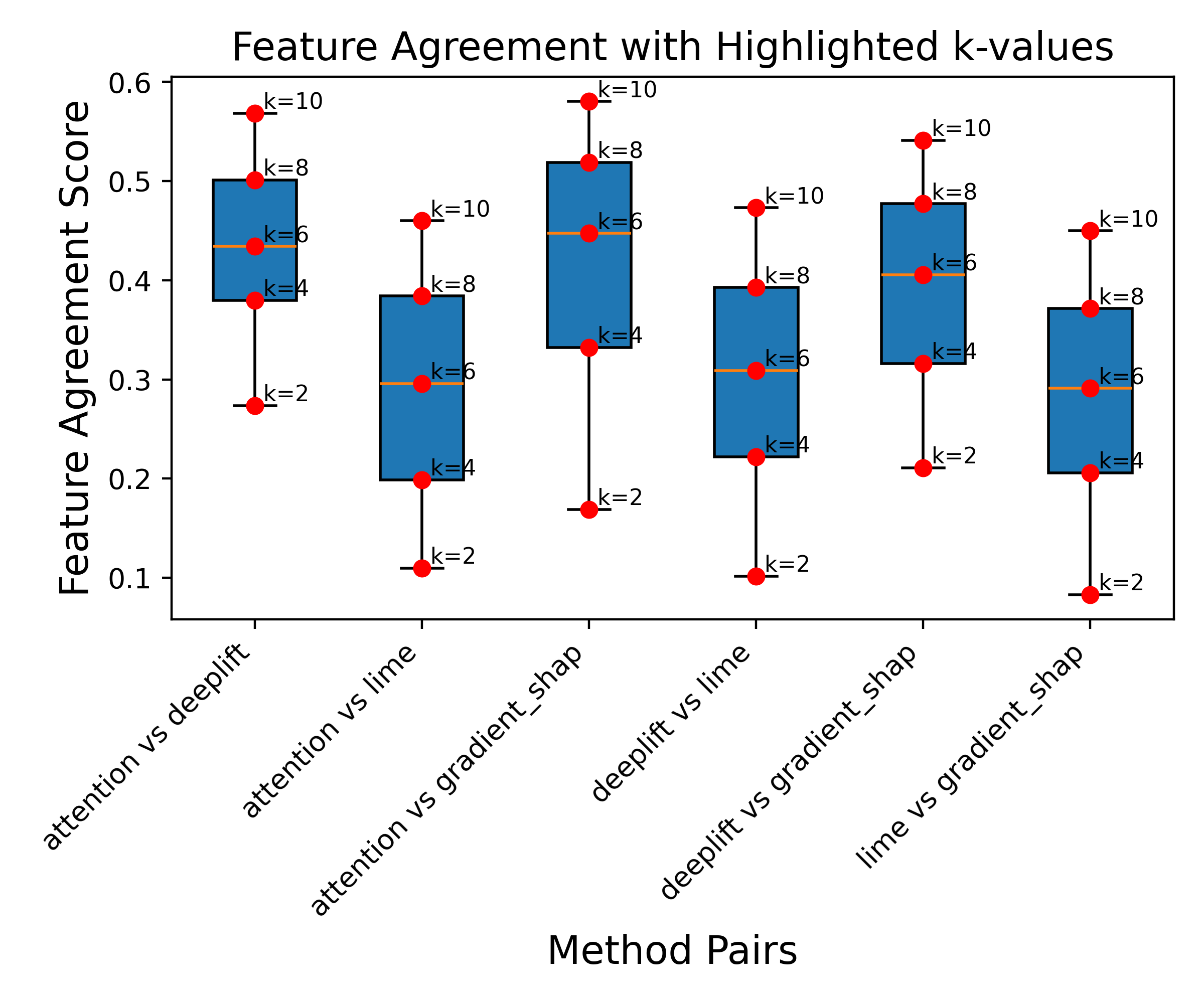}
        \caption{Box plot of feature agreement with highlighted k-values for CNN/DM.}
        \label{fig:box_plot_CNN}
    \end{subfigure}
    \caption{Box plot of Average feature agreement scores across highlighted k-values for Xsum (Fig. \ref{fig:box_plot_xsum}) and CNN/DM dataset (Fig. \ref{fig:box_plot_CNN}). The box plot of the Xsum dataset indicates low variability and high feature agreement scores compared to CNN/DM, showing high variability and lower agreement scores.}
    \label{fig:FA_boxplot_side_by_side}
\end{figure} 

\item \textbf{Box Plot of Rank Agreement}
The box plot of rank agreement scores is depicted in Fig. \ref{fig:RA_boxplot_side_by_side}. The box plot of Xsum and CNN/DM dataset shows that the variability of attention vs. DeepLIFT and DeepLIFT vs. Gradient SHAP is high for rank agreement scores compared to all other method pairs. The variability of rank agreement scores is relatively lower in CNN/DM dataset compared to Xsum. \\

\begin{figure}[H]
    \centering
    \begin{subfigure}[b]{0.49\linewidth}
        \centering
        \includegraphics[height=5.3cm]{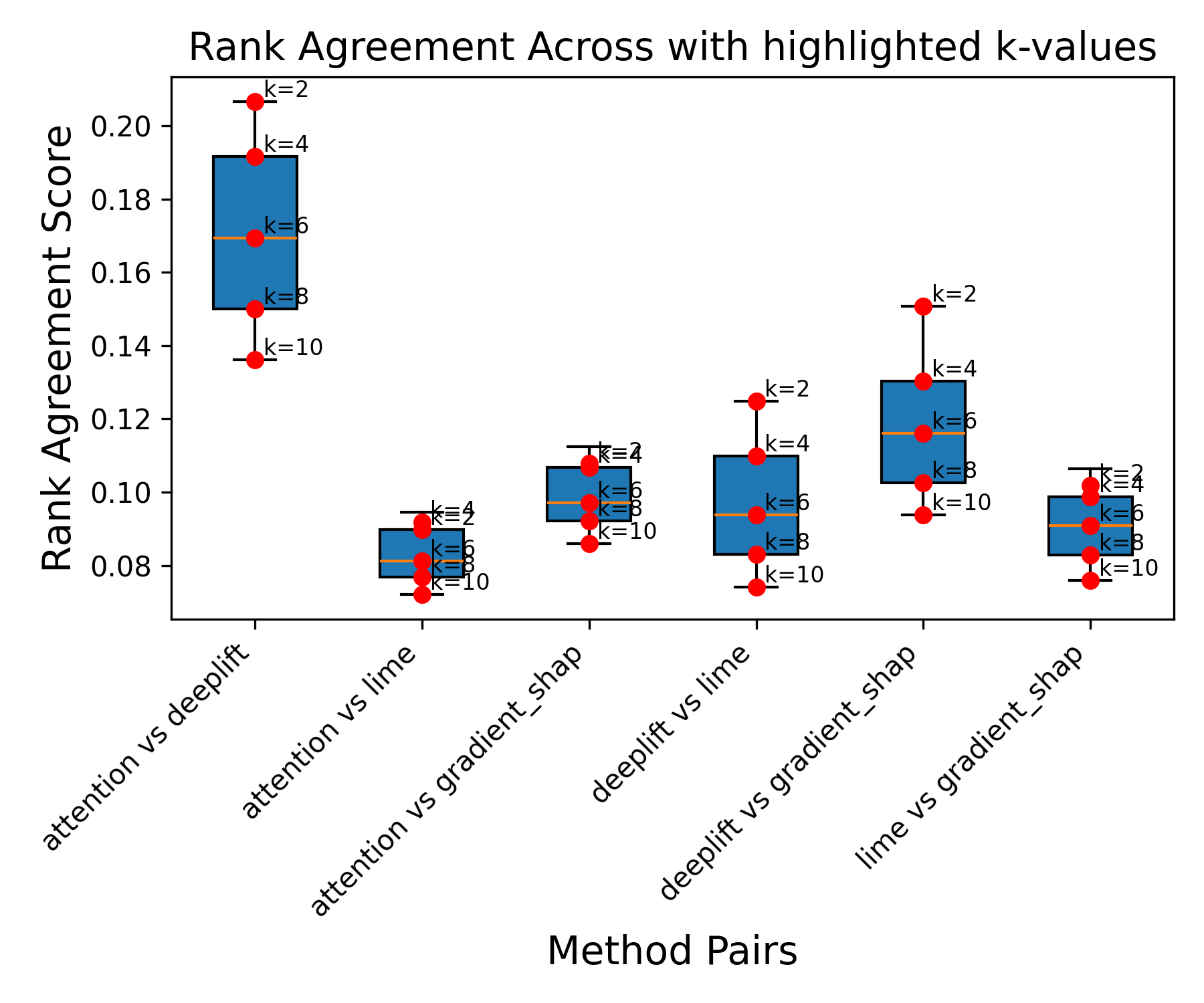}
        \caption{Box plot of rank agreement with highlighted k-values for Xsum.}
        \label{fig:RA_box_plot_xsum}
    \end{subfigure}
    \hfill
    \begin{subfigure}[b]{0.49\textwidth}
        \centering
        \includegraphics[width=\textwidth]{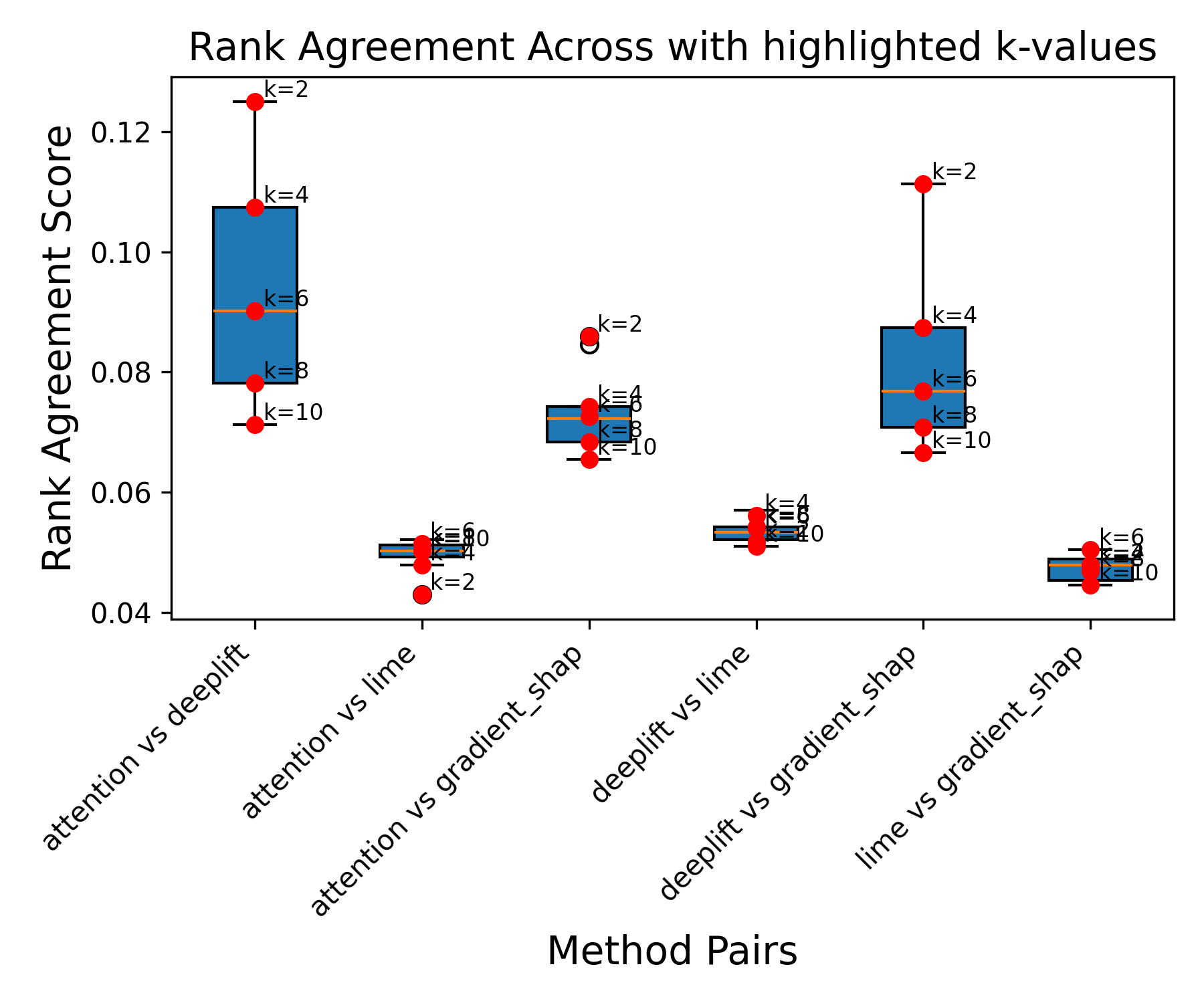}
        \caption{Box plot of rank agreement with highlighted k-values for CNN/DM.}
        \label{fig:RA_box_plot_CNN}
    \end{subfigure}
    \caption{Box plot of Average rank agreement scores across highlighted k-values for Xsum (Fig. \ref{fig:RA_box_plot_xsum}) and CNN/DM dataset (Fig. \ref{fig:RA_box_plot_CNN}). The box plot of the Xsum and CNN/DM dataset indicates low variability of rank agreement scores.}
    \label{fig:RA_boxplot_side_by_side}
\end{figure}

\item \textbf{Pairwise Rank Agreement Heatmaps:} The Fig. \ref{fig:PRA_heatmap_side_by_side} illustrates the average pairwise rank agreement scores for both datasets. The magnitude of agreement scores based on pairwise rank agreement is relatively higher compared to the scores achieved by all other metrics.

\begin{figure}[H]
    \centering
    \begin{subfigure}[b]{0.49\linewidth}
        \centering
        \includegraphics[width=\textwidth]{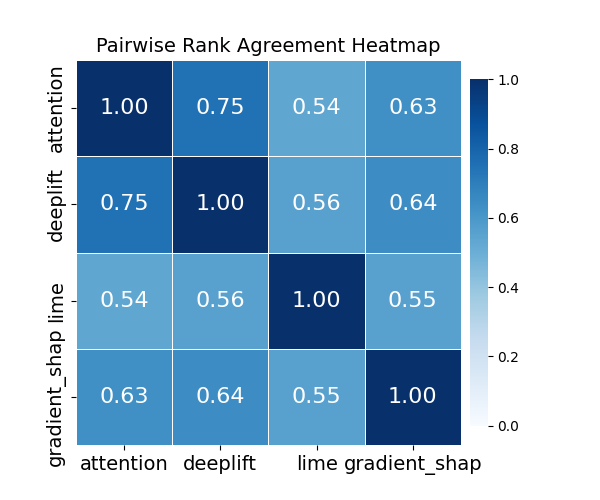}
        \caption{Pairwise rank agreement heatmap for Xsum.}
        \label{fig:PRA_heatmap_xsum}
    \end{subfigure}
    \hfill
    \begin{subfigure}[b]{0.49\linewidth}
        \centering
        \includegraphics[width=\textwidth]{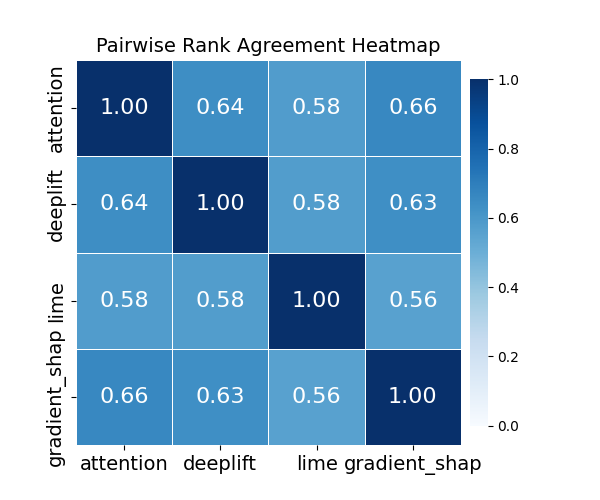}
        \caption{Pairwise rank agreement heatmap for CNN/DM.}
        \label{fig:PRA_heatmap_CNN}
    \end{subfigure}
    \caption{Average pairwise rank agreement heatmap for Xsum (\ref{fig:PRA_heatmap_xsum}) and CNN/DM (\ref{fig:PRA_heatmap_CNN}) datasets.}
    \label{fig:PRA_heatmap_side_by_side}
\end{figure}

\end{itemize}

\subsection{Disagreement analysis after applying RXAI framework (Same representative batch used in Section:\ref{Disgreement_results}) } \label{Regional_results}
\begin{itemize}

\item \textbf{Effectiveness of RXAI Framework:} The heatmaps in Figure: \ref{fig:global_vs_regional_fa_xsum} and Figure: \ref{fig:global_vs_regional_fa_cnndm} are added to compare the effectiveness of RXAI framework for k=4 across different XAI method pairs. The results clearly indicate enhanced feature agreement scores.

\begin{figure}[H]
    \centering
    \begin{subfigure}[b]{0.45\linewidth}
        \centering
        \includegraphics[width=\textwidth]{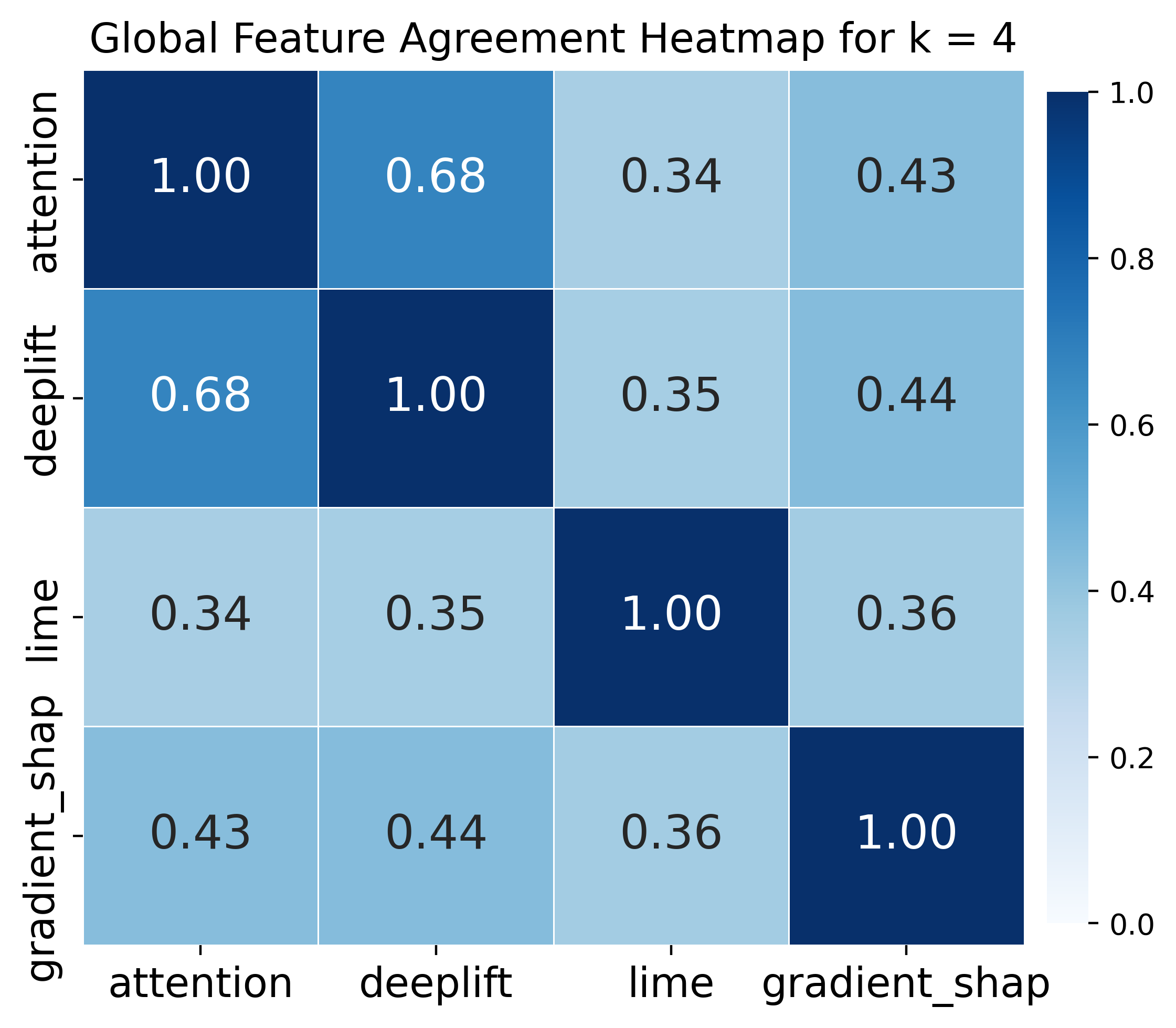}
        \caption{Global feature agreement heatmap for $k = 4$ on XSum dataset.}
        \label{fig:global_fa_xsum}
    \end{subfigure}
    \hfill
    \begin{subfigure}[b]{0.45\linewidth}
        \centering
        \includegraphics[width=\textwidth]{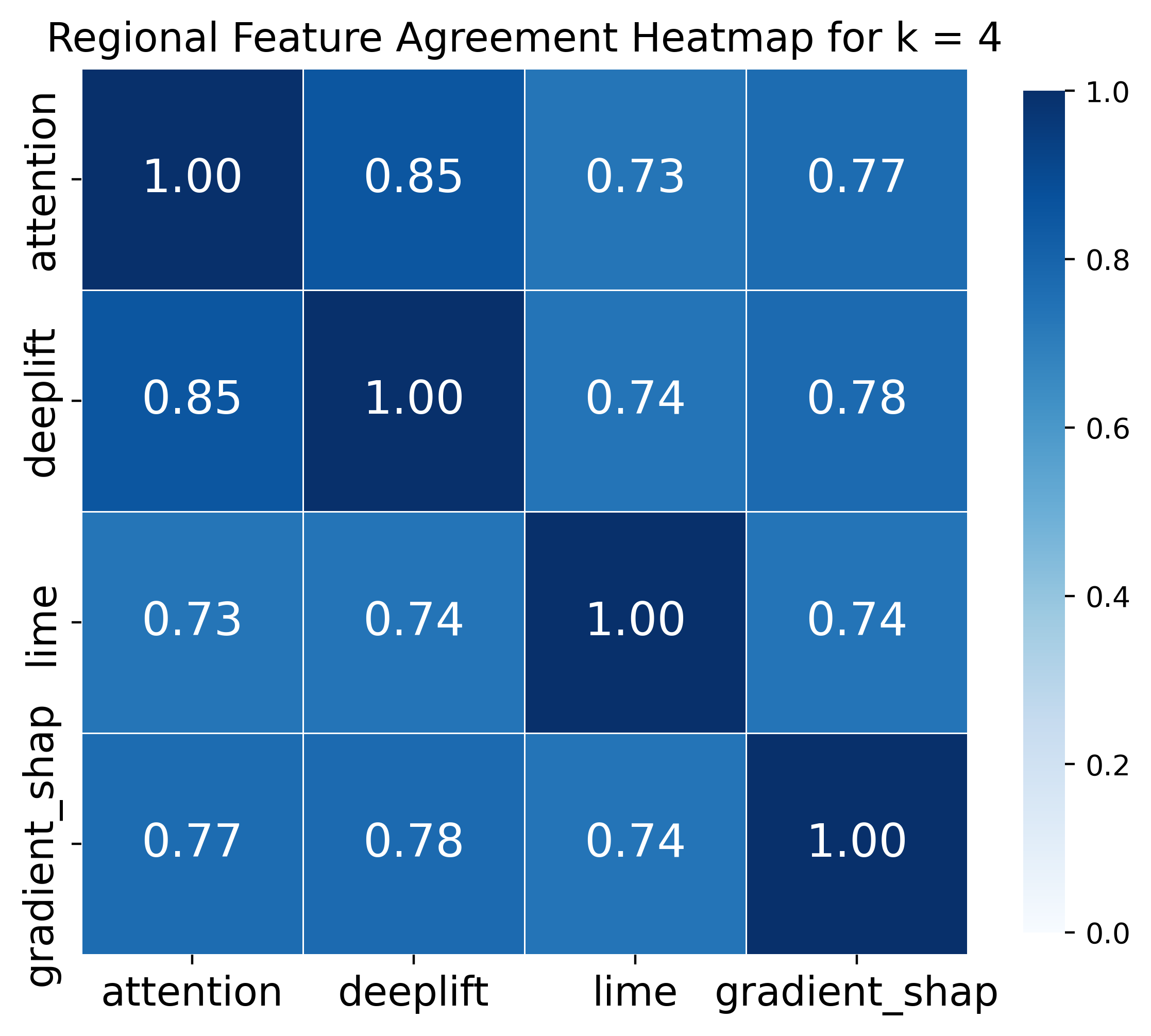}
        \caption{Regional feature agreement heatmap for $k = 4$ on XSum dataset.}
        \label{fig:regional_fa_xsum}
    \end{subfigure}
    \caption{Global vs. Regional explanation heatmap based on feature agreement for k= 4 on XSum dataset. The heatmaps demonstrate significant improvement with p-values $<$ 0.05 (refer Table: \ref{tab:wilcoxon_xsum_fa}) across all method pairs for feature agreement scores at the regional level compared to the global analysis.}
    \label{fig:global_vs_regional_fa_xsum}
\end{figure}

\begin{figure}[H]
    \centering
    \begin{subfigure}[b]{0.45\textwidth}
        \centering
        \includegraphics[width=\textwidth]{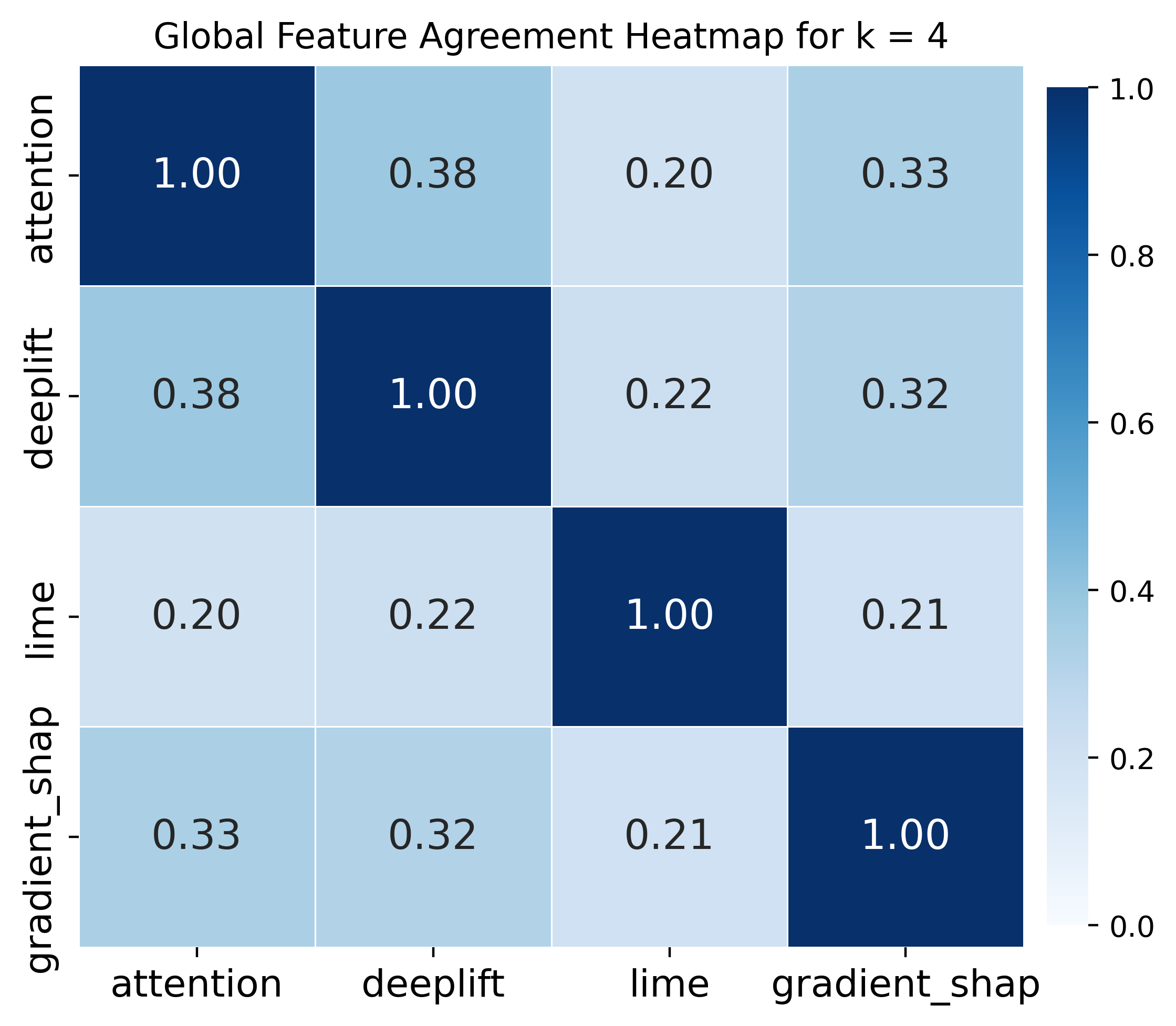}
        \caption{Global feature agreement for k = 4 on CNN/DM dataset.}
        \label{fig:global_fa_cnndm}
    \end{subfigure}
    \hfill
    \begin{subfigure}[b]{0.45\textwidth}
        \centering
        \includegraphics[width=\textwidth]{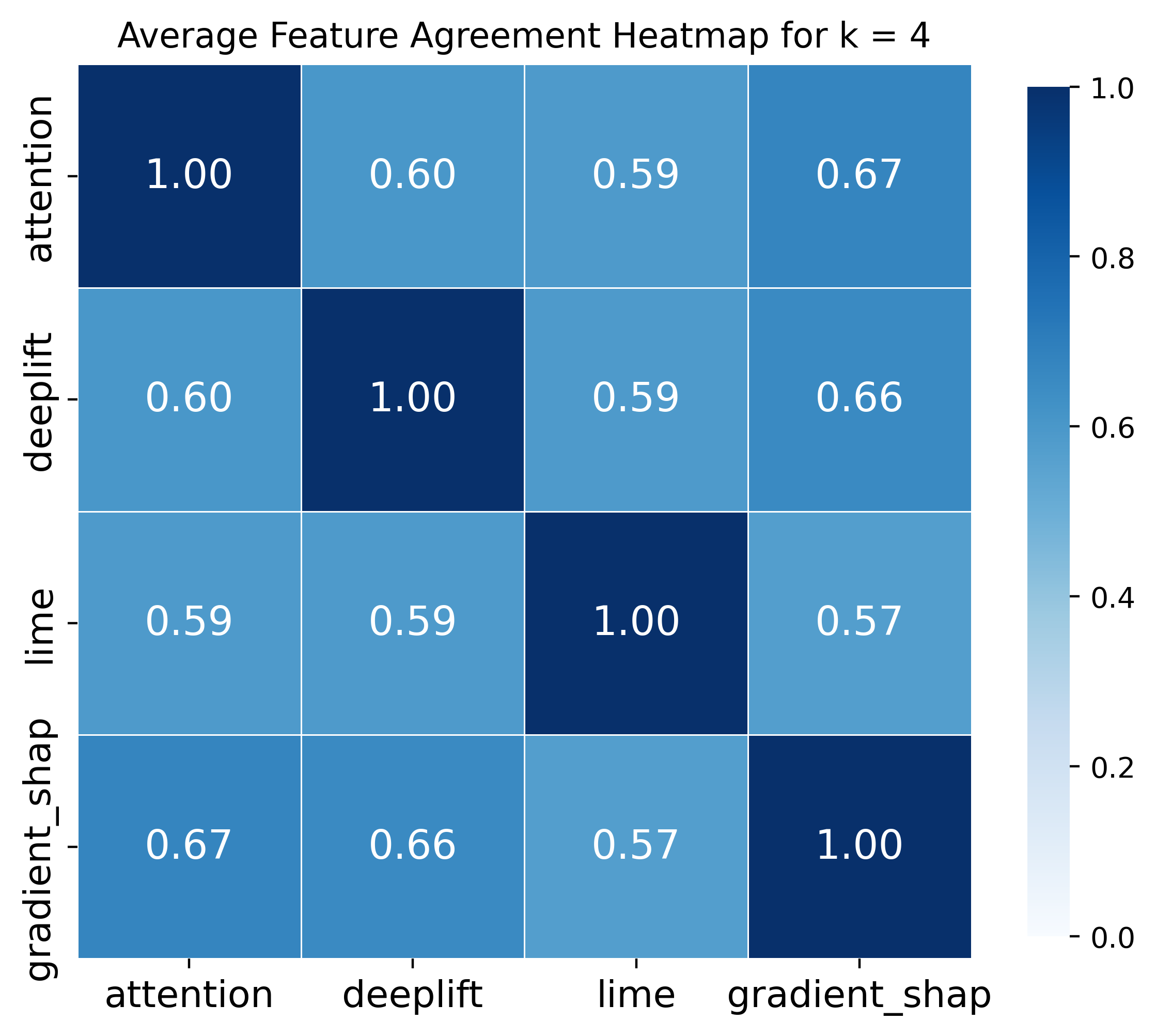}
        \caption{Regional feature agreement heatmap for k = 4 on CNN/DM dataset.}
        \label{fig:regional_fa_cnndm}
    \end{subfigure}
    \caption{Global vs. Regional explanation heatmap based on feature agreement for k = 4 on CNN/DM dataset. The heatmaps demonstrate significant improvement with p-values $<$ 0.05 (refer Table: \ref{tab:wilcoxon_batch2_cnn}) across all method pairs for feature agreement scores at the regional level compared to the global analysis.}
    \label{fig:global_vs_regional_fa_cnndm}
\end{figure}

\item \textbf{Semantic Alignment Score Analysis:} The impact of RXAI framework on semantic alignment is illustrated in Figure. \ref{fig:Regional_SAS_heatmap_side_by_side}. The heatmap presents average SAS scores aggregated across all k-values due to its consistent nature. The results indicates that RXAI framework does not lead to a significant improvement in semantic alignment scores.

\begin{figure}[H]
    \centering
    \begin{subfigure}[b]{0.49\textwidth}
        \centering
        \includegraphics[width=\textwidth]{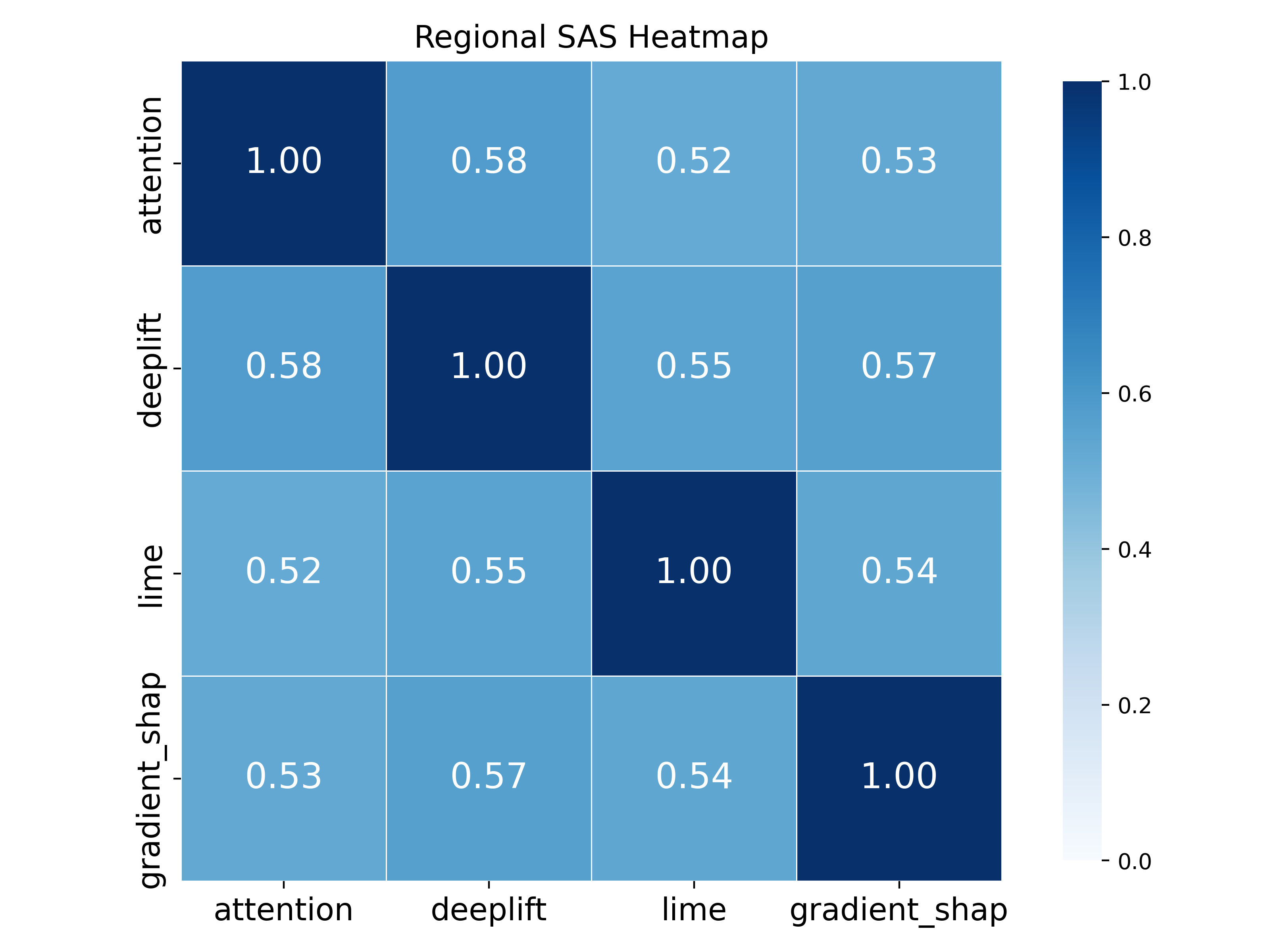} 
        \caption{SAS heatmap for segmented articles of Xsum.}
        \label{fig:Regional_SAS_heatmap_xsum}
    \end{subfigure}
    \hfill
    \begin{subfigure}[b]{0.49\textwidth}
        \centering
        \includegraphics[width=\textwidth]{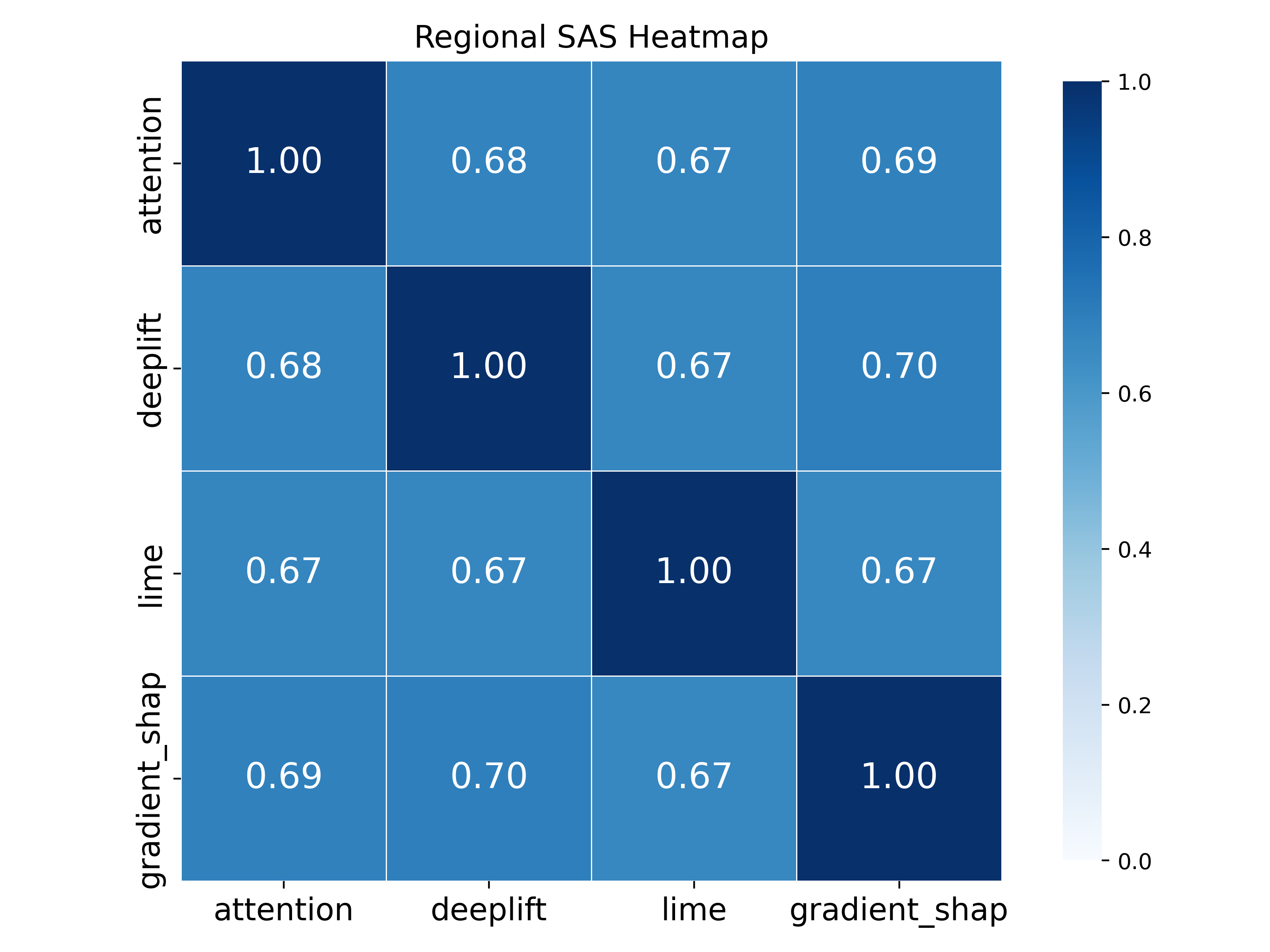}
        \caption{SAS heatmap for segmented articles of CNN/DM.}
        \label{fig:Regional_SAS_heatmap_CNN}
    \end{subfigure}
    \caption{Average SAS heatmap of Segmented articles of Xsum (Fig. \ref{fig:Regional_SAS_heatmap_xsum}) and CNN/DM \ref{fig:Regional_SAS_heatmap_CNN} datasets. Overall agreement scores based on SAS indicates no significant improvement in semantic alignment scores after segmentation.}
    \label{fig:Regional_SAS_heatmap_side_by_side}
\end{figure}

\item \textbf{Rank Correlation Analysis:} The disagreement analysis of Regional Rank Correlation analysis shows no improvement in disagreement scores after applying the RXAI framework, as shown in Figure \ref{fig:Regional_SRA_heatmap_side_by_side}. 

\begin{figure}[H]
    \centering
    \begin{subfigure}[b]{0.49\textwidth}
        \centering
        \includegraphics[width=\textwidth]{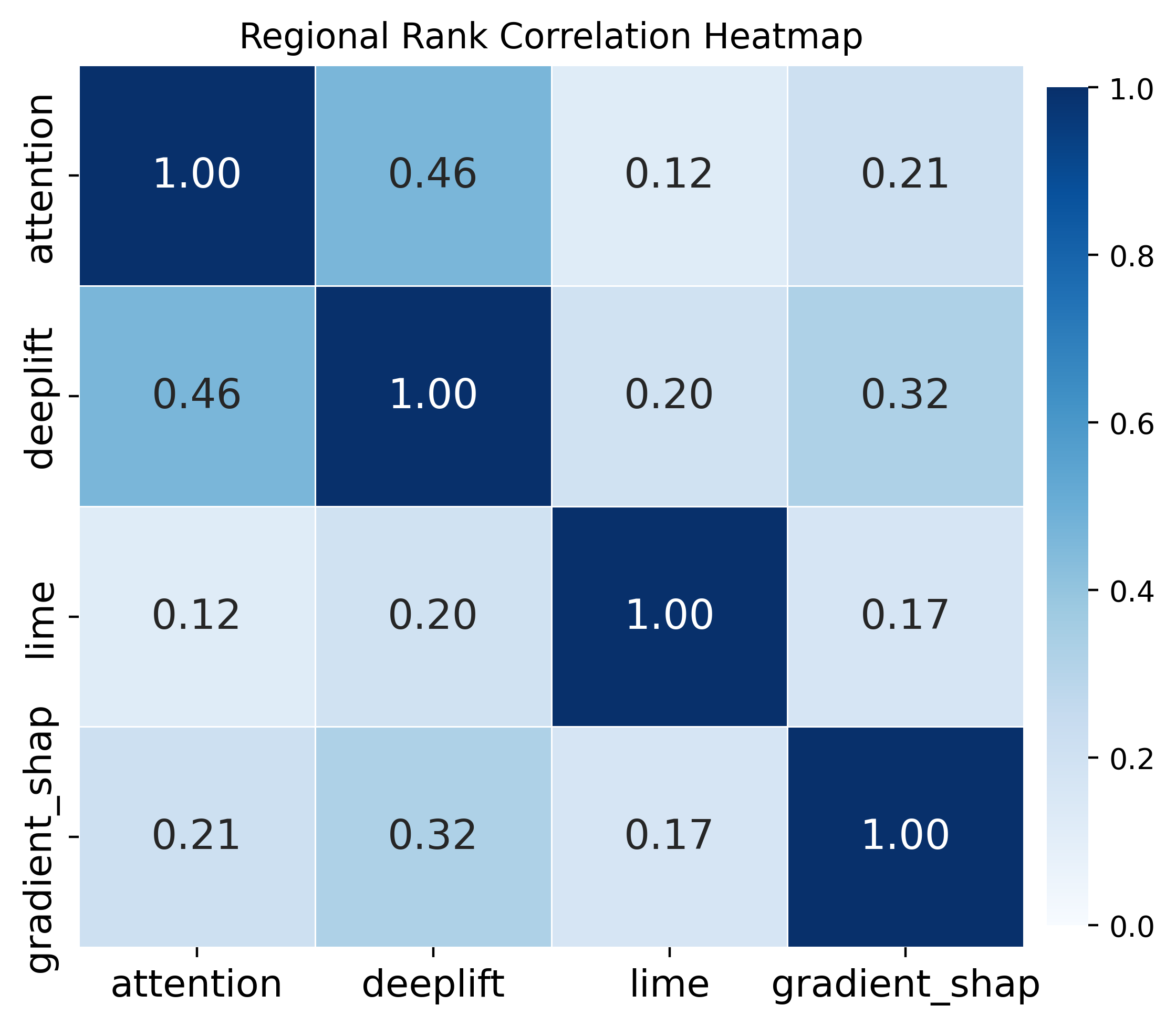}
        \caption{Rank correlation heatmap for segmented articles of Xsum.}
        \label{fig:Regional_SRA_heatmap_xsum}
    \end{subfigure}
    \hfill
    \begin{subfigure}[b]{0.49\textwidth}
        \centering
        \includegraphics[width=\textwidth]{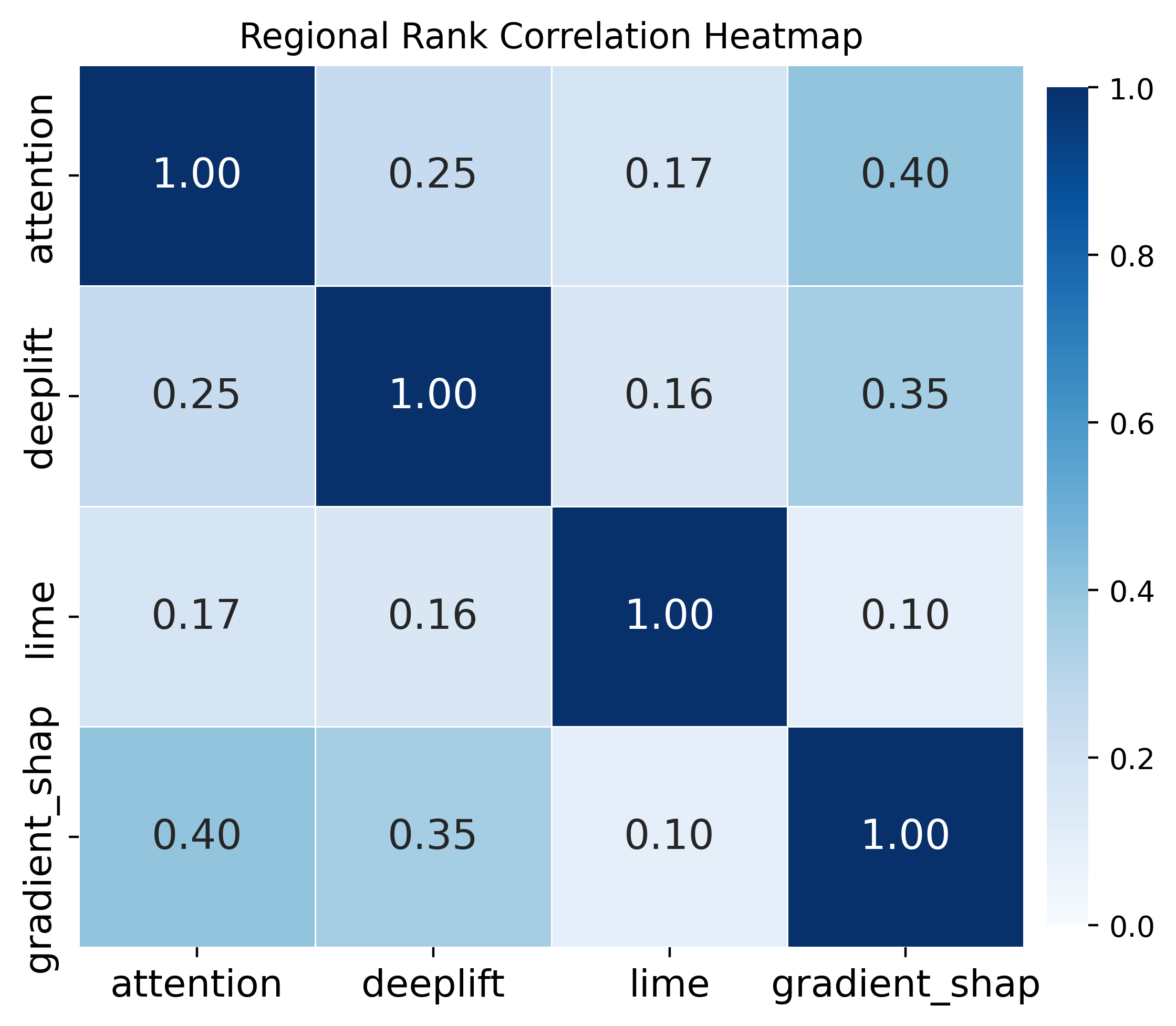}
        \caption{Rank correlation heatmap for segmented articles of CNN/DM.}
        \label{fig:Regional_SRA_heatmap_CNN}
    \end{subfigure}
    \caption{Average spearman rank correlation heatmap of Segmented articles of Xsum (Fig. \ref{fig:Regional_SRA_heatmap_xsum}) and CNN/DM \ref{fig:Regional_SRA_heatmap_CNN} datasets. Overall agreement scores based on rank correlation are low, indicating no improvement in disagreement scores after segmentation.}
    \label{fig:Regional_SRA_heatmap_side_by_side}
\end{figure}

\item \textbf{Pairwise Rank Agreement Analysis:} The disagreement analysis of Pairwise Rank Agreement metric on Xsum and CNN/DM datasets after implementing RXAI framework is depicted in Fig. \ref{fig:fig:PRA_heatmap_side_by_side_regional} . The pairwise rank agreement heatmap shows a moderate degree of agreement between most methods in both datasets. Segmentation does not help in improving the pairwise rank agreement scores. 

\begin{figure}[H]
    \centering
    \begin{subfigure}[b]{0.49\textwidth}
        \centering
        \includegraphics[width=\textwidth]{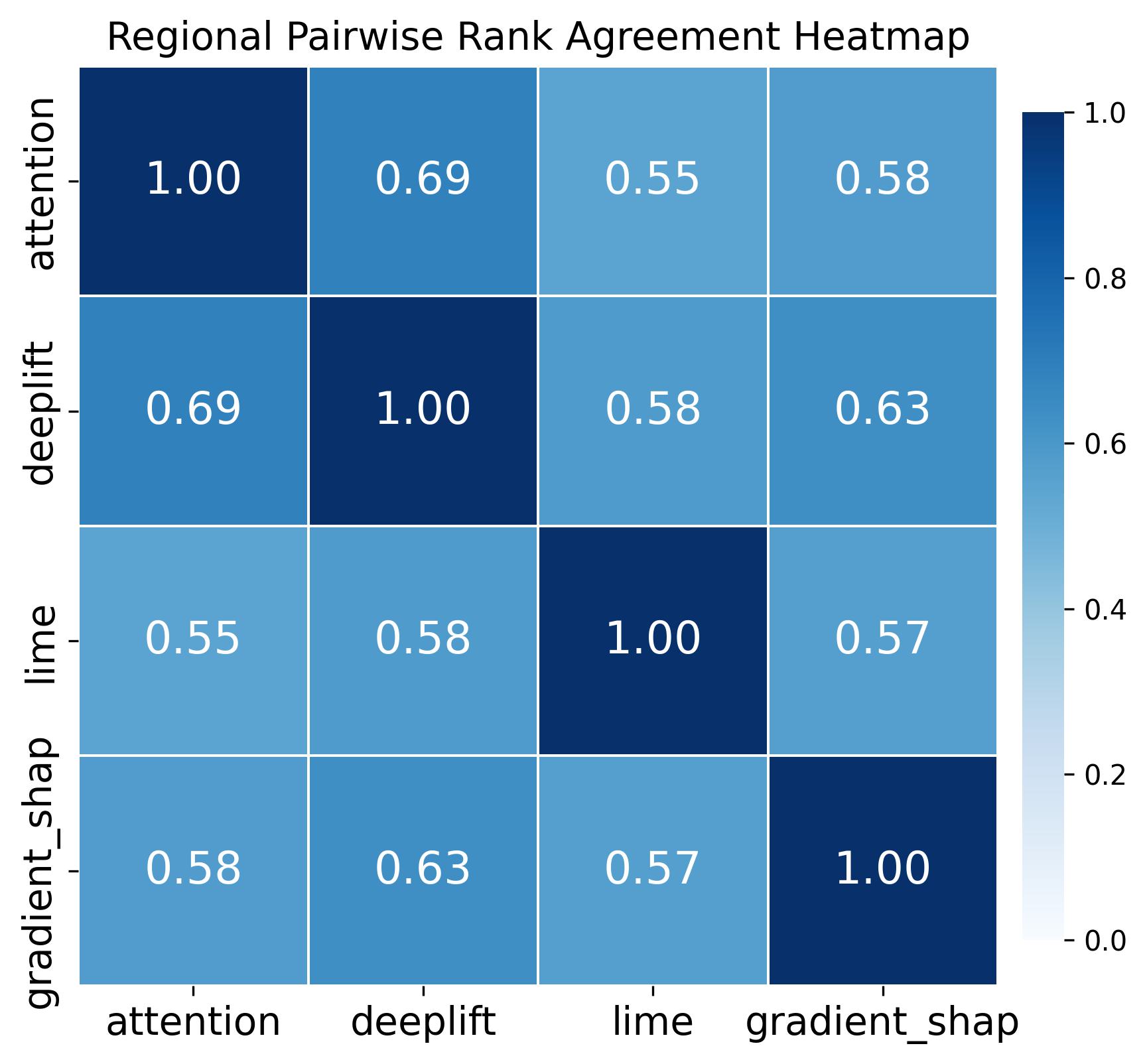}
        \caption{Pairwise Rank Agreement Heatmap (Segmented Articles).}
        \label{fig:pairwise_rank_agreement_regional_xsum}
    \end{subfigure}
    \hfill
    \begin{subfigure}[b]{0.49\textwidth}
        \centering
        \includegraphics[width=\textwidth]{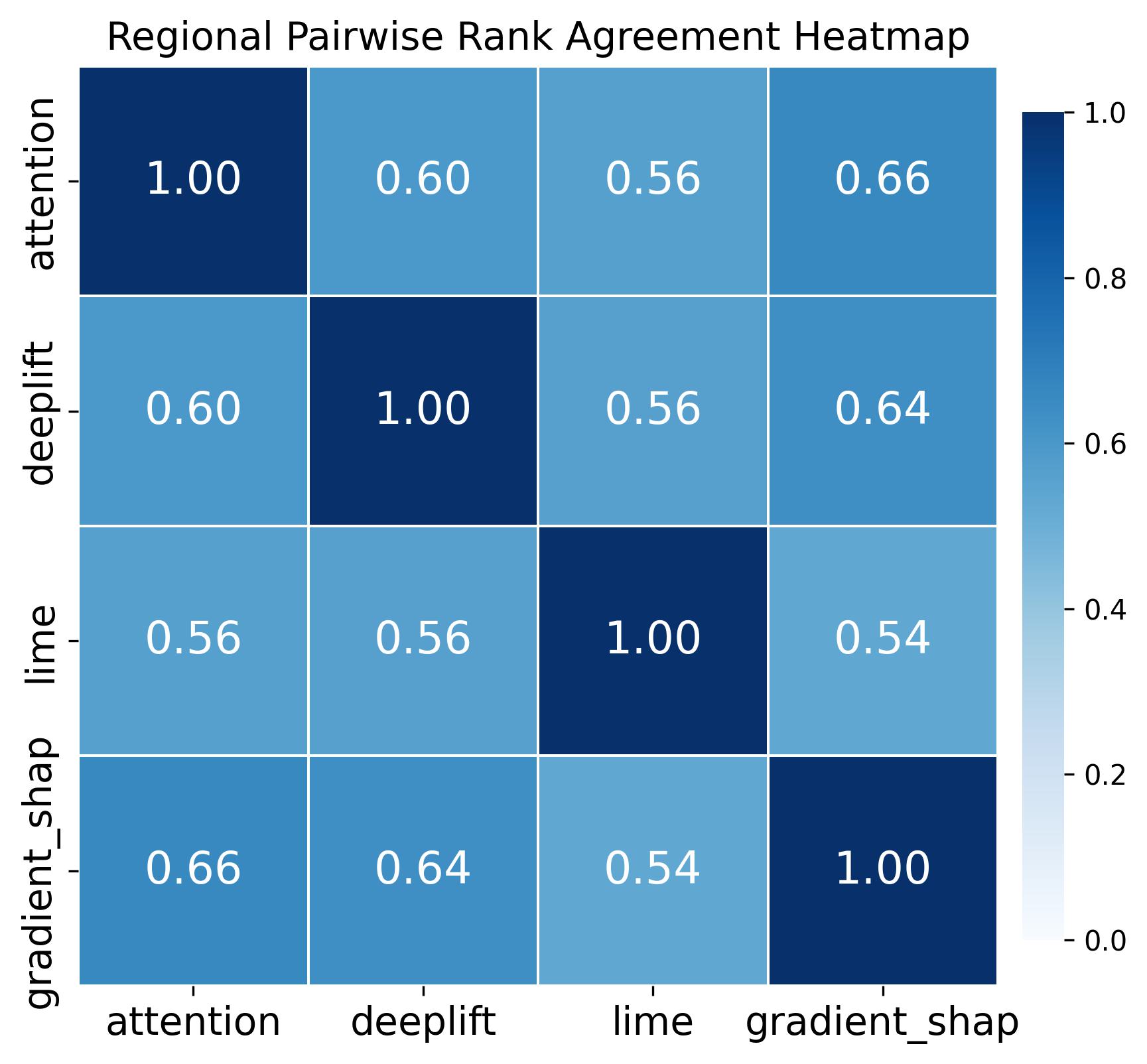}
        \caption{Pairwise Rank Agreement Heatmap (Segmented Articles).}
        \label{fig:pairwise_rank_agreement_regional_CNNDM}
    \end{subfigure}
    \caption{Average pairwise rank agreement heatmap for segmented explanations of Xsum (\ref{fig:PRA_heatmap_xsum}) and CNN/DM (\ref{fig:PRA_heatmap_CNN}) datasets. The heatmap of both datasets highlights that segmentation does not help in enhancing the pairwise rank agreement scores and mostly stays consistent with the global pairwise rank agreement score.}
    \label{fig:fig:PRA_heatmap_side_by_side_regional}
\end{figure}

\end{itemize}

\subsection{Batched Analysis of XAI Disagreement}

\subsubsection{Global Disagreement Analysis}
The disagreement analysis was carried out across 11 random batches, each with 500 samples (5500 total), to guarantee statistically robust results. While results from one representative batch are included in the main results Section: \ref{Disgreement_results}, this appendix presents findings from an additional representative batch along with average agreement scores across all 11 batches. The results from this batch, discussed in this section, demonstrate strong consistency in agreement scores across XAI methods. Notably, this consistency is not limited to these two batches. To evaluate whether the observed consistency persists across all batches, we employed a one-way ANOVA test, the results of which are discussed in Section \ref{stats_test}. Here, we present batch-level agreement comparisons (Batch 2 from the XSum and CNN/DM dataset) across four key metrics: SAS, Feature Agreement, and Spearman Rank Correlation. 

To evaluate the degree of disagreement of an individual explanation method with every other XAI method, we conducted combined comparisons at k=4. We measured how well each XAI method agrees with every other method by averaging pairwise feature agreement scores. This approach enables us to determine the consistency or divergence of a given method related to the rest, offering a combined perspective on the consistency of XAI methods. The results show that attention vs all other methods attained the FA value of 0.49, followed by DeepLIFT vs all other methods having FA value of 0.5, followed by Gradient SHAP vs all other methods having FA value of 0.43 and LIME  vs all other methods having the lowest FA value of 0.38.

A summarized view of global disagreement analysis based on Feature agreement and Rank agreement across all 11 batches is presented in Table \ref{tab:fa_summary_side_by_side} and Table \ref{tab:ra_summary_side_by_side}. The results of feature and Rank agreement at $k$=4 and $k$=8 with very low standard deviation values indicate that the global agreement scores remain highly consistent across batches.

\begin{table*}[h!]
\small
\centering
\caption{Mean Feature Agreement (FA) across 11 batches for XSum and CNN/DM datasets, computed for selected $k$ values. The table highlights average agreement consistency with low standard deviation values, indicating consistent FA values for $k=4$ and $k=8$.} 
\begin{tabular}{lcccc}
\toprule
\textbf{Method Pair} & \textbf{XSum $k=4$} & \textbf{XSum $k=8$} & \textbf{CNN $k=4$} & \textbf{CNN $k=8$} \\
\midrule
Attention vs DeepLIFT       & 0.68 ± 0.007 & 0.75 ± 0.006 & 0.40 ± 0.018 & 0.55 ± 0.026 \\
Attention vs Gradient SHAP   & 0.44 ± 0.017 & 0.62 ± 0.007 & 0.34 ± 0.013 & 0.54 ± 0.015 \\
Attention vs LIME           & 0.36 ± 0.014 & 0.55 ± 0.011 & 0.23 ± 0.018 & 0.44 ± 0.031 \\
DeepLIFT vs Gradient SHAP    & 0.45 ± 0.011 & 0.62 ± 0.009 & 0.34 ± 0.019 & 0.52 ± 0.023 \\
DeepLIFT vs LIME            & 0.37 ± 0.017 & 0.55 ± 0.011 & 0.24 ± 0.015 & 0.44 ± 0.029 \\
LIME vs Gradient SHAP        & 0.38 ± 0.011 & 0.56 ± 0.010 & 0.23 ± 0.018 & 0.42 ± 0.033 \\
\bottomrule
\end{tabular}
\label{tab:fa_summary_side_by_side}
\end{table*}

\begin{table*}[h!]
\small
\centering
\caption{Mean Rank Agreement (RA) across 11 batches for XSum and CNN/DM datasets, computed for selected $k$ values. The table highlights average agreement consistency with low standard deviation values, indicating consistency for $k=4$ and $k=8$.}
\begin{tabular}{lcccc}
\toprule
\textbf{Method Pair} & \textbf{XSum $k=4$} & \textbf{XSum $k=8$} & \textbf{CNN $k=4$} & \textbf{CNN $k=8$} \\
\midrule
Attention vs DeepLIFT       & 0.19 ± 0.005 & 0.15 ± 0.004 & 0.12 ± 0.010 & 0.09 ± 0.006 \\
Attention vs Gradient SHAP   & 0.12 ± 0.012 & 0.10 ± 0.006 & 0.08 ± 0.009 & 0.07 ± 0.006 \\
Attention vs LIME           & 0.10 ± 0.005 & 0.08 ± 0.003 & 0.06 ± 0.007 & 0.06 ± 0.006 \\
DeepLIFT vs Gradient SHAP    & 0.13 ± 0.009 & 0.11 ± 0.005 & 0.10 ± 0.009 & 0.08 ± 0.006 \\
DeepLIFT vs LIME            & 0.10 ± 0.009 & 0.08 ± 0.004 & 0.06 ± 0.004 & 0.06 ± 0.005 \\
LIME vs Gradient SHAP        & 0.10 ± 0.008 & 0.09 ± 0.005 & 0.06 ± 0.006 & 0.05 ± 0.005 \\
\bottomrule
\end{tabular}
\label{tab:ra_summary_side_by_side}
\end{table*}

\begin{figure}[htbp]
    \centering
    \begin{subfigure}[b]{0.49\textwidth}
        \centering
        \includegraphics[width=\textwidth]{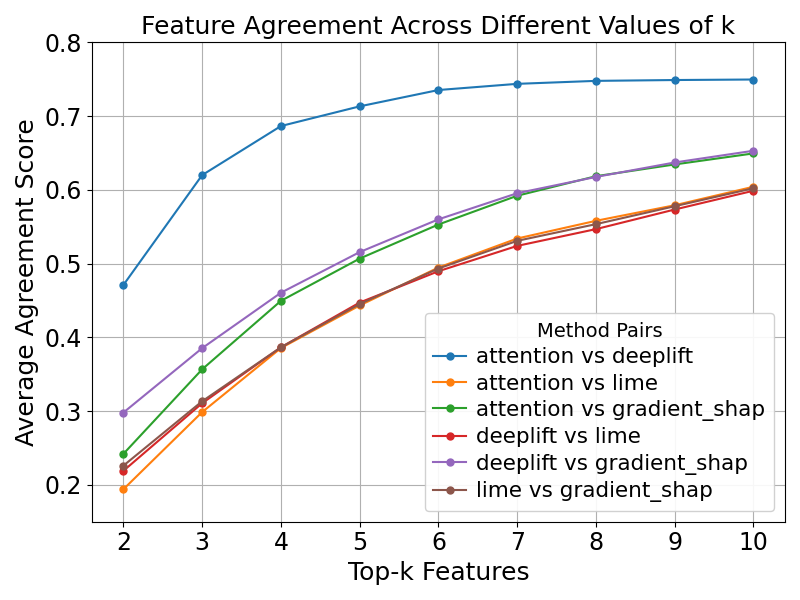}
        \caption{Global Feature Agreement scores for Xsum dataset across different k-values.}
    \end{subfigure}
    \hfill
    \begin{subfigure}[b]{0.49\textwidth}
        \centering
        \includegraphics[width=\textwidth]{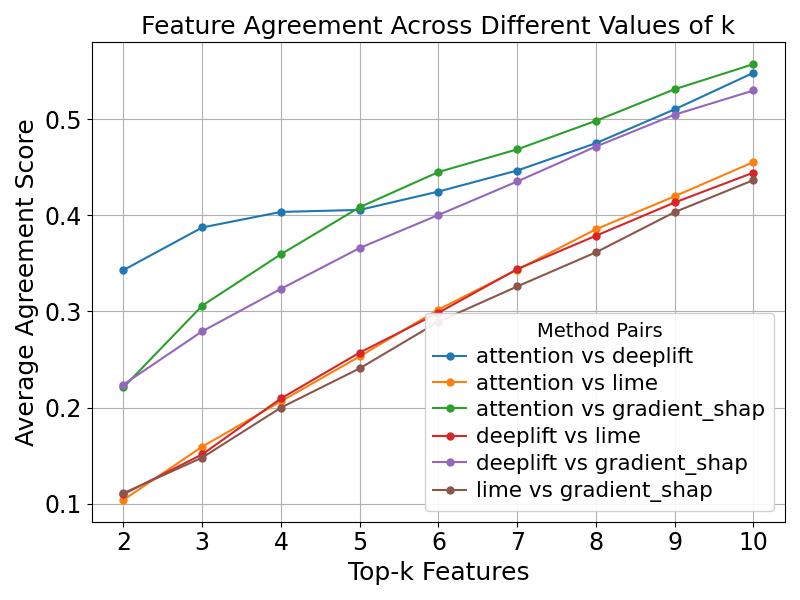}
        \caption{Global Feature Agreement scores for CNN/DM dataset across different k-values.}
    \end{subfigure}
    \caption{Global Feature Agreement across different top-$k$ values on XSum and CNN/DM datasets. The trend of global feature agreement scores across different method pairs is similar to the previous sample set analyzed in main results section \ref{Disgreement_results}.}
    \label{fig:fa-xsum-cnn}
\end{figure}

\begin{figure}[H]
    \centering
    \begin{subfigure}[b]{0.49\linewidth}
        \centering
        \includegraphics[width=\textwidth]{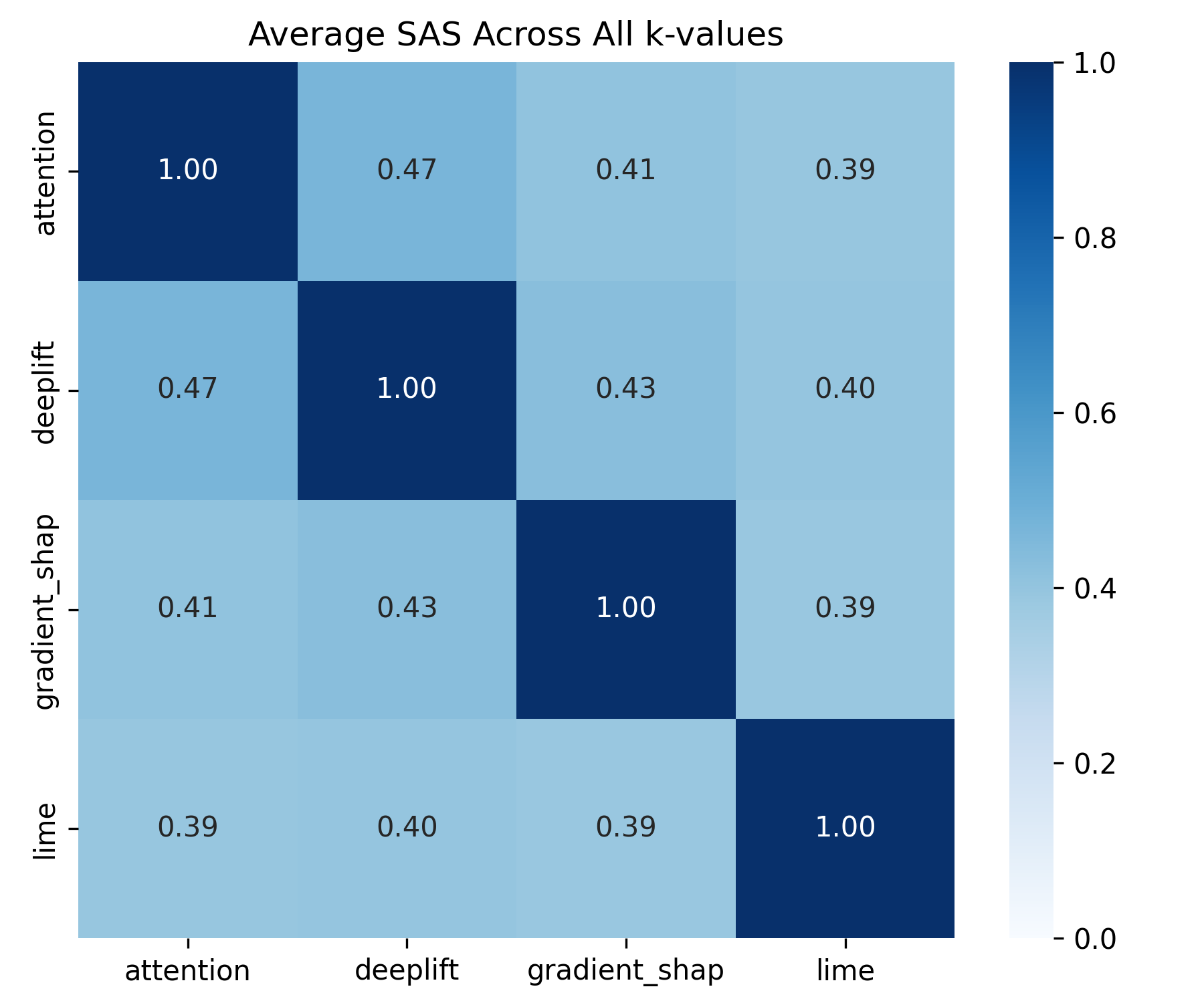}
        \caption{Global SAS Heatmap for Xsum.}
    \end{subfigure}
    \hfill
    \begin{subfigure}[b]{0.49\linewidth}
        \centering
        \includegraphics[width=\textwidth]{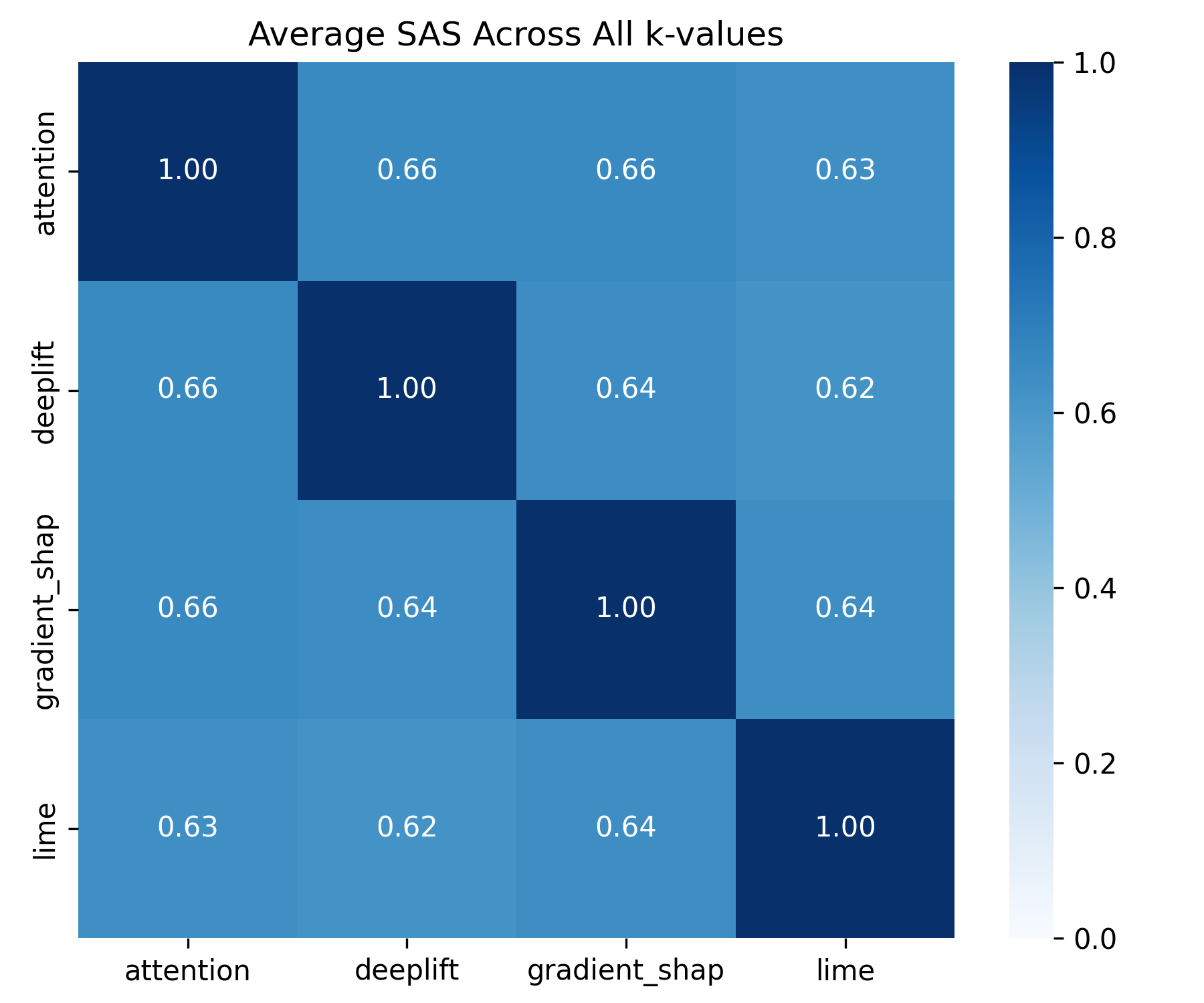}
        \caption{Global SAS Heatmap for CNN/DM.}
    \end{subfigure}
    \caption{Global Semantic Alignment Score (SAS) heatmaps for XSum and CNN/DM datasets across all k-values. The scores of global semantic alignment for this batch is consistent and depicts similar trend as other batches.}
    \label{fig:cas-xsum-cnn}
\end{figure}

\begin{figure}[H]
    \centering
    \begin{subfigure}[b]{0.49\textwidth}
        \centering
        \includegraphics[width=\textwidth]{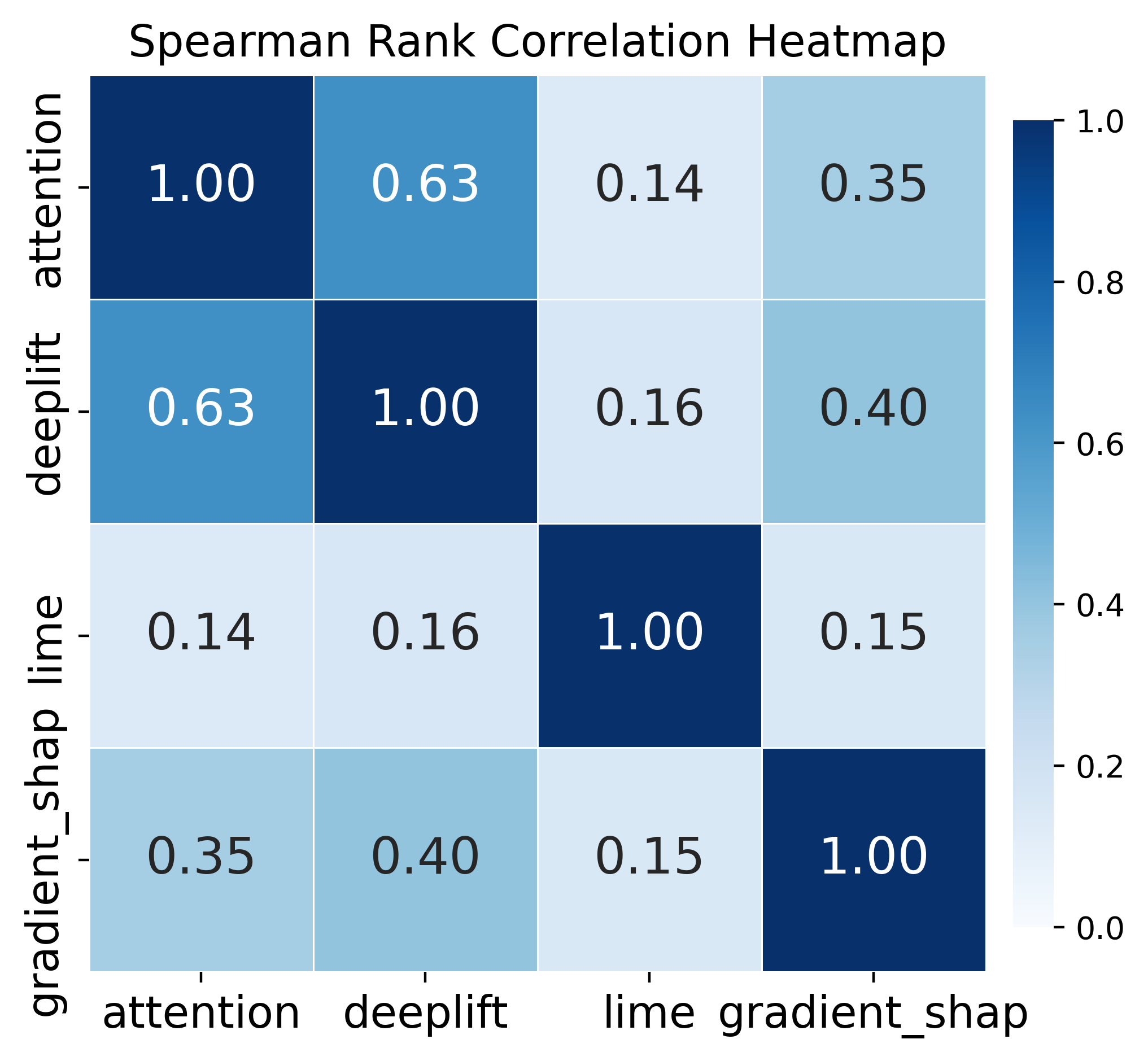}
        \caption{Global Spearman Rank Correlation Heatmap for Xsum dataset}
    \end{subfigure}
    \hfill
    \begin{subfigure}[b]{0.49\textwidth}
        \centering
        \includegraphics[width=\textwidth]{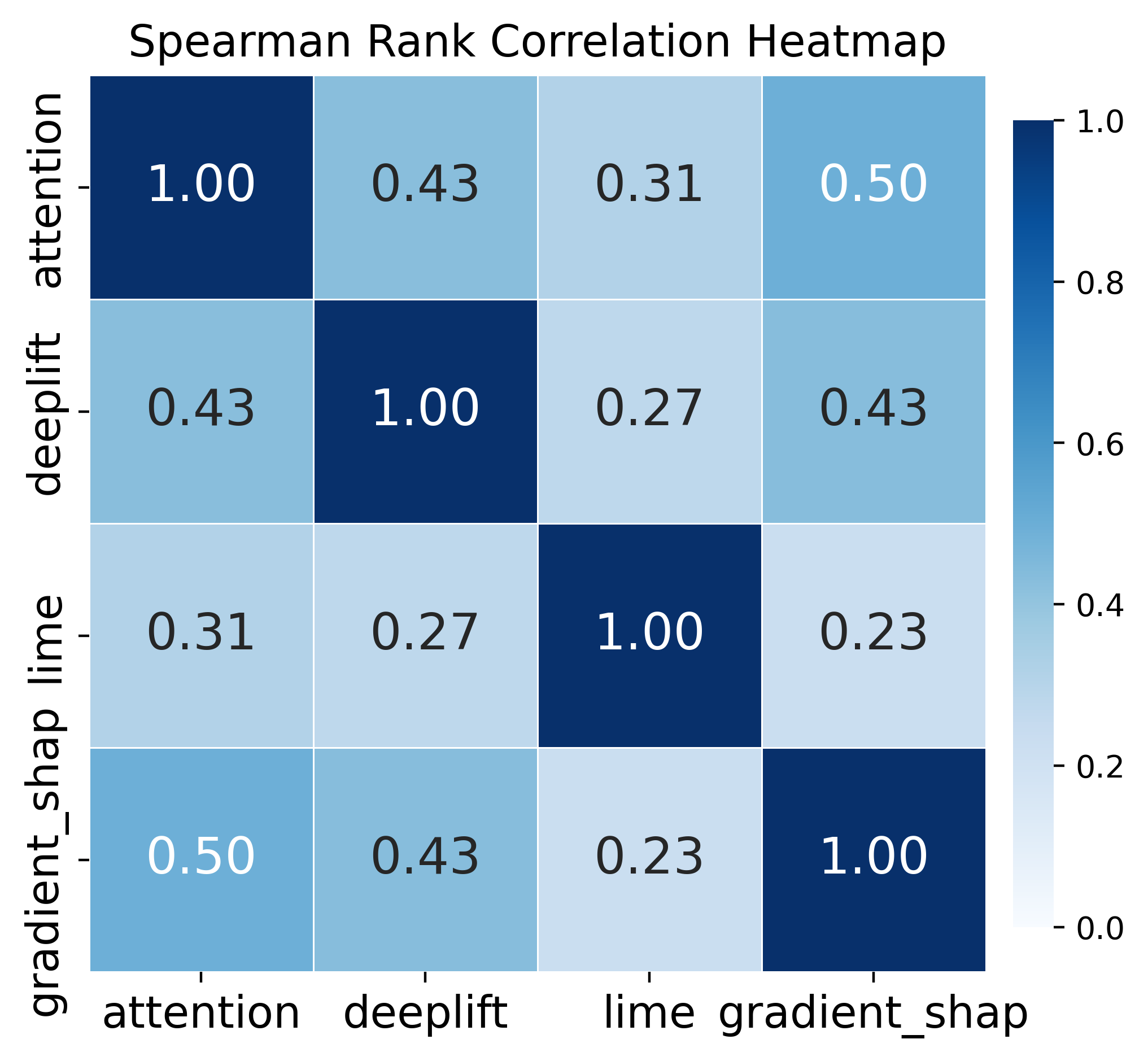}
        \caption{Global Spearman Rank Correlation for CNN/Dailymail dataset}
    \end{subfigure}
    \caption{Global Spearman Rank Correlation analysis across XAI methods on XSum and CNN/DM dataset. The heatmap depicts similar trends as observed in the previous
sample set of global rank correlation analysis.}
    \label{fig:sra-xsum-cnn}
\end{figure}

The results from Batch 2 presented in figures: \ref{fig:fa-xsum-cnn}, \ref{fig:cas-xsum-cnn}, \ref{fig:sra-xsum-cnn} illustrate similar agreement patterns between explanation methods as discussed in section: \ref{Disgreement_results}. While slight variations are observed, the scores across batches do not vary significantly. This consistency across all 11 batches supports the robustness of our findings reported in the main results section.

\subsubsection{Regional Disagreement Analysis:} We also evaluated the consistency of RXAI scores across batches; however, the agreement scores after segmentation exhibited slight variations among batches, which may be attributed to variation of segment lengths across batches. The summarized view of the RXAI framework based on Feature and Rank agreement across all 11 batches is presented in Table \ref{tab:fa_xsum_cnn} and Table \ref{tab:ra_summary_xsum_cnn}. 

\begin{table*}[h!]
\small
\centering
\caption{Mean Feature Agreement (FA) across 11 batches for XSum and CNN/DM datasets, computed for selected $k$ values. The table highlights Regional average feature agreement consistency with low standard deviation values.}
\label{tab:fa_xsum_cnn}
\begin{tabular}{lcccc}
\toprule
\textbf{Method Pair} & \textbf{XSum $k=2$} & \textbf{XSum $k=4$} & \textbf{CNN $k=2$} & \textbf{CNN $k=4$} \\
\midrule
attention vs DeepLIFT & 0.61 ± 0.050 & 0.82 ± 0.049 & 0.39 ± 0.037 & 0.65 ± 0.058 \\
attention vs Gradient\_SHAP & 0.44 ± 0.092 & 0.69 ± 0.103 & 0.45 ± 0.035 & 0.71 ± 0.052 \\
attention vs lime & 0.41 ± 0.093 & 0.64 ± 0.116 & 0.36 ± 0.039 & 0.63 ± 0.063 \\
DeepLIFT vs Gradient\_SHAP & 0.49 ± 0.096 & 0.70 ± 0.105 & 0.48 ± 0.032 & 0.69 ± 0.054 \\
DeepLIFT vs lime & 0.43 ± 0.094 & 0.65 ± 0.113 & 0.35 ± 0.037 & 0.63 ± 0.068 \\
lime vs gradient\_SHAP & 0.42 ± 0.093 & 0.65 ± 0.118 & 0.34 ± 0.038 & 0.62 ± 0.067 \\
\bottomrule
\end{tabular}
\end{table*}

\begin{table*}[h!]
 \small   
\centering
\caption{Mean Rank Agreement (RA) scores for XSum and CNN datasets, computed for selected $k$ values. The table highlights Regional average agreement consistency with low standard deviation values.}
\label{tab:ra_summary_xsum_cnn}
\begin{tabular}{lcccc}
\toprule
\textbf{Method Pair} & \textbf{XSum $k=2$} & \textbf{XSum $k=4$} & \textbf{CNN $k=2$} & \textbf{CNN $k=4$} \\
\midrule
attention vs DeepLIFT & 0.29 ± 0.030 & 0.25 ± 0.021 & 0.19 ± 0.020 & 0.18 ± 0.021 \\
attention vs Gradient\_SHAP & 0.21 ± 0.038 & 0.19 ± 0.028 & 0.23 ± 0.021 & 0.22 ± 0.018 \\
attention vs lime & 0.21 ± 0.045 & 0.17 ± 0.032 & 0.17 ± 0.022 & 0.16 ± 0.018 \\
DeepLIFT vs gradient\_SHAP & 0.28 ± 0.052 & 0.23 ± 0.037 & 0.28 ± 0.021 & 0.22 ± 0.021 \\
DeepLIFT vs lime & 0.25 ± 0.052 & 0.20 ± 0.032 & 0.17 ± 0.015 & 0.16 ± 0.015 \\
lime vs gradient\_SHAP & 0.23 ± 0.048 & 0.19 ± 0.033 & 0.17 ± 0.021 & 0.15 ± 0.020 \\
\bottomrule
\end{tabular}
\end{table*}

\begin{figure}[H]
    \centering
    \begin{subfigure}[b]{0.49\textwidth}
        \centering
        \includegraphics[width=\textwidth]{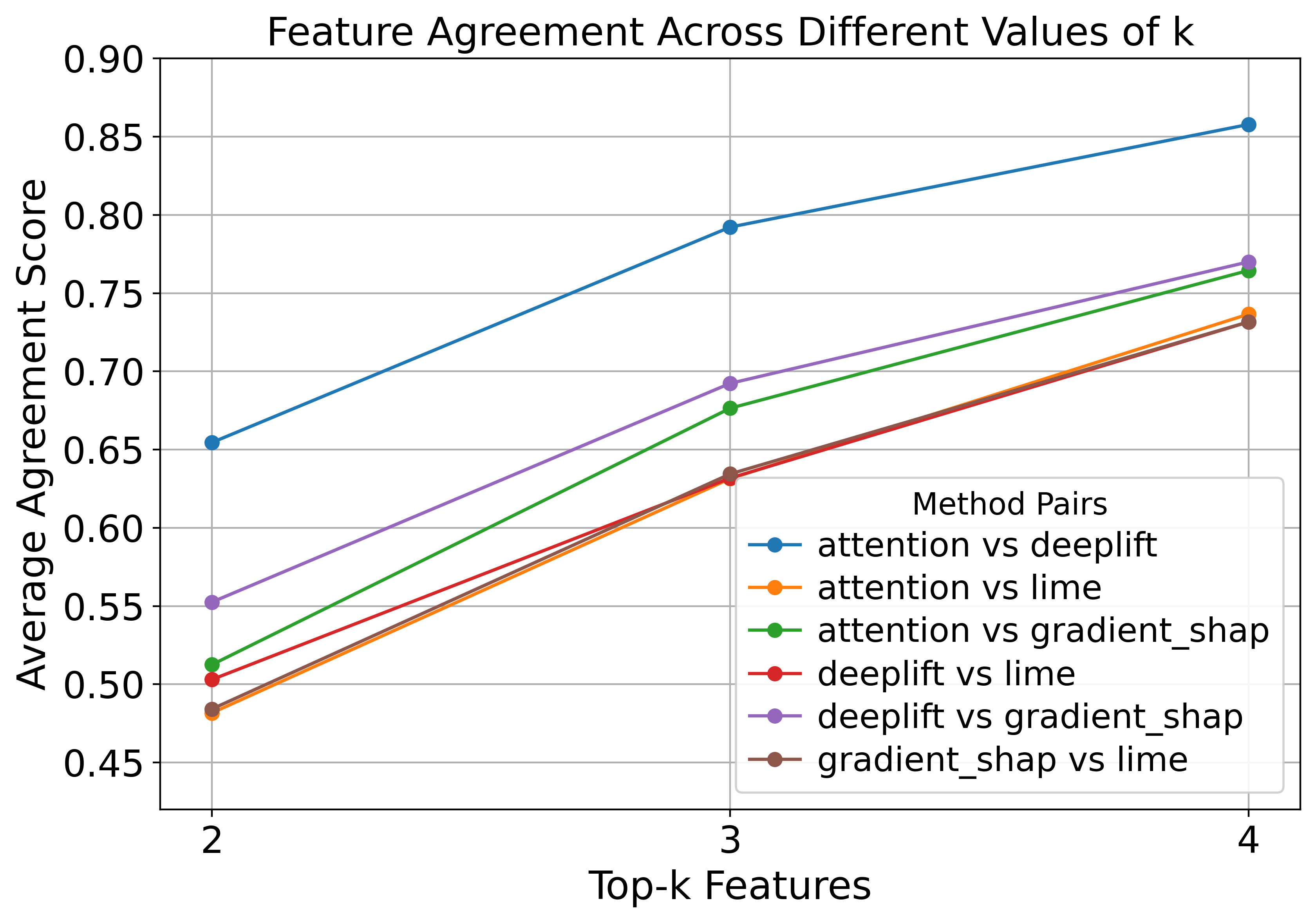}
        \caption{Feature Agreement Line-plot for Xsum dataset(segmented articles)}
    \end{subfigure}
    \hfill
    \begin{subfigure}[b]{0.49\textwidth}
        \centering
        \includegraphics[width=\textwidth]{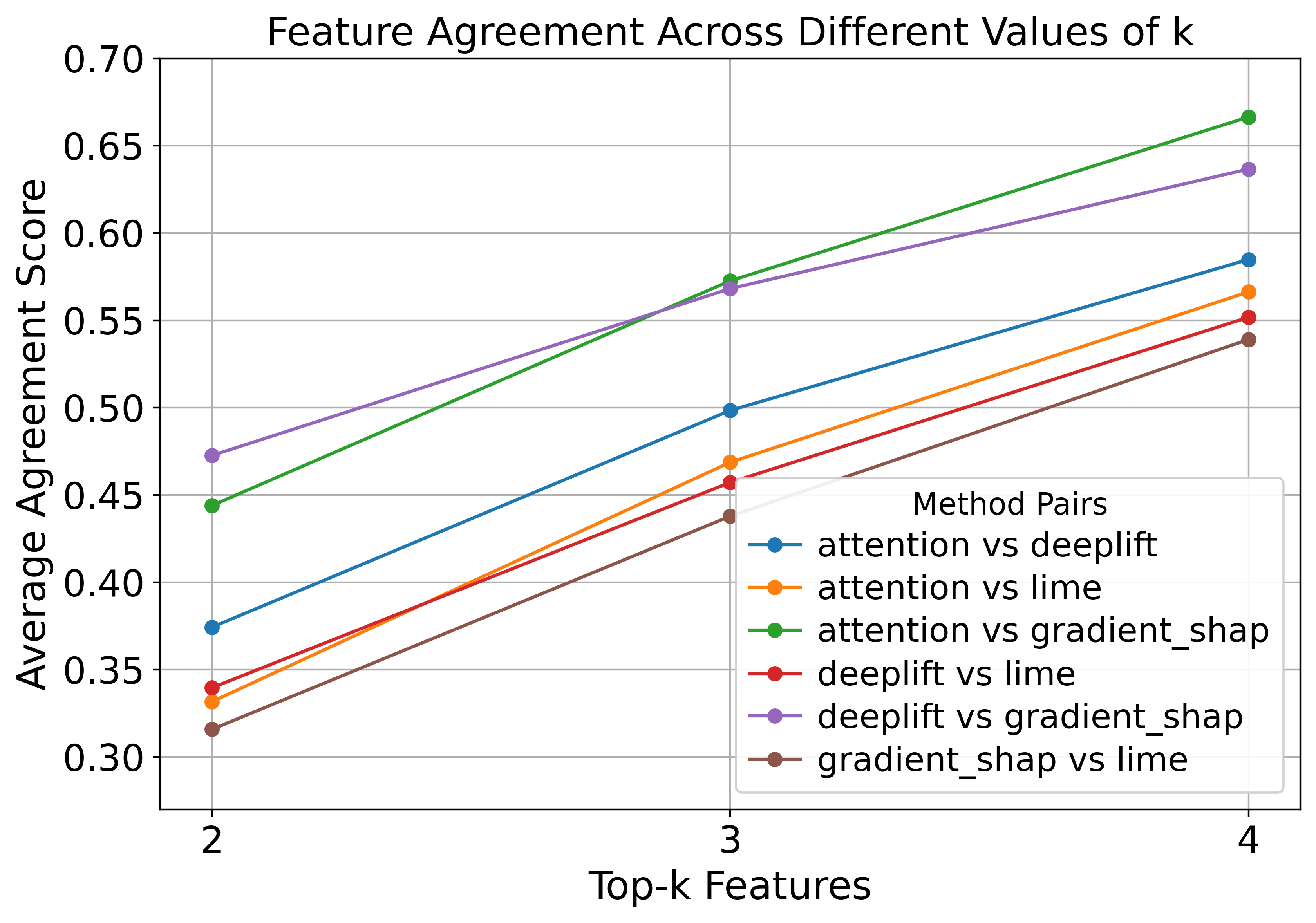}
        \caption{Feature Agreement Line-plot for CNN/DM dataset(segmented articles)}
    \end{subfigure}
    \caption{Average Feature Agreement Line-plots on XSum and CNN/DM datasets after Segmentation. The Line plot depicts similar trends as observed in the previous sample set of segmented articles.}
    \label{fig:regional_line-xsum-cnn}
\end{figure}

\begin{figure}[H]
    \centering
    \begin{subfigure}[b]{0.49\textwidth}
        \centering
        \includegraphics[width=\textwidth]{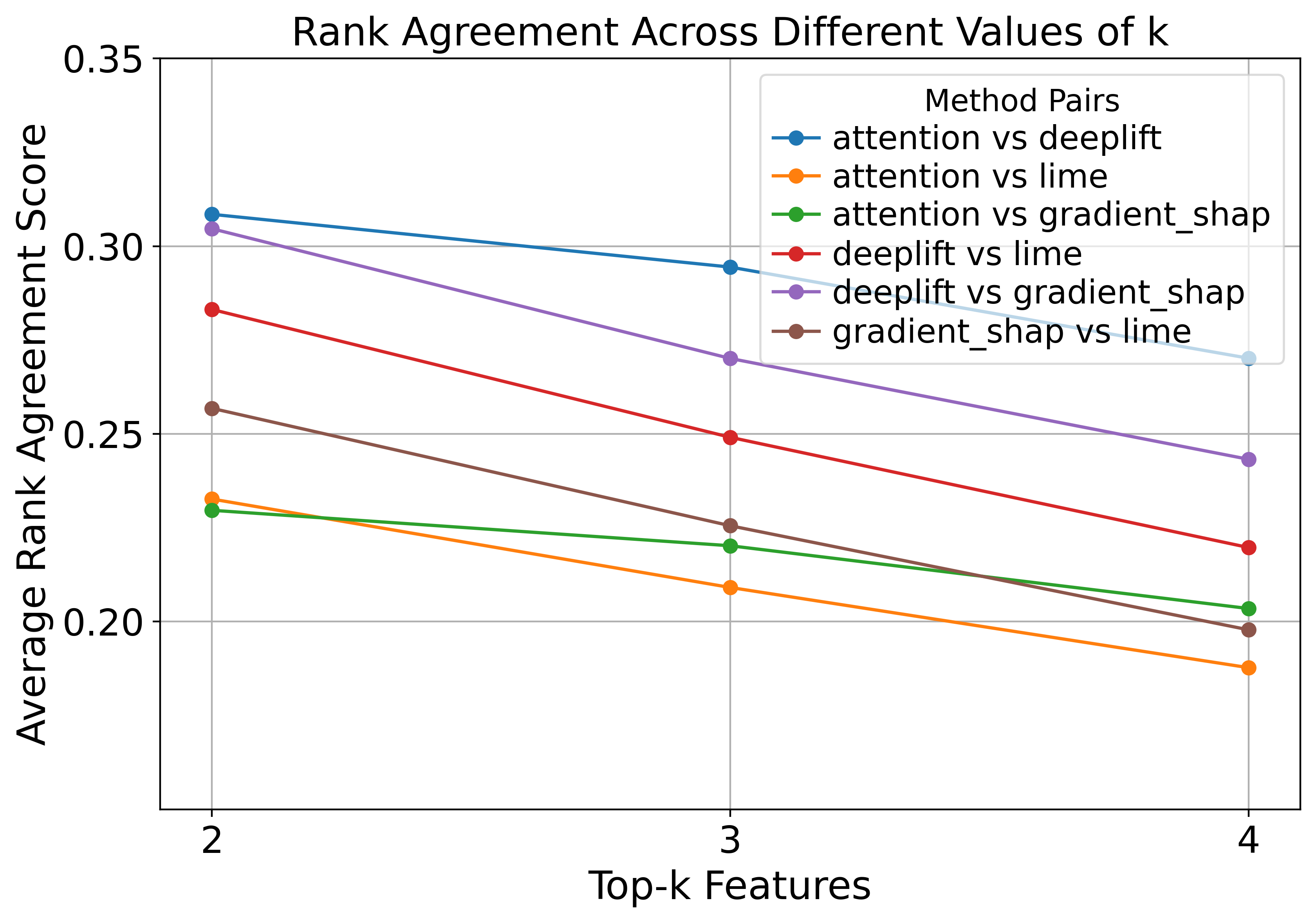}
        \caption{Rank Agreement Line-plot for Xsum dataset(segmented articles)}
    \end{subfigure}
    \hfill
    \begin{subfigure}[b]{0.49\textwidth}
        \centering
        \includegraphics[width=\textwidth]{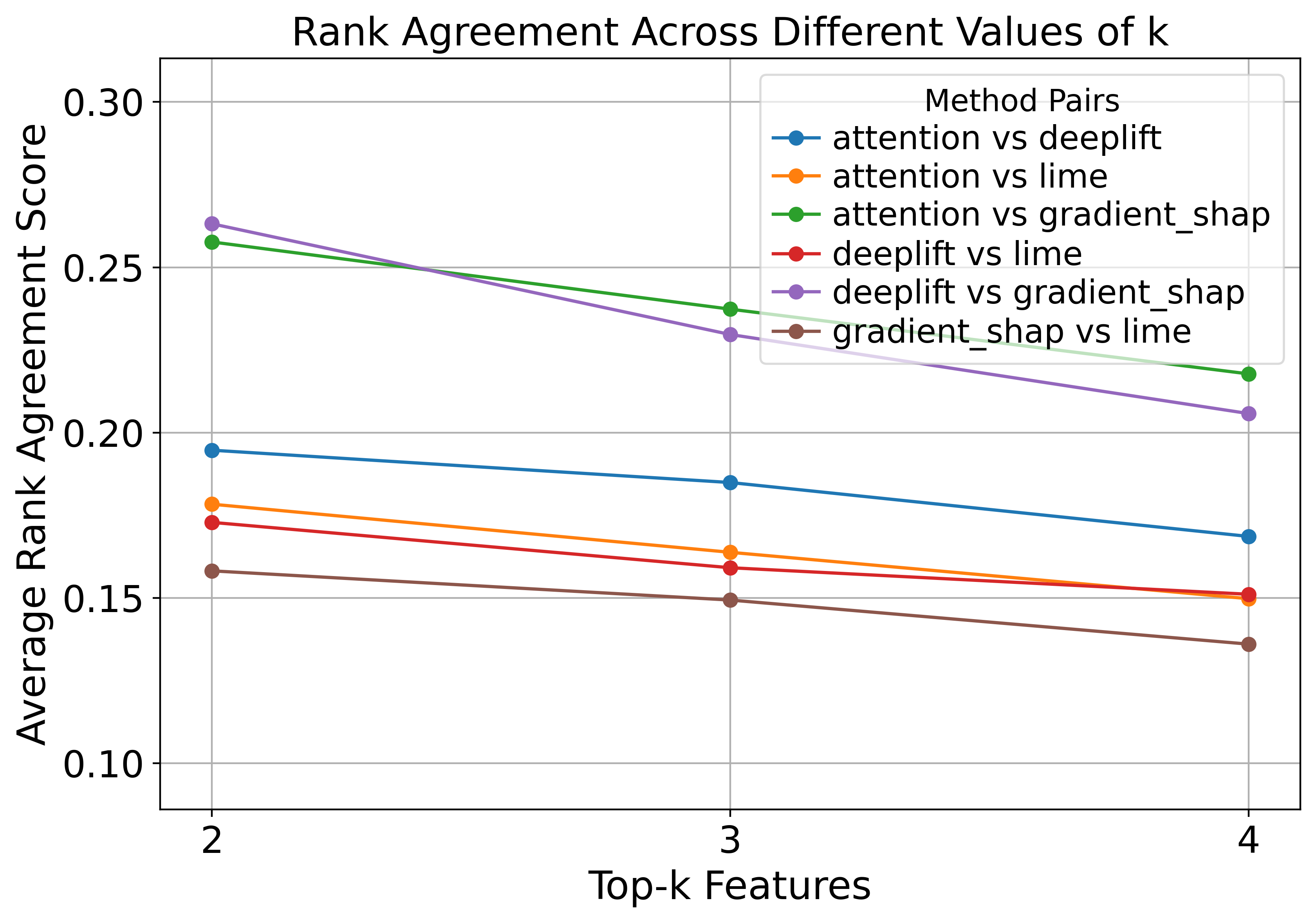}
        \caption{Rank Agreement Line-plot for CNN/DM dataset(segmented articles)}
    \end{subfigure}
    \caption{Average Rank Agreement Line-plots on XSum and CNN/DM datasets after Segmentation. The Line plot depicts similar trends as observed in the previous sample set of segmented articles.}
    \label{fig:regional_ra_line-xsum-cnn}
\end{figure}

\begin{figure}[H]
    \centering
    \begin{subfigure}[b]{0.44\textwidth}
        \centering
        \includegraphics[width=\textwidth]{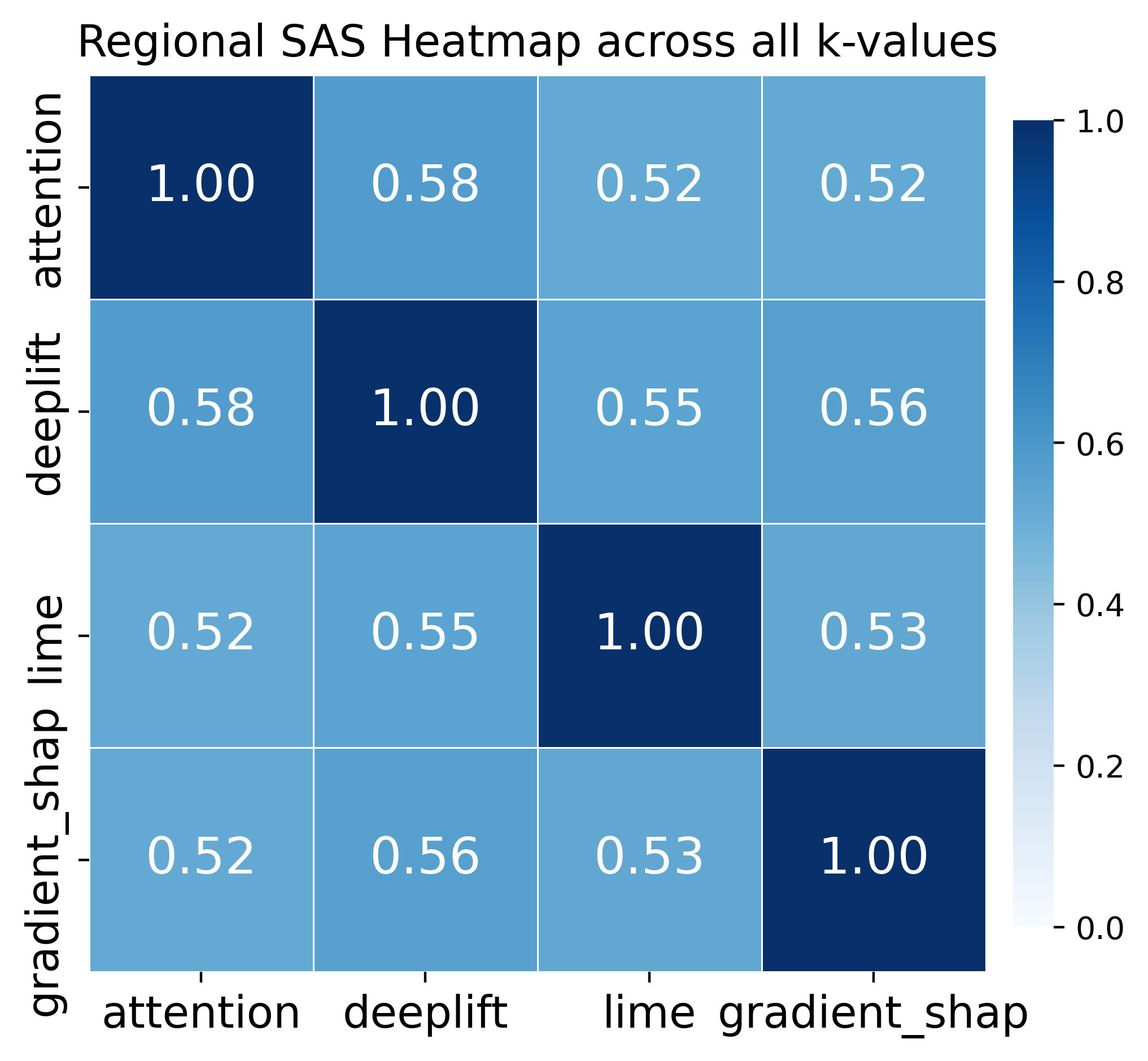}
        \caption{Regional XSum Semantic Alignment score across k values (Segmented articles)}
    \end{subfigure}
    \hfill
    \begin{subfigure}[b]{0.44\textwidth}
        \centering
        \includegraphics[width=\textwidth]{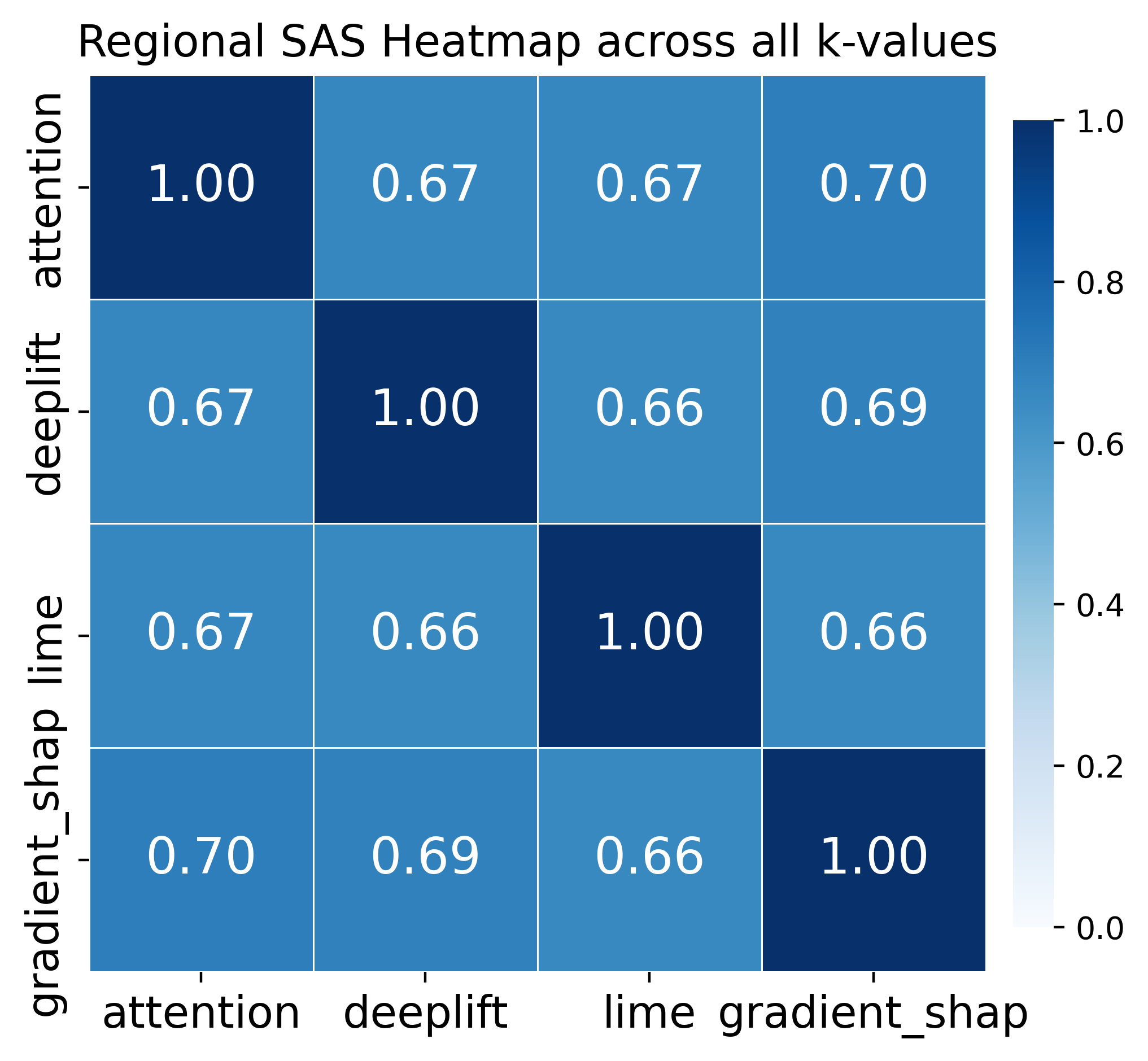}
        \caption{Regional CNN/DM Semantic Alignment Score across k values (Segmented articles)}
    \end{subfigure}
    \caption{SAS across XAI methods on XSum and CNN datasets after Segmentation. The trend of SAS scores is the same as the previously analyzed sample.}
    \label{fig:regional_CAS-xsum-cnn}
\end{figure}

\begin{figure}[H]
    \centering
    \begin{subfigure}[b]{0.44\textwidth}
        \centering
        \includegraphics[width=\textwidth]{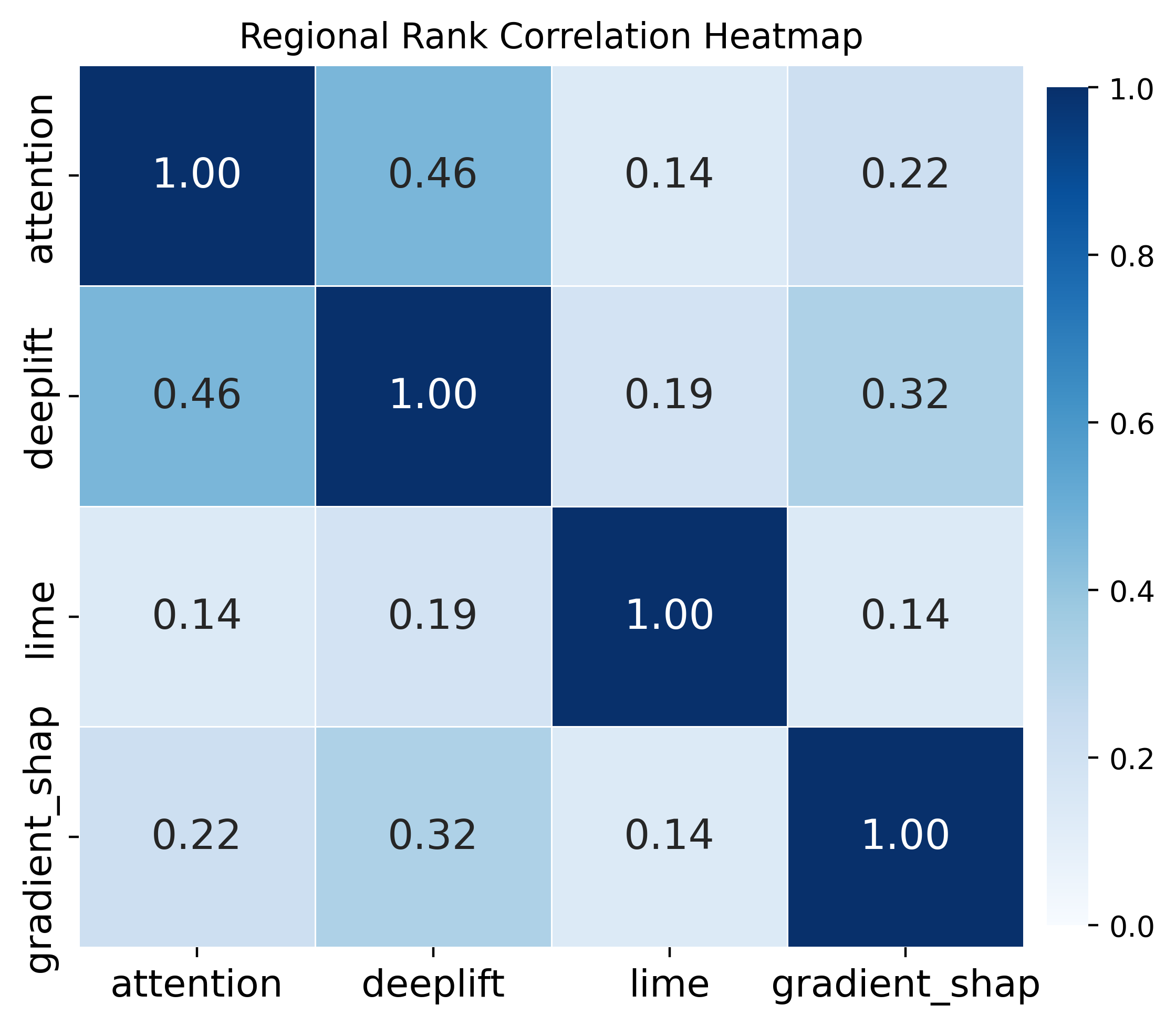}
        \caption{Spearman Rank Correlation for Xsum dataset (Segmented articles)}
    \end{subfigure}
    \hfill
    \begin{subfigure}[b]{0.44\textwidth}
        \centering
        \includegraphics[width=\textwidth]{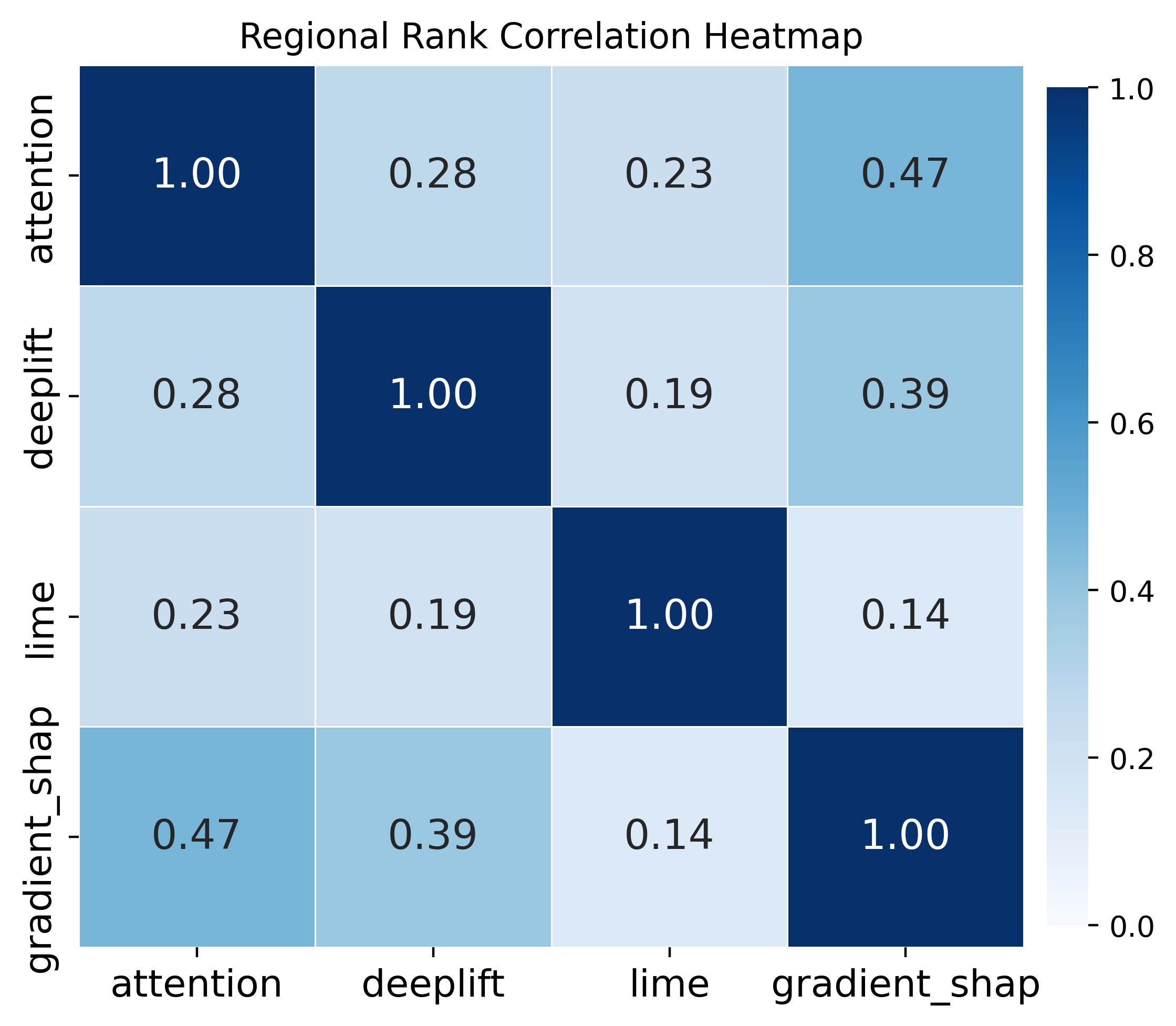}
        \caption{Spearman Rank Correlation for CNN/DM dataset (Segmented articles)}
    \end{subfigure}
    \caption{Spearman Rank Correlation across XAI methods on XSum and CNN datasets after segmentation. The results of Spearman Rank Agreement in this batch are also consistent.}
    \label{fig:regional_SRA-xsum-cnn}
\end{figure}

To assess the effectiveness of the RXAI framework and how consistent it is in enhancing agreement between XAI methods, we have added the RXAI analysis on batch 2 below. Here, we present batch-level regional agreement comparisons (Batch 2 from the XSum and CNN/DM dataset) across four key metrics: SAS, Spearman Rank Correlation, Feature Agreement, and Rank Agreement. The results of the RXAI approach on Batch 2, presented in figures: \ref{fig:regional_line-xsum-cnn}, \ref{fig:regional_ra_line-xsum-cnn}, \ref{fig:regional_CAS-xsum-cnn} and \ref{fig:regional_SRA-xsum-cnn} illustrate similar agreement enhancement patterns between XAI methods as observed in section: \ref{Disgreement_results}. While the variations are observed between the enhanced agreement scores, they vary with a low standard deviation value as highlighted in Table \ref{tab:fa_xsum_cnn} and Table \ref{tab:ra_summary_xsum_cnn}. This consistency of improvement across all 11 batches supports the hypothesis and robustness of our findings reported in the main results section.

\subsubsection{Statistical Test Results and analysis} \label{stats_test}

\begin{enumerate}
    \item \textbf{One-way ANOVA:}  
    \begin{enumerate}
        \item \textbf{Conducting ANOVA on Global Disagreement Results:}
    
    This test was used to determine whether the mean global agreement scores (Feature Agreement and Rank Agreement) significantly differ across 11 independently sampled batches. The test was conducted for each method pair and top-k value ($k \in \{2, \dots, 10\}$), resulting in 54 tests per agreement type. As shown in Table~\ref{tab:anova_summary}, the XSum dataset showed relatively stable behavior across batches, with only 6\% of FA tests and 6\% of RA tests found to be significant ($p < 0.05$). In contrast, the CNN dataset exhibited high batch sensitivity in FA scores (88.9\% significant) and moderate variation in RA scores (42.6\% significant).
    
  \item \textbf{Conducting ANOVA on Local Disagreement Results:} The same test was executed across all 11 batches of Regional agreement analysis scores. The RXAI framework helps in improving agreement; however, the magnitude of improvement varies across batches for both Xsum and CNN/DM datasets. Hence, the p-values across batches of RXAI analysis were below 0.05, indicating that the agreement scores are not consistent across batches after applying RXAI, which may be attributed to segment length across batches. However, the analysis done in table \ref{tab:ra_summary_xsum_cnn} and \ref{tab:fa_xsum_cnn} shows a very low value of variance across the mean score of agreement for all batches. 
\begin{table}[h]
\centering
\caption{One-way ANOVA summary for global agreement score consistency across 11 batches.}
\label{tab:anova_summary}
\begin{tabular}{lcccc}
\toprule
\textbf{Dataset} & \textbf{Agreement} & \textbf{Significant} & \textbf{Not-Significant} & \textbf{\% Not Significant} \\
\midrule
XSum & Feature Agreement (FA) & 3 & 51 & 94\% \\
XSum & Rank Agreement (RA)    & 3 & 51 & 94\% \\
CNN  & Feature Agreement (FA) & 48 & 6  & 11.0\% \\
CNN  & Rank Agreement (RA)    & 23 & 31 & 57.6\% \\
\bottomrule
\end{tabular}
\end{table}

\end{enumerate}

    \item \textbf{Wilcoxon Signed-Rank Test:} To evaluate the impact of RXAI approach on agreement between XAI methods, we applied the Wilcoxon signed-rank test, a non-parametric test for paired data. We compared agreement scores before (global) and after segmentation (regional) for each article, across method pairs and $k \in \{2, 3, 4\}$, over 11 batches. The results show strong improvements. For the XSum dataset, $\sim$99\% of Feature Agreement(FA) and $\sim$83\% of Rank Agreement (RA) comparisons across 11 batches showed statistically significant improvement ($p < 0.05$ with very low p-values). Similarly for CNN/DM dataset approximately $\sim$97\% of FA and $\sim$85\% of RA comparisons across all batches showed significant improvement. These results confirms the effectiveness of RXAI approach, which helps in enhancing explanation consistency. Table: \ref{tab:wilcoxon_xsum_fa} and \ref{tab:wilcoxon_batch2_cnn} shows sample results of p-values.

    \begin{table}[h]
\centering
\caption{Wilcoxon signed-rank test $p$-values for Feature Agreement values across Xsum dataset's representative batch of 500 samples discussed in section \ref{Disgreement_results}. The p-values clearly indicate that there is a significant difference between agreement scores before and after segmentation for each method pair at different k-values, supporting our hypothesis.}
\label{tab:wilcoxon_xsum_fa}
\begin{tabular}{lccc}
\toprule
\textbf{Method Pair} & \textbf{$k=2$} & \textbf{$k=3$} & \textbf{$k=4$} \\
\midrule
attention vs DeepLIFT         & $7.751 \times 10^{-20}$ & $2.156 \times 10^{-29}$ & $1.622 \times 10^{-43}$ \\
attention vs lime             & $2.586 \times 10^{-42}$ & $1.547 \times 10^{-54}$ & $1.610 \times 10^{-66}$ \\
attention vs Gradient\_SHAP   & $2.287 \times 10^{-38}$ & $2.131 \times 10^{-47}$ & $5.710 \times 10^{-61}$ \\
DeepLIFT vs lime              & $1.858 \times 10^{-42}$ & $2.456 \times 10^{-56}$ & $1.942 \times 10^{-65}$ \\
DeepLIFT vs Gradient\_SHAP    & $1.662 \times 10^{-36}$ & $1.713 \times 10^{-49}$ & $6.463 \times 10^{-61}$ \\
lime vs Gradient\_SHAP        & $8.405 \times 10^{-41}$ & $2.384 \times 10^{-52}$ & $1.052 \times 10^{-64}$ \\
\bottomrule
\end{tabular}
\end{table}

\end{enumerate}

\begin{table}[h]
\centering
\caption{Wilcoxon signed-rank test $p$-values for FA values across CNN/DM dataset's representative batch of 500 samples discussed in section \ref{Disgreement_results}. The p-values clearly indicate that there is a significant difference between agreement scores before and after segmentation for each method pair at different k-values, supporting our hypothesis. }
\label{tab:wilcoxon_batch2_cnn}
\begin{tabular}{lccc}
\toprule
\textbf{Method Pair} & \textbf{$k=2$} & \textbf{$k=3$} & \textbf{$k=4$} \\
\midrule
attention vs DeepLIFT         & $5.185 \times 10^{-1}$  & $1.370 \times 10^{-7}$  & $1.312 \times 10^{-25}$ \\
attention vs lime             & $6.122 \times 10^{-37}$ & $1.441 \times 10^{-65}$ & $4.503 \times 10^{-76}$ \\
attention vs Gradient\_SHAP   & $1.990 \times 10^{-27}$ & $2.399 \times 10^{-50}$ & $3.426 \times 10^{-71}$ \\
DeepLIFT vs lime              & $5.400 \times 10^{-37}$ & $3.888 \times 10^{-65}$ & $2.578 \times 10^{-74}$ \\
DeepLIFT vs Gradient\_SHAP    & $1.510 \times 10^{-37}$ & $1.699 \times 10^{-57}$ & $8.874 \times 10^{-72}$ \\
lime vs Gradient\_SHAP        & $1.631 \times 10^{-33}$ & $7.157 \times 10^{-66}$ & $1.634 \times 10^{-74}$ \\
\bottomrule
\end{tabular}
\end{table}

\subsubsection{Impact of RXAI Framework:} \label{rxai:impact} From the empirical analysis and the statistical test results it is evident that RXAI framework helps in mitigating the disagreement problem. However, the framework relies on clustering the text data in meaningful, coherent segments using k-means clustering, which may introduce semantic or coherence drift between input articles and segments. To assess this, we have conducted an analysis in two steps:

\begin{enumerate}
    \item Comparing semantic similarity between input articles and their segmented version across both datasets: Table \ref{tab:semantic_similarity_all_batches} illustrates the semantic similarity results on two representative samples of Xsum and CNN/DM datasets. The similarity is measured between the Input article and the segmented article (concatenated), as well as the input article with individual segments. The global similarity score with the concatenated segment text remains high ($\geq$0.90) for both datasets. The segment-wise score is, however, slightly lower than the concatenated score. This analysis confirms that segmentation largely preserves the meaning of the article, validating the reliability of the RXAI framework.

\begin{table}[h]
\centering
\caption{Semantic similarity between original articles and segmented versions across two batches from XSum and CNN/DM datasets. Global similarity is computed using concatenated segments; segment-wise reflects the average of individual segment similarities.}
\label{tab:semantic_similarity_all_batches}
\begin{tabular}{lcccc}
\toprule
\textbf{Metric} & \textbf{XSum Batch-1} & \textbf{XSum Batch-2} & \textbf{CNN Batch-1} & \textbf{CNN Batch-2} \\
\midrule
\makecell Avg Segment-wise Similarity    & 0.77 ± 0.10 & 0.80 ± 0.10 & 0.76 ± 0.08 & 0.75 ± 0.01 \\
\makecell{Global Semantic Similarity \\ (Concatenated)} & 0.94 ± 0.07 & 0.97 ± 0.06 & 0.90 ± 0.08 & 0.92 ± 0.08 \\
\bottomrule
\end{tabular}
\end{table}

\item The Second step is to compare the Coherence score of the article and its segmented version across both datasets.
To verify coherence drift, we utilized the Coherence Momentum model from SGNLP \citep{jwala-etal-2022-rethinking}. This neural model employs a momentum encoder and hard negative mining during training to assess coherence. Given a piece of text, it outputs a coherence score, which is meaningful only in comparative settings. The text with the higher score is considered more coherent. In our study, we use this model to compare coherence between the original article and its segmented version. The coherence score of each input article and segmented article is calculated, along with the drift between the two scores. The average coherence value for the full article vs. the Segmented article is computed to analyze coherence preservation. In some cases, the coherence score of segmented articles is higher than the full article. 

The results of coherence analysis are depicted in Table \ref{tab:coherence_drift_split}. From the overall average coherence scores of Input articles vs segmented articles as well as the drift score highlights that the segmentation does introduce coherence drift; however, the drift is moderate, ranging from 4.8 to 8.1, with low standard deviation values. Overall, these results highlight that while RXAI introduces moderate coherence drift, it largely preserves the semantic structure and coherence between the input article and the segmented article.



\begin{table}[h]
\centering
\vspace{0.5em}

\begin{subtable}[t]{0.45\textwidth}
\centering
\small
\caption{Coherence analysis of XSum dataset across two batches}
\begin{tabular}{lcc}
\toprule
\textbf{Metric} & \textbf{Batch-1} & \textbf{Batch-2} \\
\midrule
Coherence Full (Mean)         & -9.499  & -8.223  \\
Coherence Segmented (Mean)    & -15.763 & -13.477 \\
Drift (Mean)                  & 6.264   & 5.255   \\
\midrule
Coherence Full (Std Dev)      & 9.356   & 8.645   \\
Coherence Segmented (Std Dev) & 8.378   & 8.386   \\
Drift (Std Dev)               & 6.279   & 7.450   \\
\bottomrule
\end{tabular}
\end{subtable}
\hspace{0.05\textwidth}
\begin{subtable}[t]{0.45\textwidth}
\centering
\small
\caption{Coherence analysis of CNN/DM dataset across two batches}
\begin{tabular}{lcc}
\toprule
\textbf{Metric} & \textbf{Batch-1} & \textbf{Batch-2} \\
\midrule
Coherence Full (Mean)         & -8.542  & -9.360  \\
Coherence Segmented (Mean)    & -13.365 & -17.466 \\
Drift (Mean)                  & 4.822   & 8.106   \\
\midrule
Coherence Full (Std Dev)      & 8.148   & 8.769   \\
Coherence Segmented (Std Dev) & 7.709   & 7.289   \\
Drift (Std Dev)               & 6.848   & 7.083   \\
\bottomrule
\end{tabular}
\end{subtable}

\vspace{0.5em}

\caption{Coherence scores and drift for XSum and CNN/DM datasets across two batches. Coherence Full refers to the original article, Coherence Segmented to the segmented version, and Drift is their absolute difference.}
\label{tab:coherence_drift_split}

\end{table}

\end{enumerate}





%

\section{JavaScript Visualization tool}
\begin{figure}[H]
    \centering
    \includegraphics[height=8.3cm]{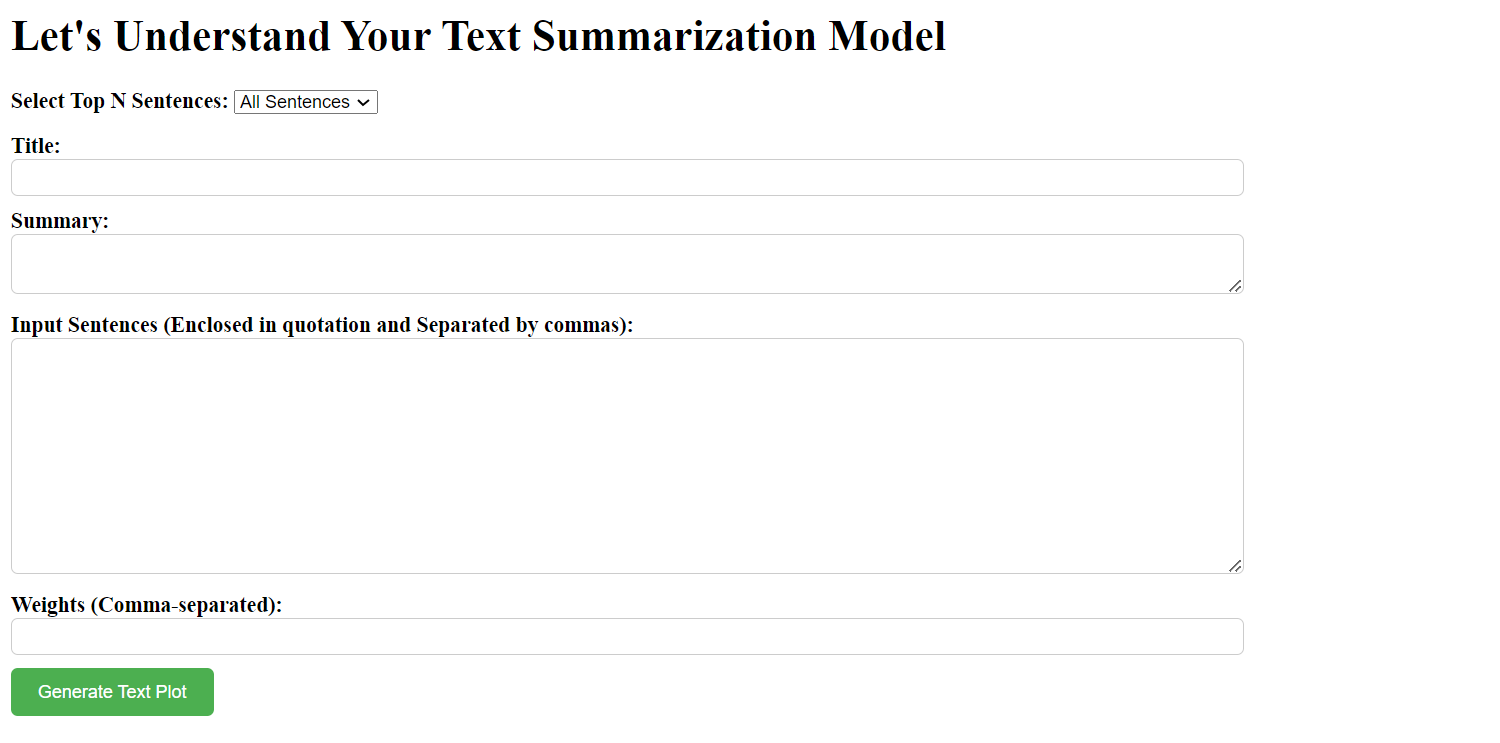}  
    \caption{JavaScript tool interface along with the input fields}
    \label{fig:Js_textplot}  
\end{figure}

The JavaScript visualization tool features a user-friendly interface for inputting text, attribution scores, and summaries. Fig. \ref{fig:Js_textplot} showcases the tool's interface, which accepts source sentences and their corresponding normalized attribution scores, allowing users to explore the influence of each sentence on the summary generation.

\end{appendices}

\end{document}